\documentclass[10pt,journal,compsoc]{IEEEtran}
\ifCLASSOPTIONcompsoc
  \usepackage[nocompress]{cite}
\else
  \usepackage{cite}
\fi
\ifCLASSINFOpdf
\else
\fi

\usepackage{amsfonts}
\usepackage{amsmath}
\usepackage{graphicx}
\usepackage{caption}
\usepackage{subcaption}
\usepackage{bm}
\usepackage{multirow}
\graphicspath{{./}{./figures/}}

\usepackage{xcolor}
\usepackage[linesnumbered,ruled,vlined]{algorithm2e}

\SetCommentSty{mycommfont}
\SetKwInput{KwInput}{Input}                
\SetKwInput{KwOutput}{Output} 

\usepackage[bookmarks,backref=true,linkcolor=black]{hyperref} 
\hypersetup{
  pdfauthor = {},
  pdftitle = {},
  pdfsubject = {},
  pdfkeywords = {},
  colorlinks=true,
  linkcolor= black,
  citecolor= black,
  pageanchor=true,
  urlcolor = black,
  plainpages = false,
  linktocpage
}

\usepackage{color} 
\usepackage{colortbl}
\usepackage{cases}
\definecolor{light-gray}{gray}{0.6}
\definecolor{lavender}{rgb}{0.5,0.5,1.0}

\newcommand{\hot}[1]{{#1}}

\usepackage{verbatim}
\usepackage{url}
\usepackage{gensymb}

\usepackage{xifthen}
\newcommand{\matD}{{\mathbf{D}}}
\newcommand{\matA}{{\mathbf{A}}}
\newcommand{\matT}{{\mathbf{T}}}
\newcommand{\matM}{{\mathbf{M}}}
\newcommand{\matTAt}[1]{{\matT^{(#1)}}}

\newcommand{\vecB}{{\mathbf{b}}}
\newcommand{\vecS}{{\mathbf{s}}}
\newcommand{\vecN}{{\mathbf{n}}}
\newcommand{\estVecB}{{\hat{\vecB}}}
\newcommand{\vecBAt}[1]{{\vecB^{(#1)}}}
\newcommand{\vecSAt}[1]{{\vecS^{(#1)}}}
\newcommand{\vecNAt}[1]{{\vecN^{(#1)}}}
\newcommand{\estVecBAt}[1]{{\hat{\vecB}^{(#1)}}}

\newcommand{\estVecNAt}[1]{{\hat{\vecN}^{(#1)}}}
\newcommand{\estB}[2]{{\hat{b}_{#1}^{(#2)}}}

\newcommand{\fon}{FON}
\newcommand{\fonPlus}{FON+}
\newcommand{\semantic}{VAR}
\newcommand{\semanticPlus}{VAR+}
\newcommand{\fixed}{Fixed}
\newcommand{\fixedPlus}{Fixed+}
\newcommand{\ours}{FlowHON}

\newcommand{\bernard}{B{\'e}nard}

\DeclareMathOperator{\kld}{d_{KL}}
\DeclareMathOperator{\euclid}{d_E}

\DeclareMathOperator{\nonzero}{nonzero}

\begin{document}
%
\title{FlowHON: Representing Flow Fields Using Higher-Order Networks}
%
%

\author{
  Nan~Chen,
  Zhihong~Li,
  and Jun~Tao*,~\IEEEmembership{Member,~IEEE}
\IEEEcompsocitemizethanks{
\IEEEcompsocthanksitem * denote the corresponding author.
\IEEEcompsocthanksitem N. Chen is with the School of Computer Science and Engineering, Sun Yat-sen University and National University of Singapore. E-mail: chenn53@mail2.sysu.edu.cn. The majority of this work was conducted during his undergraduate studies at SYSU.
\IEEEcompsocthanksitem Z. Li and J. Tao are with the School of Computer Science and Engineering, Sun Yat-sen University and the National Supercomputer Center in Guangzhou, China. J. Tao is the corresponding author. E-mail: lizhh236@mail2.sysu.edu.cn, taoj23@mail.sysu.edu.cn. 
\protect
}
}


%
%
%
\markboth{IEEE Transactions on Visualization and Computer Graphics}{Chen et al.: FlowHON: Representing Flow Fields Using Higher-Order Networks}
%



\IEEEtitleabstractindextext{%
\begin{abstract}
Flow fields are often partitioned into data blocks for massively parallel computation and analysis based on blockwise relationships. However, most of the previous techniques only consider the first-order dependencies among blocks, which is insufficient in describing complex flow patterns. In this work, we present FlowHON, an approach to construct higher-order networks (HONs) from flow fields. FlowHON captures the inherent higher-order dependencies in flow fields as nodes and estimates the transitions among them as edges. We formulate the HON construction as an optimization problem with three linear transformations. The first two layers correspond to the node generation and the third one corresponds to edge estimation. Our formulation allows the node generation and edge estimation to be solved in a unified framework. With FlowHON, the rich set of traditional graph algorithms can be applied without any modification to analyze flow fields, while leveraging the higher-order information to understand the inherent structure and manage flow data for efficiency. We demonstrate the effectiveness of FlowHON using a series of downstream tasks, including estimating the density of particles during tracing, partitioning flow fields for data management, and understanding flow fields using the node-link diagram representation of networks.
\end{abstract}

\begin{IEEEkeywords}
 Flow visualization, higher-order network, data transformation, data partition, and task distribution.
\end{IEEEkeywords}

}

\maketitle

\IEEEdisplaynontitleabstractindextext

%

\IEEEraisesectionheading{\section{Introduction}\label{sec:introduction}}

Flow visualization plays a vital role in understanding dynamic systems for various domains and applications. In the past decades, flow visualization has been studied extensively, and many techniques were developed to effectively visualize and analyze flow fields. Recently, due to the increasing size and complexity of the simulated flow fields, many approaches partition the flow data into blocks for further processing or analysis. The data partitioning reduces the size of data processed by each computing node for scalability and allows the structure of flow fields to be understood at the block level. 
At the core of these techniques, the graph is used either as a data structure for graph algorithms to analyze flow fields, or as a visual representation to enable clear observation and easy interaction in 2D.

Although developed for different scenarios, existing graph-based techniques usually share a similar construction process. Nodes in a graph represent data blocks, and edges represent the transition probabilities among data blocks. The transition probabilities are estimated empirically based on the number of particles moving between blocks. In this way, the graph provides affinity relationships between blocks and captures the structure of the flow field. Analysis of the graph facilitates a series of downstream tasks, including data partition ~\cite{nouanesengsy2011load}, data prefetching ~\cite{guo2014advection, gerndt2004viracocha, zhang2016efficient}, and particle advection scheduling~\cite{chen2013graph}. By applying layout algorithms, the graph may be used to present the flow structure in a compact way without occlusion~\cite{xu2010flow} as well. 

However, the conventional graph-based flow visualization techniques usually assume the Markovian process in describing the relationships among blocks, which could be inaccurate. The Markovian assumption implies that the particles in the same block will follow the same transition probability distribution when moving to the next block. This assumption does not hold in most cases, especially for blocks with complex flow behavior. Some approaches may employ a multi-resolution partitioning strategy to further divide the complicated block. For example, the FlowGraph~\cite{ma2013flowgraph} evaluates the entropy of flow directions in blocks to guide the partitioning. But this strategy may require a great amount of blocks to precisely describe the curvy boundaries of regions.

To the best of our knowledge, Zhang et al.\cite{zhang2016efficient}) is the only existing approach that considers higher-order dependencies. But this technique still fails to incorporate the connections among higher-order dependencies. It captures only the local higher-order patterns but not the global structure of entire flow fields. Therefore, this technique may not be easily extended to support tasks such as data partitioning, workload balancing, and decomposition of flow fields, where the global structure is often needed.

In this paper, we aim to capture block-wise higher-order dependencies in a flow field at a global scale. Toward this end, we extract the higher-order dependencies among blocks and organize these dependencies as a higher-order network. The higher-order dependencies allow different flow behaviors in a single data block to be separated so that the flow transition patterns among data blocks can be accurately described. And the network connecting the higher-order dependencies allows the dependencies to be studied at a larger scale, and provides a compatible interface for existing network analytic algorithms to be applied. The main challenges to achieving this goal can be summarized into three aspects. The first challenge is to extract higher-order dependencies that can precisely model diverse flow patterns in the flow field and avoid redundant higher-order dependencies at the same time. The second challenge is to approximate the transition probabilities between nodes in the network, so that the flow behavior can be accurately described. The final challenge is to establish a connection between network analytic methods with flow fields, which makes it possible to analyze flow fields by analyzing corresponding higher-order networks.

To tackle the above challenges, we propose FlowHON, a unified framework to construct higher-order networks from flow fields. The framework formulates the HON construction problem as an optimization problem with three linear transformations, including two linear layers for node generation and one layer for transition estimation. This formulation generalizes the existing HON construction algorithms. Therefore, it may lead to potentially better performance with the existing approaches being special solutions to our optimization problem. We propose an efficient approach to optimize the node generation and transition estimation in a unified framework and examine the performance of our approach using several downstream tasks with a variety of data sets. The tasks include estimating particle transitions by using random walks on our network, data partitioning by applying a community detection algorithm, and visualizing the flow field structure by leveraging graph layout. We demonstrate the effectiveness by comparing our approach with the existing graph-based approaches using these tasks.
\section{Related Work}

{\bf Graph-based techniques for flow visualization.}
Graph-based approaches have received considerable attention from the scientific visualization community in various kinds of applications~\cite{wang2017graphs}. In flow visualization, several graph-based techniques were developed to describe the access pattern among blocks during particle tracing.
Bhatia et al.~\cite{bhatia2011flow} designed edge maps for triangular meshes, which mapped the entry and exit points of streamlines on the boundary of individual triangles.
Chen et al.~\cite{chen2011flow} proposed the access dependency graph to assess the dependencies between different data blocks in the flow field and used it to guide the file layout for improved I/O performance.

Chen et al.~\cite{chen2012flow} proposed the N-hop access dependency graph that further considers the N-hop transitions.
Chen et al.~\cite{chen2013graph} applied discrete-time Markov chains on node-link graphs to predict particle trajectories on time-varying flow fields to guide seed advection schedule.

Nouanesengsy et al.~\cite{nouanesengsy2011load} utilized a flow graph with initial seed locations to generate an estimate of the workload of each data block during parallel streamline generation.
Guo et al.~\cite{guo2014advection} developed a graph-based model that could be constructed on the fly to predict data access for data block prefetching. 
Gerndt et al.~\cite{gerndt2004viracocha} applied a similar strategy to build a first-order probability graph that characterized the successor relation of blocks in CFD data sets. 
Zhang et al.~\cite{zhang2016efficient} applied higher-order dependencies among data blocks to predict the data access pattern and guide the data prefetching.
Zhang et al.~\cite{Zhang2018PacificVis} built an access dependency graph to estimate workload.
Other approaches applied graphs to understand the structures of flow fields, such as Morse Connection Graphs~\cite{Chen2008TVCGMorse}, Flow Web~\cite{xu2010flow}, FlowGraph~\cite{ma2013flowgraph,ma2013graph}, Flow topology graph (FTG)~\cite{Aldrich2017FlowTopologyGraph}, and Semantic Flow Graph~\cite{tao2017semantic}.

{\bf Parallel tracing and data management.}

Parallel particle tracing algorithms generally fall into three main categories: data-parallelism, task-parallelism, and hybrid-parallelism~\cite{zhang2018survey}. The data-parallel mechanism distributes data blocks to computation nodes and exchanges particles during tracing.

Yu et al.\cite{Yu2007ParallelHierarchical} partitioned flow data based on the hierarchical representation of data blocks. Moloney et al. \cite{Moloney2007EuroVis} applied k-d tree to divide the dataset of uniform grid for load balancing in sort-first parallel direct volume rendering. Based on a similar idea, Zhang et al. \cite{Zhang2018TVCGDynamic} applied k-d tree decomposition to balance workload during parallel particle tracing. Chen et al. ~\cite{Chen2008PacificVis} partitioned flow data based on flow direction and features. Peterka et al. \cite{Peterka2011Astudy} employed a static round-robin partition algorithm to distribute data blocks among processes. Nouanesengsy et al.\cite{Nouanesengsy2012Parallel} partitioned flow data into mutually exclusive spans of time for high-resolution FTLE computation. Graph-based models are generally employed to estimate the workload of each data block during running time, which cooperates with workload-aware allocation methods to generate the optimal data distribution among processors~\cite{nouanesengsy2011load}.

The task-parallel mechanism distributes the tracing tasks to computation nodes, which load data blocks on demand. 

Task-parallel tracing frameworks usually integrate methods such as data prefetching as well as file layout rearrangement to exploit data locality and to boost I/O performance. Many of these approaches leveraged graph-based representation to guide the file layout~\cite{chen2011flow,chen2012flow}, data prefetching~\cite{guo2014advection,gerndt2004viracocha,zhang2016efficient}, and task grouping~\cite{chen2013graph}.

Hong et al. ~\cite{Hong2018PacificVis} employed an LSTM based model to estimate the access pattern for parallel particle tracing in flow fields. 

{\bf Particle density estimation.}

Reich et al.~\cite{Reich2012TVCG} applied time-discrete Markov-chains on static unstructured flow fields to estimate particle distributions over time given the initial particle distributions. Hollt et al.~\cite{envirvis.20151090} used first-order forward tracking to estimate the trajectory of a particle originating in a specific cell. Guo et al.~\cite{guo2018extreme} introduced a divide-and-conquer mechanism to compute stochastic flow maps, where they decoupled the time domain into short periods, performed Monte Carlo particle tracing for each subinterval independently, and then composed the results to approximate the particle distribution for a longer period.

Our approach falls into the category of graph-based approaches. But different from the existing approaches, ours is the only one that leverages the higher-order dependencies and their connections to describe flow fields at a refined level. Most similar to ours are the approaches that consider N-hop or higher-order dependencies. Chen et al.~\cite{chen2012flow} included N-hop transitions among blocks into dependency graphs. But this construction only compensates the underestimated long-term dependencies, but does not provide a refined level of behaviors inside each individual block. Zhang et al.~\cite{zhang2016efficient} considered higher-order dependencies to distinguish different flow behaviors in individual blocks. But the dependencies are only used for data prefetching, and their connections are not considered. Therefore, this approach does not describe the higher-order dependencies at a global level.

\hot{{\bf Higher-order network related.}

Conventional network models are usually based on the assumption of Markovian behavior, meaning that the transition probability on a specific node only depends on its current state. Recent work has empirically demonstrated that this assumption is insufficient to model the real-world transitions~\cite{song2010limits,takaguchi2011predictability, chierichetti2012web}. Rosvall et al.~\cite{rosvall2014memory} proposed a second-order Markov model composed of memory nodes that encode the currently visited node as well as a previously visited node. Xu et al.~\cite{xu2016representing} put forward a higher-order network (HON) with nodes of variable orders. This approach only generates necessary higher-order nodes which behave significantly differently from their respective lower-order nodes. Edler et al.~\cite{edler2017mapping} abstracted different forms of higher-order networks as a sparse memory network that distinguished physical nodes from state nodes encoding higher-order dependencies, and employed a generalized map equation algorithm on it to detect overlapping module patterns. The higher-order network is also used in visualization approaches to examine long-term dependencies. Tao et al.~\cite{tao2017honvis} developed a visual analytic framework of HON that allows users to examine higher-order Markov dependencies interactively at different levels of granularity.
}
\begin{figure}[t]
\begin{center}
$\begin{array}{c@{\hspace{0.05in}}c}
\includegraphics[height=0.9in]{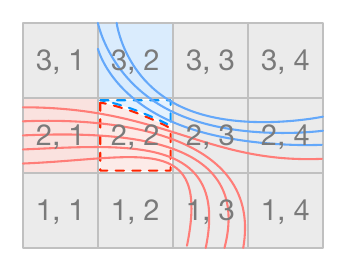}&
\includegraphics[height=0.9in]{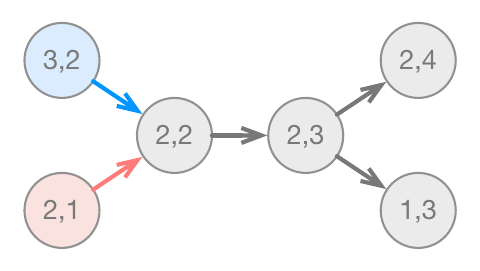}\\
\mbox{(a)} & \mbox{(b)}
\\
\end{array}$
$\begin{array}{c}
\includegraphics[height=1.0in]{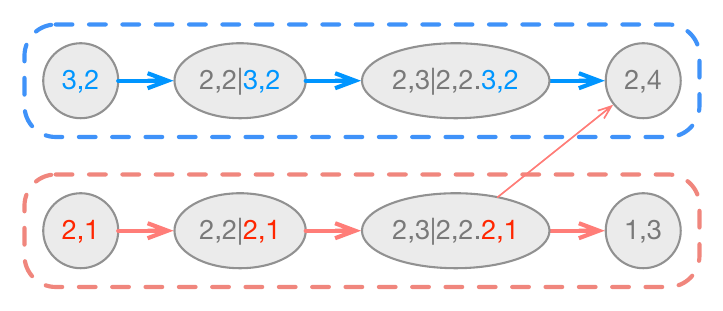}\\
\mbox{(c)}
\end{array}$
\end{center}
\caption{An illustrative example of the higher-order network. (a) shows a vector field uniformly partitioned into a $4 \times 3$ grid. (b) shows the corresponding first-order network. (c) shows the corresponding higher-order network.}
\label{fig:intro}
\end{figure}

\section{Higher-order Network for Flow}

We introduce \emph{FlowHON}, a higher-order network (HON) for flow visualization. Our goal is to encode the higher-order Markov dependencies in the graph representation, which can be leveraged by the existing graph-based visualization techniques to improve their performance without modification. The higher-order dependency indicates that the transition probability relies on not only the current state but also a series of previous states. The higher-order dependency commonly exists in many real-world applications, but it is not exploited by the conventional graph-based approaches. The conventional approaches are often built upon first-order networks (FON), which is insufficient to describe complicated transition patterns.
 
In this section, we will briefly introduce the concepts of HON in the context of flow visualization and explain how the HON facilitates the analysis of flow fields.

{\bf Higher-order dependency}.
Figure~\ref{fig:intro} (a) illustrates an example of a flow field uniformly partitioned into a $4 \times 3$ grid. The streamlines exhibit two movement patterns, which are distinguished by their colors. The trajectories of particles starting from block $(3,2)$ are colored in blue, and those from block $(2,1)$ are in red. 

Traditional approaches (e.g., \cite{xu2010flow,ma2013flowgraph,ma2013graph}) model the block-wise relationships by collecting statistics of sampled particles moving between consecutive blocks, resulting into a directed graph, as shown in Figure~\ref{fig:intro} (b). This graph encodes the \emph{first-order Markov dependency}, meaning that the distribution of the next block to visit only depends on the current block where a particle resides. For example, all particles in block $(2,2)$ will move to $(2,3)$ (i.e, $p((2,2) \rightarrow (2,3))=1.0$), and a particle in $(2,3)$ has equal chance to visit either $(2,4)$ or $(1,3)$ (i.e., $p((2,3)\rightarrow(2,4))=0.5$ and $p((2,3)\rightarrow(1,3))=0.5$). However, this assumes that all particles in a block share the same distribution, which may be inaccurate. In Figure~\ref{fig:intro} (a), we can see that the blue particles in $(2,3)$ will move to $(2,4)$ and most of the red ones will move to $(1,3)$.

The \emph{higher-order Markov dependency} encodes the transition from a sequence of possible events to the next event. In our scenario, this means that the probability of the next block to visit not only depends on the current block where a particle resides, but also a series of blocks this particle has visited before. For example, using higher-order dependencies, the red particles in $(2,3)$ will be denoted as $(2,3)|(2,2).(2,1)$, meaning particles currently in $(2,3)$ given that they come from $(2,2)$ and $(2,1)$. For these particles, the probability to visit $(2,4)$ becomes $p((2,3)|(2,2).(2,1) \rightarrow (2,4))=0.2$ and the probability to visit $(1,3)$ becomes $p((2,3)|(2,2).(2,1) \rightarrow (1,3))=0.8$. Similarly, for the blue particles in $(2,3)$, their transition becomes $p((2,3)|(2,2).(3,2)\rightarrow(2,4))=1.0$. Note that patterns of the blue and red streamlines become more distinguishable using this representation. Therefore, \emph{the higher-order Markov dependencies provide a clearer picture of particle movements between blocks}. To avoid confusion, we refer to the evidence sequence of events as a \emph{higher-order state} (e.g., $(2,3)|(2,2).(3,2)\rightarrow(2,4)$ is a third-order dependency and $(2,3)|(2,2).(3,2)$ is a third-order state).

{\bf Higher-order network}.
The higher-order dependencies only describe the local transition patterns. To further capture the global structure of a flow field, transitions among higher-order states must be incorporated. The higher-order network (HON) is a directed graph, whose nodes are higher-order states and edges encode transition probabilities between nodes. Figure~\ref{fig:intro} (c) illustrates such an example. For particles in block $(3,2)$, as no preceding block is given, these particles start from a first-order state $(3,2)$. After moving to $(2,2)$, the particles have a second-order state $(2,2)|(3,2)$. An edge is added to the graph to connect the two states $(3,2)$ and $(2,2)|(3,2)$. Comparing to the corresponding FON (Figure~\ref{fig:intro} (b)), the higher-order network splits the first-order state $(2,2)$ into two second-order states $(2,2)|(3,2)$ and $(2,2)|(2,1)$. These second-order states ``remember" the history of particles to better distinguish different flow behaviors. Similarly, the node $(2,3)$ is split into two nodes $(2,3)|(2,2).(3,2)$ and $(2,3)|(2,2).(2,1)$, and the edge $(2,2)\rightarrow(2,3)$ is split into two respective edges.

{\bf Why do we need higher-order networks in flow analysis?}
An interpretation of the HON is that \emph{the HON implicitly subdivides the blocks in a regular grid along the flow}. \hot{As illustrated by Figure~\ref{fig:intro} (a), the particles in block $(2,2)$ form two groups based on which blocks they have previously visited. This implicitly subdivides block $(2,2)$ into a red and a blue region along the streamlines, corresponding to the flows going to the right and those going downward, respectively. Similar subdivisions of blocks can be observed in real-world data sets, which will be discussed in Section~\ref{sec:optimize}.}

\emph{The subdivision behavior of the HON provides a finer-level description of the flow field, where the global structure can be better studied}. In Figure~\ref{fig:intro} (c), we can easily identify two communities in the HON, where there is only a weak transition between the two communities. However, in Figure~\ref{fig:intro} (b), this structure is not available in the FON, as the two nodes $(2,2)$ and $(2,3)$ mix different movement patterns. Although multi-resolution techniques, such as octree, may be used to further subdivide blocks with complicated patterns, these techniques partition the blocks regularly along the axes, which may require a much higher number of small blocks to approximate the irregular flow boundaries.

Additionally, \emph{as a directed graph, the HON allows all existing graph analysis approaches to be applied directly, with better accuracy}. For example, the random walk can be used to approximate the particle movement on the graph. In Figure~\ref{fig:intro}, a particle starting from $(3,2)$ will reach either $(2,4)$ or $(1,3)$ with $50\%$ of chance using the FON, while this particle will reach $(2,4)$ for sure using the HON, which is more accurate. For another example, community detection algorithms can be used to identify the two different streamline bundles using the HON but not using the FON.

\begin{figure}[t]
\begin{center}
\includegraphics[width=0.8\linewidth]{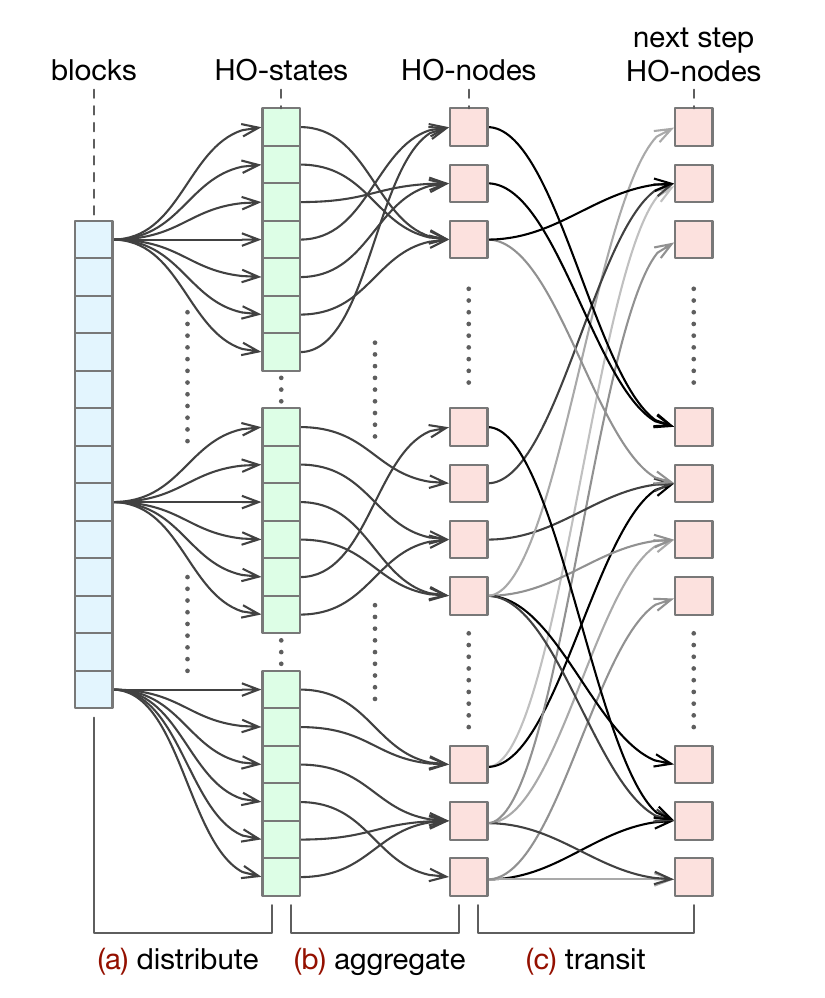}
\end{center}
\caption{An illustration of our formulation of the HON construction from data blocks. This formulation can be considered as three linear layers: (a) distributing the particles from data blocks to HO-states; (b) aggregating HO-states into HO-nodes; and, (c) transiting from current HO-nodes to next-step HO-nodes.}
\label{fig:formulation}
\end{figure}

\section{Our Construction Algorithm}
\label{sec:algorithm}

We formulate the HON construction process as a network optimization problem. The network contains three linear layers to connect the particle distributions in data blocks, higher-order states (HO-states), and higher-order nodes (HO-nodes). 
The HO-nodes are formed by aggregating similar HO-states to reduce the size of a fixed-order network~\cite{rosvall2014memory}. 
Specifically, these three layers represent \emph{the distribution of particles from blocks to HO-states}, \emph{the aggregation of HO-states into HO-nodes}, and \emph{the transitions between HO-nodes}, respectively, as shown in Figure~\ref{fig:formulation}. By connecting the three layers, our approach allows different steps in HON construction to be optimized in a unified framework. In this section, we will start by introducing existing HON construction algorithms and discussing their limitations. Then, we will introduce our formulation and show that the existing approaches can be generalized by our formulation. Finally, we will cover the basic components of the optimization in detail.

{\bf Notations}. For clarity, we distinguish two similar concepts as follows. The \emph{higher-order state} is the finest level of elements in the HON. A HO-states represents a sequence of blocks visited by a particle consecutively (e.g., $(2,3)|(2,2).(3.2)$). The \emph{higher-order node} is a node in the HON, representing a group of HO-states exhibiting similar transition behavior. By aggregating HO-states into HO-nodes, we compress the size of the HON to achieve more efficient computation and more effective visualization. 

\begin{figure}[t]
\begin{center}
$\begin{array}{c@{\hspace{0.05in}}c}
\includegraphics[height=1.1in]{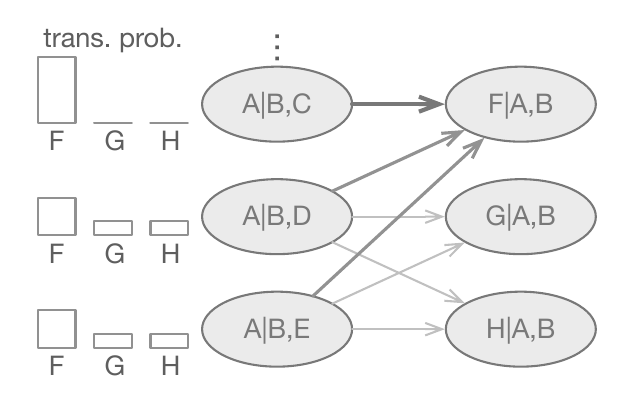}&
\includegraphics[height=1.1in]{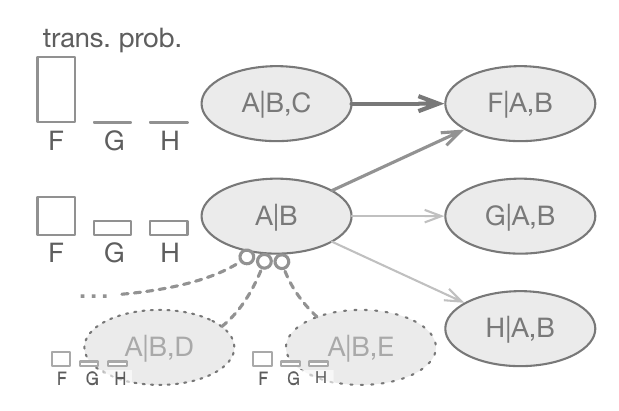}\\
\mbox{(a)} & \mbox{(b)}\\
\end{array}$
\end{center}
\caption{Existing higher-order network construction algorithms. (a) shows a fixed-order network~\cite{rosvall2014memory}, where all nodes share a fixed order. (b) shows the corresponding variable-order network~\cite{xu2016representing}, where similar higher-order nodes are merged into a lower-order node.}
\label{fig:fixed-vary}
\end{figure}

\subsection{Existing HON construction approaches}
Higher-order networks are generally categorized into two types: the fixed-order~\cite{rosvall2014memory} and variable-order networks~\cite{xu2016representing}.
Figure~\ref{fig:fixed-vary} (a) shows a portion of a fixed-order network. In this network, each node is a third-order state, and each edge represents the transition probability between two corresponding states. The transition probabilities are initially recorded between a higher-order state and an individual event, and later expanded into transitions between two states of the same order. Note that the higher-order state provides enough previous history to expand an individual event to a state of the same order or one order higher. For example, the transition $A|B.C\rightarrow F$ is expanded into $A|B.C\rightarrow F|A.B$. The fixed-order network provides the complete dependency information up to a fixed order, but the number of nodes may increase exponentially with the order.

Figure~\ref{fig:fixed-vary} (b) illustrates a variable-order network, which contains nodes of various orders to reduce the unnecessary higher-order nodes. A higher-order node is included in the variable-order network only if its transition behavior differs from the corresponding lower-order node. \hot{The behavior difference is measured between the transition probability distribution of a higher-order node and that of the corresponding lower-order node. The Kullback-Leibler divergence (KLD)~\cite{kullback1951information} is used to calculate the difference between two distributions based on information theory.} For example, in Figure~\ref{fig:fixed-vary} (b), the two third-order nodes $A|B.D$ and $A|B.E$ have similar distributions to a second-order node $A|B$, while the third-order node $A|B.C$ has a different distribution. In this case, $A|B.D$ and $A|B.E$ are considered to be redundant, as they provide no additional information than $A|B$. Therefore, the variable-order network will include $A|B$  to represent both $A|B.D$ and $A|B.E$, and include $A|B.C$ to preserve its unique transition behavior.

However, the variable-order network reduces the size from the fixed-order network based on handcrafted rules, which does not guarantee optimal performance for three reasons.
First, the construction process only considers the similarities between states of different orders but does not take into account the similarities between states of the same order. As a result, it fails to combine similar states corresponding to different lower-order nodes.
Second, the transition probability distribution of the lower-order state may not be the most appropriate one to represent all states reduced to it. For example, in Figure~\ref{fig:fixed-vary}, $A|B$ represents $A|B.D$ and $A|B.E$, but its distribution includes the transitions through $A|B.C$ as well, which may lead to inaccurate probabilities.
Third, both the variable-order and fixed-order networks estimate the transition probabilities by counting transitions over the entire history, but ignore the differences residing in the transition patterns due to the change of particles' spatial distribution.

\begin{figure}[t]
\begin{center}
\includegraphics[width=0.9\linewidth]{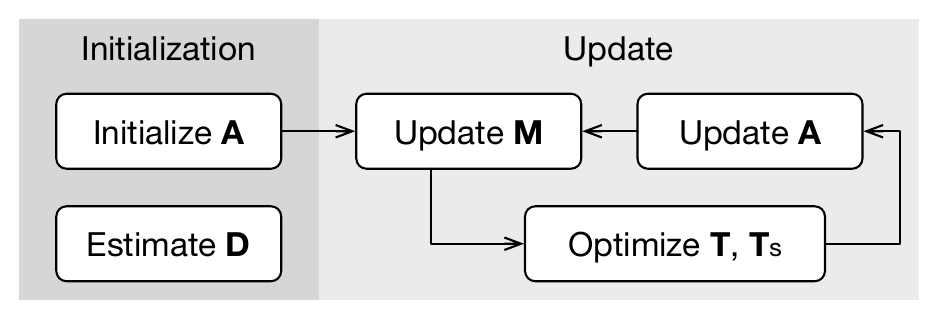}
\end{center}
\caption{The optimization routine for our FlowHON construction. The initialization stage initializes $\matA$ and $\matD$. The update stage iteratively optimizes $\matT$ based on the current $\matA$, and then updates $\matA$ based on $\matT$, so that the interaction between $\matA$ and $\matT$ is incorporated. The optimization is guided by a loss function based on $\matD$, $\matA$, and $\matT$. The loss function is evaluated in the optimization and the validation, but they are hidden in this figure for visual compactness.}
\label{fig:optimize}
\end{figure}

\subsection{Problem formulation}
\label{sec:formulate}
We consider two essential problems in the HON construction jointly: how to generate nodes in the HON and how to estimate the transition probabilities between nodes.
Toward this end, we formulate the HON construction process as an optimization problem for a three-layer network. Each layer is a linear transformation represented by a weight matrix, which is elaborated as follows.

{\bf Distribution}. The first layer is the distribution layer, represented by the \emph{distribution matrix} $\matD$. It distributes particles in each data block to corresponding HO-states. Formally, let two column vectors $\vecB$ and $\vecS$ represent the numbers of particles in blocks and HO-states, where elements $b_i$ and $s_j$ are the number of particles in the $i$-th block and the $j$-th HO-state, respectively. The distribution matrix $\matD$ estimates $\vecS$ by $\vecS = \matD \cdot \vecB$. Each element $D_{j,i}$ in the distribution matrix can be seen as the fraction of particles in block $b_i$ corresponding to HO-state $s_j$ (i.e., $s_j=b_i\cdot D_{j,i}$). The matrix $\matD$ should fulfill two requirements: each column should be a partition of unity, meaning that the numbers of particles in HO-states should sum up to the number in the corresponding block; and, each row should be a one-hot vector, meaning that the particles of a HO-state should only come from the corresponding block.

{\bf Aggregation}. The second layer is the aggregation layer, represented by the \emph{aggregation matrix} $\matA$. This layer aggregates the HO-states into HO-nodes to reduce the size of the network. Similarly, the matrix $\matA$ estimates a column vector $\vecN$ representing the number of particles in each HO-node by computing $\vecN=\matA \cdot \vecS$. The matrix $\matA$ should only contain binary values. Specifically, the element $\matA_{i,j}$ equals one if the $j$-th HO-state is assigned to the $i$-th HO-node. Note that only the HO-states sharing the same current block (meaning that the particles are residing in the same block) could be aggregated into the same HO-node. Additionally, each column in $\matA$ should be a one-hot vector, meaning that a HO-state should be assigned to a single HO-node. 

{\bf Transition}. The third layer is the transition layer, represented by the \emph{probability transition matrix} $\matT$. This layer approximates the movement of particles between HO-nodes. Formally, given a column vector $\vecNAt{t}$ representing the number of particles in each HO-node at time $t$, the matrix $\matT$ estimates the numbers at time $t+1$ as $\vecNAt{t+1}=\matT \cdot \vecNAt{t}$. Specifically, element $\matT_{i,j}$ represents the transition probability from the $j$-th HO-node to the $i$-th HO-node. Note that the transition is not always valid between any two HO-states. In our scenario, we enforce two constraints to avoid the violation of physical rules (e.g., a transition should not appear between two spatially disjoint blocks) and the violation of the semantic meaning of HO-states (e.g., a valid transition from state $A|B.C$ should move to states in the form of ``$*|A.B$"). Accordingly, transitions between HO-nodes should be constrained as well. In our implementation, this constraint is enforced by a \emph{mask matrix} $\matM$, where the element $\matM_{i,j}$ indicates whether the transition between the $i$-th and $j$-th HO-nodes is valid.

{\bf Connection to existing approaches}. This three-layer construction model generalizes both the fixed-order~\cite{rosvall2014memory} and variable-order~\cite{xu2016representing} networks.
The fixed-order network represents each HO-state of a fixed-order as a node, and the probability of starting from a certain node is given by the sampled data. This can be seen as using our approximate assignment (which will be discussed later) to produce the distribution matrix $\matD$ and using an identity matrix as the aggregation matrix $\matA$. 
The variable-order network always starts from a first-order state and generates the history in random walks on the network. This can be seen as a distribution matrix $\matD$, where $D_{i,j}=1$ implies that the $i$-th HO-state is a first-order one.
The variable-order network uses HO-states of different orders as nodes. This can be seen as using a handcrafted rule to produce the aggregation matrix $\matA$, which aggregates HO-states of higher orders into the corresponding lower-order ones.
Both the fixed-order and the variable-order networks estimate the transition probabilities using the sampled data, which can be seen as the computation of $\matT$.
Therefore, these two network construction algorithms can be special solutions to our problem. Ideally, our approach should produce the optimal performance given the same network size.

\begin{figure*}[t]
\begin{center}
 $\begin{array}{c@{\hspace{0.1in}}c@{\hspace{0.1in}}c}
\includegraphics[width=0.3\linewidth]{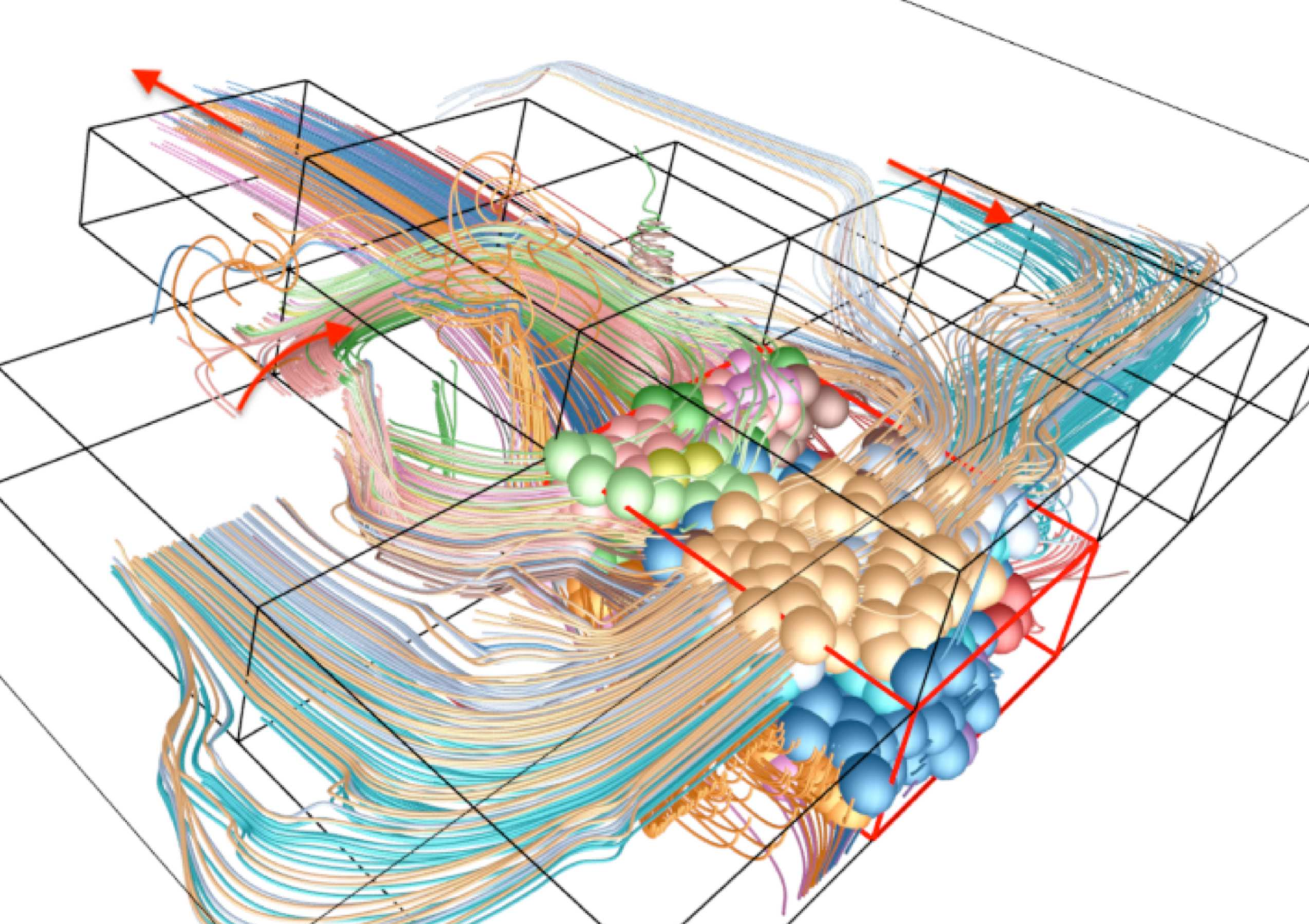}&
\includegraphics[width=0.3\linewidth]{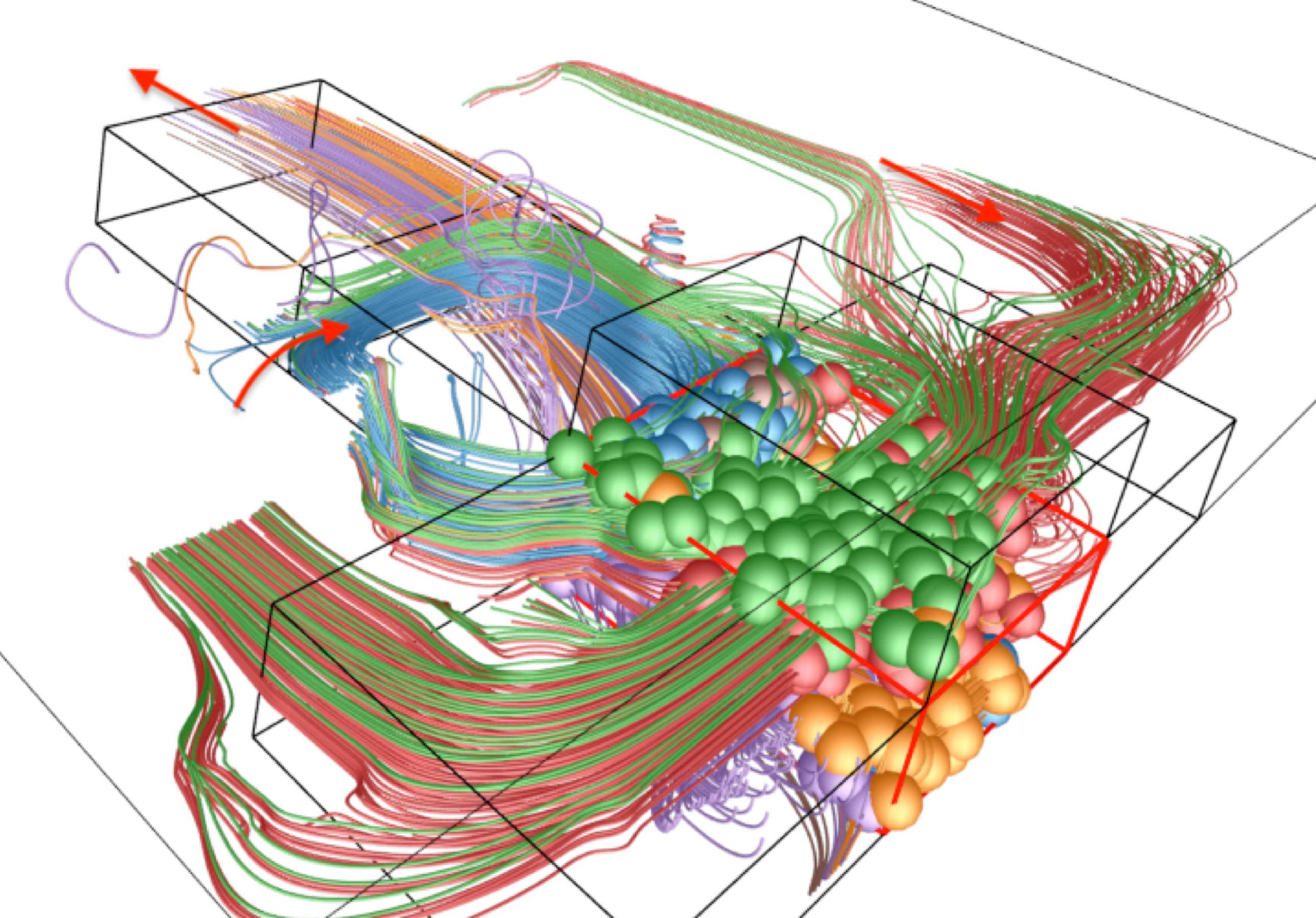}&
\includegraphics[width=0.3\linewidth]{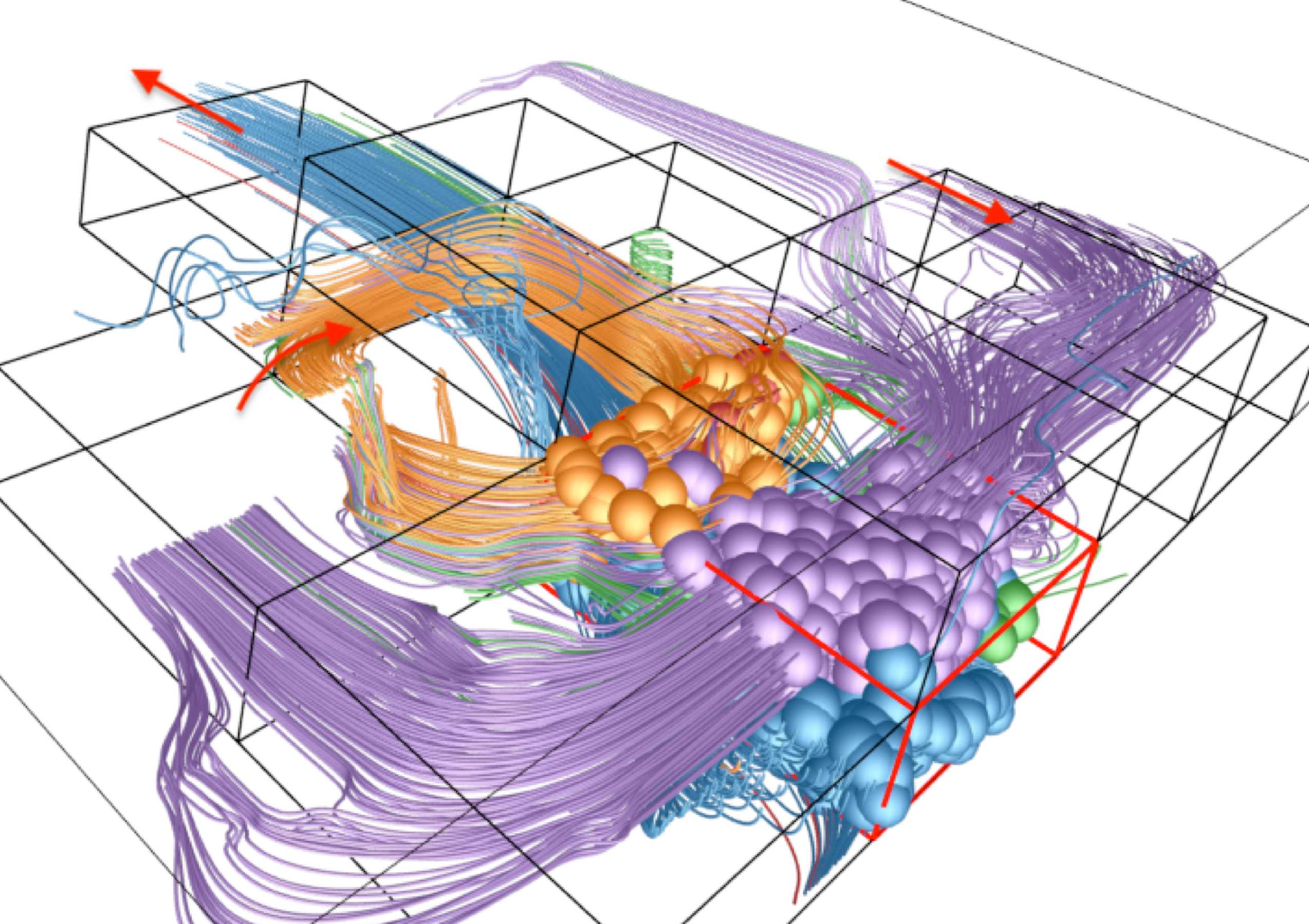}\\
\mbox{(a)} & \mbox{(b)} & \mbox{(c)}
\end{array}$
\end{center}
\caption{Visualization of higher-order nodes produced by different schemes to group higher-order states. (a) shows the individual higher-order states. (b) shows the higher-order nodes generated by the variable-order network~\cite{xu2016representing}, or the semantic scheme. (c) shows the higher-order nodes generated by hierarchical clustering. The red boxes indicate the current blocks, and the red arrows indicate the flow directions. The spheres represent sampled particles, and their colors encode higher-order nodes.}
\label{fig:subdiv}
\end{figure*}

\subsection{Optimization}
\label{sec:optimize}
The optimization routine for our FlowHON is illustrated in Figure~\ref{fig:optimize}. This routine aims at identifying the optimal aggregation of HO-states and transitions between nodes to best mimic the transition statistics in the sampled data. The optimization is performed in two stages: the initialization stage and the update stage. In the initialization stage, the initial values of the aggregation matrix $\matA$ and the distribution matrix $\matD$ are computed. These two matrices provide the initial HO-states and HO-nodes. In the update stage, a repeated procedure is performed to update the transition matrix $\matT$ and the aggregation matrix $\matA$ iteratively. In each iteration, we first optimize the transition matrix $\matT$ based on the current HO-nodes ($\matA$), and then update the HO-nodes ($\matA$) based on the optimized transitions. \hot{The source code is provided at GitHub with the following link: \url{https://github.com/NanChanNN/FlowHON}}. In this section, we will elaborate on each component in the routine, starting from the loss function of the optimization. 

{\bf Loss function}. The loss function evaluates the error of estimating the particle distribution over blocks using the model. We use the estimation error to guide the optimization, as estimation of particle movement is an essential task in graph-based flow visualization, on top of which many other applications are built. Given the initial numbers of particles in the blocks $\vecBAt{0}$ at step $0$, we use the matrices $\matD$, $\matA$, $\matT$ to estimate the numbers up to a predefined step $k$ (i.e., $\estVecBAt{1}, \cdots, \estVecBAt{k}$). The estimated numbers are compared to the actual numbers from tracing (i.e., $\vecBAt{1}, \cdots, \vecBAt{k}$) using KL-Divergence (KLD). Each component in a vector $\vecB$ or $\estVecB$ should be divided by the total number of particles to convert $\vecB$ to a distribution for the KLD computation. As this division only scales the loss by a constant factor, we use the vector $\vecB$ in our loss function for simplicity:

\begin{align}
L = \sum_{t=1}^k \kld (\vecBAt{t} || \estVecBAt{t}) = \sum_{t=1}^k \sum_{i} b_i^{(t)} \cdot \log \frac{b_i^{(t)}}{\estB{i}{t} + \epsilon},
\label{eq:loss}
\end{align}

\noindent where $b_i^{(t)}$ and $\estB{i}{t}$ are the actual number and the estimated number of particles in the $i$-th block at step $t$, respectively, and $\epsilon$ is a small constant to avoid division by zero. 

The vector $\estB{i}{t}$ is estimated using the matrices $\matD$, $\matA$, and $\matT$. Given the initial particle numbers in blocks $\vecBAt{0}$, the particle numbers in HO-states and in HO-nodes can be derived by $\vecSAt{0} =\matD \cdot \vecBAt{0}$ and $\vecNAt{0}=\matA \cdot \vecSAt{0} = \matA \cdot \matD \cdot \vecBAt{0}$, respectively. Then, the movement of particles between HO-nodes can be approximated using the transition matrix $\matT$, and the particle numbers in HO-nodes at step $t$ can be estimated as $\estVecNAt{t} = \matT^t \cdot \vecNAt{0}$. Note that we use parentheses in superscripts to distinguish time steps (e.g., $^{(t)}$) from the power exponent (e.g., $^t$). At each step, we can aggregate the particle numbers in HO-nodes to the numbers in blocks using $\estVecBAt{t} = \nonzero (\matD^T \cdot \matA^T) \cdot \estVecNAt{t}$, where $\matD^T$ and $\matA^T$ are the transposed matrices of $\matD$ and $\matA$, respectively, and $\nonzero(\cdot)$ is an element-wise function that sets the non-zero elements in a matrix to be $1$. In summary, the estimated particle numbers in blocks $\estVecBAt{t}$ at step $t$ is given by:

\begin{align}
\estVecBAt{t} = \nonzero(\matD^T \cdot \matA^T) \cdot \matT^t \cdot (\matA \cdot \matD \cdot \vecBAt{0}).
\label{eq:estB}
\end{align}

\noindent Therefore, we can see that the loss function only relies on $\matD$, $\matA$, and $\matT$, which are the parameters to be optimized in our framework.

{\bf Estimate $\matD$}. The construction of matrix $\matD$ is part of the initialization step. \hot{The matrix $\matD$ transforms the number of particles in each block to the number of particles in each HO-state. We provide two strategies for the initialization: \emph{exact assignment} and \emph{approximate assignment}. The exact assignment uses backward tracing of particles to determine the exact visiting history. Given the visiting history, we could assign a particle to the corresponding HO-state precisely. While accurate, exact assignment suffers from the heavy computation of backward tracing for each particle. During experiments, we observe that the distribution from the blocks to the HO-states is quite stable at the beginning, which means that we could assign particles on a block to HO-states based on the statistics of previous samples. Motivated by this, we put forward the second strategy, namely approximate assignment.
This strategy uses statistics from sampled particles to determine the fraction of particles corresponding to each HO-state in a block.
The distribution matrix only reflects the statistics of HO-states when the visiting history is not given. As the distribution from the blocks to the HO-states is not involved in the later transitions, this matrix will remain constant after initialization.}

{\bf Initialize $\matA$ (node generation)}. \hot{The aggregation matrix $\matA$ has more columns than rows, which reduces the amount of HO-nodes in the resulted network. This matrix can be derived from any method that groups similar HO-states into HO-nodes. For example, we may use the variable-order network~\cite{xu2016representing} to produce the aggregation matrix by considering the lower-order node to be a group of higher-order states. We refer to this node generation scheme as the \emph{semantic} scheme, as the HO-states are grouped based on their semantic meanings. However, this may fail to group HO-states with similar behavior but different previous blocks. In our implementation, we apply the \emph{hierarchical clustering} to group similar HO-states in each block.} The hierarchical clustering starts with clusters of individual HO-states and merges the two closest clusters in each iteration if their distance is smaller than a predefined threshold. The distance between two HO-states is defined as the Euclidean distance between their corresponding transition probability distributions, and the distance between two clusters is defined as the weighted average between HO-states in the two clusters $\mathbf{c}_i$ and $\mathbf{c}_j$:

\begin{align}
d(\mathbf{c}_i, \mathbf{c}_j) = \frac{\sum_{u \in \mathbf{c}_i, v \in \mathbf{c}_j} w_u \cdot w_v \cdot \euclid (p_u, p_v)}{\sum_{u \in \mathbf{c}_i } w_u \cdot \sum_{v \in \mathbf{c}_j} w_v},
\end{align}

\noindent where $w_u$ and $w_v$ are the numbers of transitions related to HO-states $u$ and $v$, respectively, and $p_u$ and $p_v$ are the transition probability distributions from HO-states $u$ and $v$ to the next blocks, respectively. Note that we use transition probabilities from HO-states to blocks instead of between HO-states to avoid overfitting of the clustering. 

\hot{Figure~\ref{fig:subdiv} compares the individual HO-states, the HO-nodes generated by the semantic scheme, and the HO-nodes generated by this hierarchical clustering. The spheres represent particles in the block highlighted in red, and the colors of spheres indicate the respective higher-order nodes of the particles. The streamline segments corresponding to the particles are displayed in the same respective color. We can see that particles of the same color reside in similar regions inside the block. These regions partition the block, where streamlines of the same region have similar behaviors. The individual HO-states represent flow patterns at the finest level, as shown in Figure~\ref{fig:subdiv} (a). The HO-nodes generated by the semantic scheme cannot combine patterns from different blocks. Therefore, they may fail in grouping similar patterns together. In Figure~\ref{fig:subdiv} (b), we can see some of the green particles share a similar transition pattern as the red ones, while others share a pattern with the blue ones. In contrast, the HO-nodes generated from hierarchical clustering are able to provide a more concise summarization of transition patterns, where particles of the same color often heading to the same block, as shown in Figure~\ref{fig:subdiv} (c).}

\begin{table*}[th]
\centering
\caption{\hot{The data sets used in the experiment. Block dimensions show the numbers of blocks along each axis. ``init" and ``train" show the node initialization and edge training time in seconds, respectively. ``ours" shows the construction time of the third-order {\ours}.}}

\begin{tabular}{l|c|c|r|r|r|r|r|r|r|r}
\hline\hline
                     & data                                                        & block                  & \multicolumn{2}{c|}{FON} & \multicolumn{2}{c|}{VAR} & \multicolumn{2}{c|}{Fixed} & \multicolumn{2}{c}{ours} \\ \cline{4-11}
data set             & dimension                                                   & dimension              & init       & train      & init         & train         & init       & train        & init       & train       \\ \hline
ABC                  & $64 \times 64 \times   64$    & $6 \times 6 \times 6$  & 0.38       & 2.48       & 0.55         & 31.98         & 0.46       & 223.39       & 3.00       & 51.62       \\
\bernard              & $128 \times 32 \times   64$   & $8 \times 4 \times 4$  & 1.50       & 1.30       & 2.36         & 30.11         & 1.21       & 242.54       & 9.35       & 124.90      \\
combustion           & $506 \times 400 \times 100$ & $10 \times 8 \times 2$ & 0.39       & 1.56       & 0.57         & 20.38         & 0.64       & 142.80       & 7.92       & 64.52       \\
computer room        & $417 \times 345 \times 60$  & $6 \times 6 \times 3$  & 0.32       & 1.19       & 0.44         & 16.83         & 0.49       & 99.54        & 4.92       & 54.73       \\
crayfish             & $322 \times 162 \times 119$ & $10 \times 5 \times 4$ & 0.74       & 2.16       & 1.18         & 66.86         & 1.29       & 732.16       & 17.38      & 127.70      \\
electron             & $64 \times 64 \times   64$    & $5 \times 5 \times 5$  & 0.25       & 1.51       & 0.30         & 5.57          & 0.26       & 37.46        & 3.02       & 22.49       \\
five critical points & $51 \times 51 \times   51$    & $4 \times 4 \times 4$  & 0.24       & 0.87       & 0.30         & 3.82          & 0.20       & 21.51        & 1.34       & 22.94       \\
hurricane            & $500 \times 500 \times 100$ & $6 \times 6 \times 2$  & 0.35       & 0.84       & 0.50         & 4.88          & 0.27       & 29.28        & 1.43       & 41.82       \\
solar plume          & $126 \times 126 \times 512$ & $4 \times 4 \times 10$ & 0.61       & 1.66       & 0.93         & 33.33         & 0.88       & 243.57       & 10.76      & 84.42       \\
square cylinder      & $192 \times 64 \times   48$   & $10 \times 3 \times 2$ & 0.39       & 0.85       & 0.52         & 2.96          & 0.14       & 7.67         & 0.40       & 21.87       \\
tornado              & $64 \times 64 \times   64$    & $5 \times 5 \times 5$  & 0.29       & 1.51       & 0.37         & 5.65          & 0.14       & 24.04        & 0.47       & 16.78       \\
two swirls           & $64 \times 64 \times   64$    & $5 \times 5 \times 5$  & 1.08       & 1.49       & 1.70         & 26.90         & 1.02       & 184.57       & 9.52       & 109.83      \\
\hline
average              & / & / & 0.54       & 1.45       & 0.81         & 20.77         & 0.58       & 165.71       & 5.79       & 61.97 \\
\hline\hline
\end{tabular}
\label{tab:data}
\end{table*}

{\bf Update $\matM$}. An element $M_{i,j}$ in the mask matrix $\matM$ records whether the transition between the $i$-th and $j$-th HO-nodes is physically meaningful. Given the aggregation matrix $\matA$, this can be easily done by checking whether the transition between any pair of their HO-states is possible. In practice, we may simply count the number of transitions between two HO-nodes in the sampled data. If that number is zero, then we set the corresponding element in $\matM$ to be zero. This will prevent the transition probability between corresponding HO-nodes from being updated in the transition matrix optimization. \hot{The mask matrix update is involved in the update loop in Figure~\ref{fig:optimize} as it relies on the aggregation matrix $\matA$. Therefore, once $\matA$ is updated, the mask matrix $\matM$ should be updated as well.}

{\bf Optimize $\matT$ (edge optimization)}. The transition matrix $T$ encodes the edges among HO-nodes in a HON. Traditional approaches~\cite{xu2010flow,rosvall2014memory,xu2016representing} count the transitions between the HO-nodes and compute the probability. However, these approaches do not consider the dynamic patterns of transitions, and, therefore, they may overemphasize the transition patterns when a block contains a large number of particles. \hot{To avoid this issue, instead of using the statistics collected from sampled particles, we aim at learning a matrix $\matT$ that will produce the observed particle distributions at each step.} We start with a matrix $\matTAt{0}$ from sampled particles and update the matrix to minimize the loss function $L$ (Equation~\ref{eq:loss}) using the gradient descent algorithm. Formally, the update process is performed iteratively using the following equation:

\begin{align}
\matTAt{i+1} = \matTAt{i} + \alpha \frac{\partial L}{\partial \matTAt{i}} \odot \matM,
\end{align}

\noindent where $\alpha$ is the learning rate and ``$\odot$" denotes element-wise multiplication. For simplicity, we use $\nonzero(\matTAt{0})$ as the mask matrix $\matM$. 

For $\matT$ to be a transition probability matrix, two additional constraints are enforced. First, all elements in $\matT$ should be non-negative. We utilize projected gradient descent~\cite{lin2007projected} to replace all negative values by zero after every weight update. Second, the summation of elements in each column in $\matT$ must equal to one for this column to be a distribution. We implement this by adding a regulation term in the loss function to punish columns whose element summations deviate from one:

\begin{align}
L_T = L + \sum_{j}(\sum_{i}T_{i,j}-1)^2.
\end{align}

\noindent Besides, we normalize the transition matrix at the end of the optimization process to make the summation of each column strictly equal to one.


{\bf Update $\matA$ (node update)}. \hot{Note that an HO-state should be aggregated into the HO-node with the most similar transition behavior. Therefore, when the transition probability ($\matT$) of HO-nodes is updated, the aggregation from HO-states to HO-nodes should be updated accordingly. Because the transitions from the HO-states to HO-nodes are unknown, we use optimization to learn a transition matrix $\matT_s$ using a similar scheme as the $\matT$ update.} Note that each column in $\matT_s$ contains the transition probabilities from a HO-state to all HO-nodes, and each column in $\matT$ contains the probabilities from a HO-node to all HO-nodes. Therefore, for each HO-state, we can identify the most similar HO-node by comparing the corresponding columns in $\matT_s$ and $\matT$ using Euclidean distance. \hot{The aggregation matrix $\matA$ is constructed so that each HO-state is assigned to the most similar HO-node.}

{\bf Termination}. We terminate the optimization routine if one of the following two criteria is fulfilled. First, the node assignment does not change for a predefined number of iterations. This is done by checking whether any value in $\matA$ has changed during the update of $\matA$. Second, the performance of the model does not improve for a predefined number of iterations. We use a validation set of sampled streamlines to evaluate the performance, and the model with the minimum validation loss is stored for later use.

\begin{figure}[t]
\begin{center}
\includegraphics[width=0.9\linewidth]{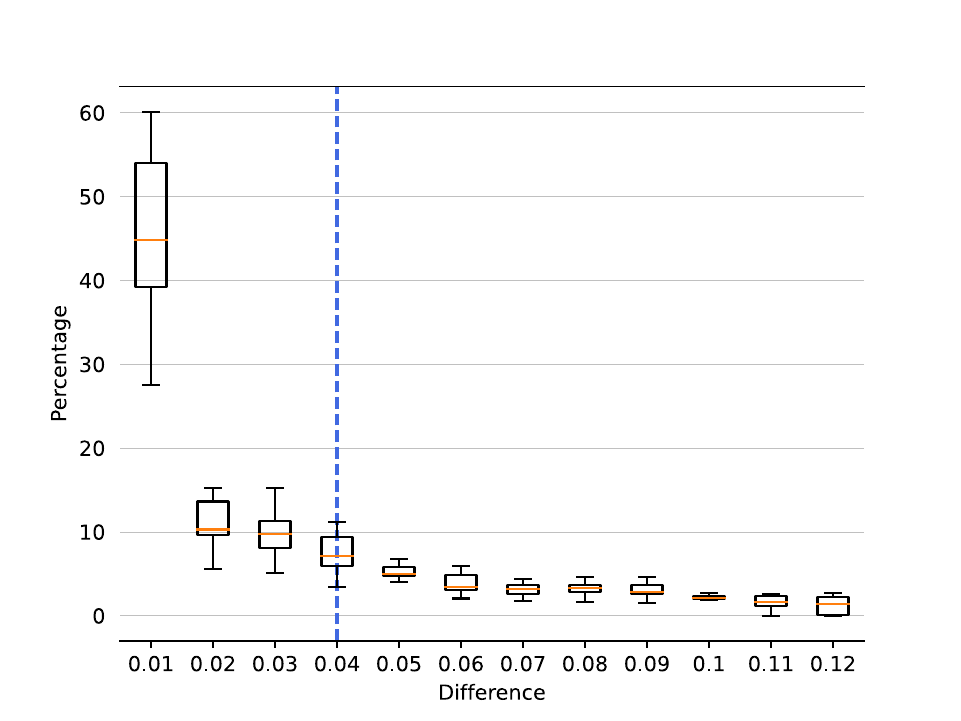}
\end{center}
\caption{The difference distribution of all data sets. The x-axis represents the difference range. The Y-axis represents the percentage of difference values in a range. The box plot shows the $25\%$, $50\%$, and $75\%$ quantiles of the percentage over data sets. The blue dashed line shows the threshold used in our experiment.}
\label{fig:difference_distribution_cruve}
\end{figure}

\section{Results and Discussion}

\subsection{Experiment Configuration}
In the experiment, we compare our approach with the traditional first-order network (\fon) and the variable-order network (\semantic)~\cite{xu2016representing}. \hot{The fixed-order network (\fixed)~\cite{rosvall2014memory} can be considered as a reference, because it represents the finest level of transitions with the largest number of nodes. Both \semantic\ and our FlowHON are approximating the behavior of \fixed\ with fewer nodes.} Since \fon, \semantic, and \fixed\ can be considered as special solutions of $\matD$, $\matA$, and $\matT$, they can all benefit from our transition optimization that updates $\matT$. Therefore, we also compare the optimized versions of \fon, \semantic, and \fixed, denoted as {\fonPlus}, {\semanticPlus}, and {\fixedPlus}, respectively.

\begin{table*}[ht]
\footnotesize
\centering 
\caption{\hot{Average estimation error and network sizes of different network construction approaches. The estimation error is given by KLD using Equation~\ref{eq:loss}. ``VAR" denotes the variable-order network up to third-order, and ``\fixed" denotes the third-order fixed-order network. The smallest estimation error for each data set among the third-order FlowHON, FON, VAR, FON+, and VAR+ is highlighted in bold font.}
} 
\begin{tabular}{l|l|l|l|l|l|l|l|l|l|r|r|r|r|r|r} 
\hline\hline 
                        & \multicolumn{9}{c|}{estimation error}                                                                                & \multicolumn{6}{c}{network size}                                 \\ \cline{2-16} 
                        & \multicolumn{3}{c|}{other techniques} & \multicolumn{3}{c|}{with edge optimization} & \multicolumn{3}{c|}{ours}      & \multicolumn{3}{c|}{other techniques} & \multicolumn{3}{c}{ours} \\ \cline{2-16} 
data set                & FON         & VAR        & Fixed        & FON+       & VAR+               & \fixedPlus      & 2nd   & 3rd            & 4th   & FON        & VAR         & Fixed        & 2nd     & 3rd    & 4th    \\ \hline
ABC                     & 0.029       & 0.022      & 0.013      & 0.029      & 0.019              & 0.012     & 0.019 & \textbf{0.016} & 0.016 & 216        & 978         & 2180       & 469     & 641    & 735    \\
\bernard & 0.057       & 0.062      & 0.055      & 0.022      & 0.017              & 0.014     & 0.017 & \textbf{0.016} & 0.016 & 128        & 968         & 2260       & 361     & 412    & 556    \\
combustion              & 0.024       & 0.021      & 0.017      & 0.022      & \textbf{0.014}     & 0.014     & 0.018 & 0.017          & 0.016 & 141        & 761         & 1879       & 376     & 593    & 798    \\
computer room           & 0.028       & 0.026      & 0.012      & 0.023      & 0.015              & 0.011     & 0.015 & \textbf{0.014} & 0.012 & 107        & 698         & 1631       & 348     & 593    & 747    \\
crayfish                & 0.091       & 0.084      & 0.054      & 0.027      & 0.019              & 0.014     & 0.019 & \textbf{0.019} & 0.019 & 200        & 1486        & 3249       & 562     & 796    & 973    \\
electron                & 0.006       & 0.008      & 0.006      & 0.007      & 0.008              & 0.007     & 0.007 & \textbf{0.006} & 0.006 & 125        & 348         & 1018       & 263     & 330    & 370    \\
5 critical points    & 0.016       & 0.015      & 0.008      & 0.011      & 0.010              & 0.007     & 0.009 & \textbf{0.008} & 0.006 & 64         & 272         & 751        & 214     & 296    & 360    \\
hurricane               & 0.030       & 0.037      & 0.017      & 0.019      & 0.015              & 0.010     & 0.013 & \textbf{0.010} & 0.011 & 72         & 354         & 880        & 247     & 289    & 393    \\
solar plume             & 0.122       & 0.115      & 0.082      & 0.074      & \textbf{0.045}     & 0.029     & 0.053 & 0.051          & 0.051 & 160        & 1009        & 2252       & 410     & 606    & 738    \\
square cylinder         & 0.006       & 0.006      & 0.002      & 0.005      & 0.002              & 0.002     & 0.004 & \textbf{0.002} & 0.001 & 60         & 185         & 384        & 104     & 183    & 227    \\
tornado                 & 0.014       & 0.016      & 0.011      & 0.015      & 0.011              & 0.012     & 0.012 & \textbf{0.011} & 0.010 & 125        & 348         & 754        & 259     & 309    & 357    \\
two swirls              & 0.077       & 0.059      & 0.052      & 0.026      & 0.019              & 0.019     & 0.021 & \textbf{0.019} & 0.016 & 125        & 904         & 2041       & 287     & 404    & 594    \\ \hline
average                 & 0.042       & 0.039      & 0.028      & 0.023      & 0.016              & 0.013     & 0.017 & \textbf{0.016} & 0.015 & 127        & 693         & 1607       & 325     & 454    & 570    \\
\hline\hline
\end{tabular}
\label{table:estimate} 
\end{table*}

{\bf Tasks}. We evaluate the performance of these approaches based on three tasks: namely, \emph{particle density estimation}, \emph{community detection}, and \emph{graph visualization}. The particle density estimation starts from a set of uniformly sampled particles and uses the networks to estimate the numbers of particles over blocks at each step. This task quantitatively evaluates the ability of a network to approximate the transition patterns between blocks. The community detection partitions the flow field into communities and examines the average number of times that a particle moves between two communities. This task quantitatively evaluates the effectiveness of data partition based on different networks. A network is favored if a particle is less likely to move between different communities, because this indicates that the communities are more independent and that less task exchange or data loading is needed in parallel particle tracing. Finally, we visualize the networks using node-link diagrams to evaluate them qualitatively. We explore the graph visualization to see what kind of structural information can be revealed by different networks.

{\bf Data sets and training configuration}. \hot{We use twelve steady flow data sets with different characteristics in the experiment, which are summarized in Table~\ref{tab:data}. The experiment with these data sets are described in Section~\ref{sec:density_estimate}, Section~\ref{sec:community}, and Section~\ref{sec:visualization}. We further include preliminary experiments with two unsteady flow data sets, which will be discussed in Section~\ref{sec:unsteady}}. For the steady fields, we select the block dimension proportional to the data dimension so that each block is roughly a cube. For each data set, we sample 15,000 particles for the network construction (among which 10,000 are used for training and 5,000 are used for validation) and 15,000 for testing. We only use forward tracing to produce streamlines to avoid all streamlines starting from boundary blocks. The streamlines are then converted into a sequence of blocks. A special block ``$-1$" is used to denote the end of streamlines, either due to going out of boundaries or reaching critical points. 
For the node generation, we use a difference threshold of $0.04$ for the hierarchical clustering based on the distribution of differences. As shown in Figure~\ref{fig:difference_distribution_cruve}, the difference value $0.04$ locates at the ``elbow" of the difference distribution, meaning that there is limited room for node compression beyond this point. We train each transition matrix in 100 iterations and set the learning rate to 0.01 with a decay rate of 0.9 every 10 iterations. We terminate the optimization if the matrix $\matA$ does not change for four consecutive iterations. 

\hot{
{\bf Construction time.} The construction of the third-order {\ours} requires 5.79 seconds for node initialization and 61.97 seconds for edge optimization, leading to a total construction time of 67.94 seconds on average, as shown in Table~\ref{tab:data}. The construction time ranges from 17.25 seconds for the tornado data set to 145.08 seconds for the crayfish data set. Compared with the other approaches, it requires the most time in the initialization stage, as hierarchical clustering is used for each node. {\fon} (0.54 seconds) and {\fixed} (0.58 seconds) required the least time to initialize, as they only connect the nodes based on statistics without any further analysis. In terms of edge optimization, {\fon} (1.45 seconds) is the fastest with the smallest number of nodes and {\fixed} (165.7 seconds) is the slowest with the most number of nodes. {\ours} (61.97 seconds) is slower than {\semantic} (20.77 seconds) with similar amount of nodes. {\semantic} does not update the node aggregation, and, therefore, it does not benefit from the iterative optimization scheme. Instead, {\semantic} performs a single iteration of optimization, leading to a faster edge training time.
}


\begin{figure*}[h]
\begin{center}
$\begin{array}{c@{\hspace{0.02in}}c@{\hspace{0.02in}}c@{\hspace{0.02in}}c}
\includegraphics[width=0.24\linewidth]{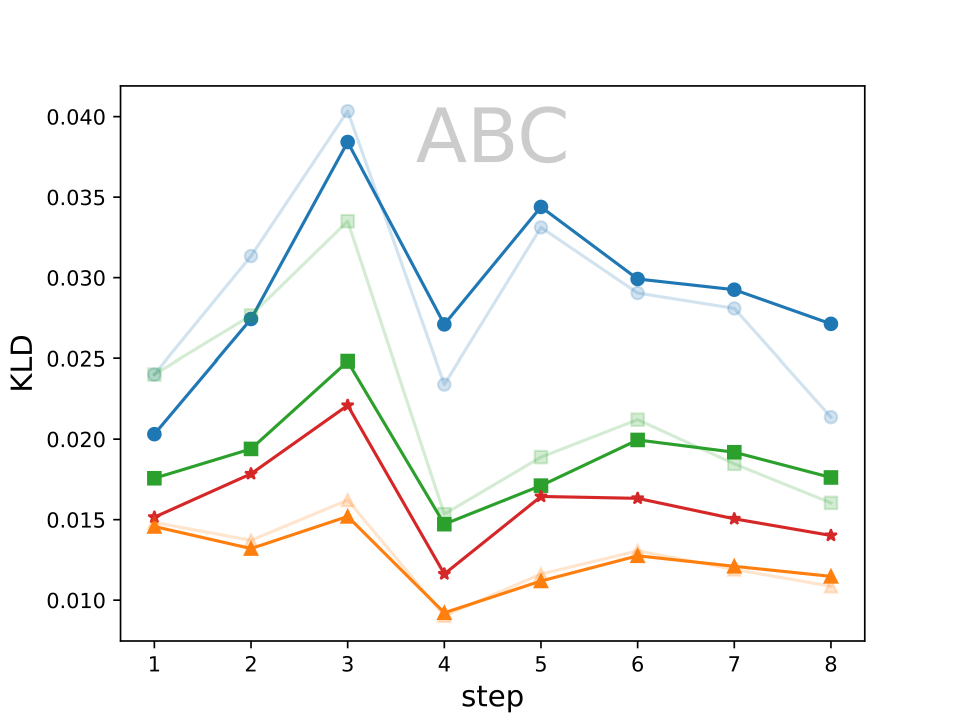}&
\includegraphics[width=0.24\linewidth]{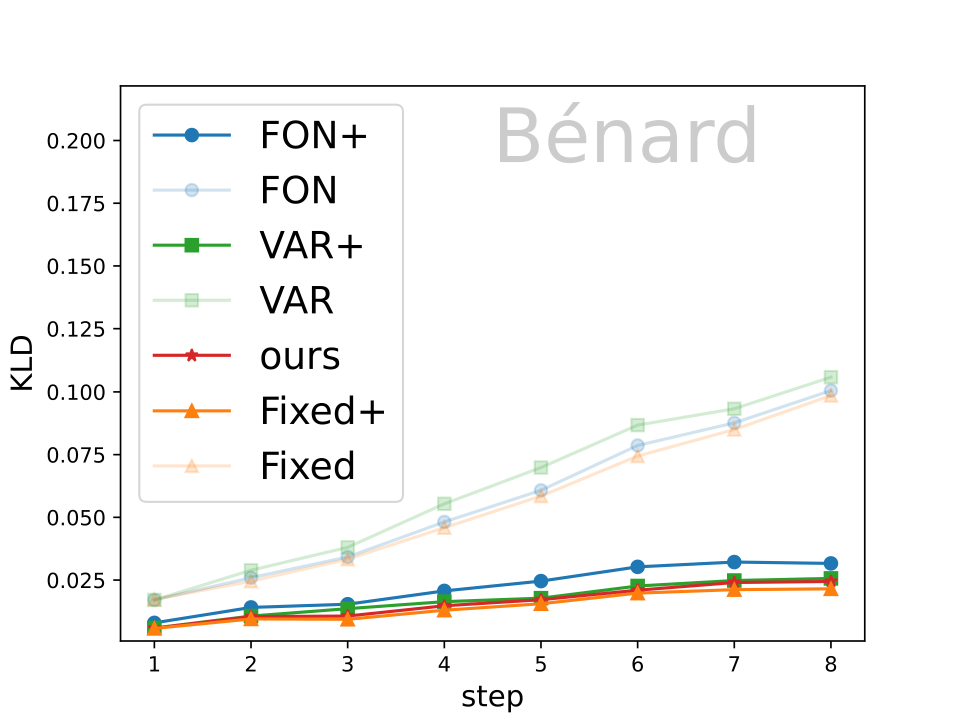}&
\includegraphics[width=0.24\linewidth]{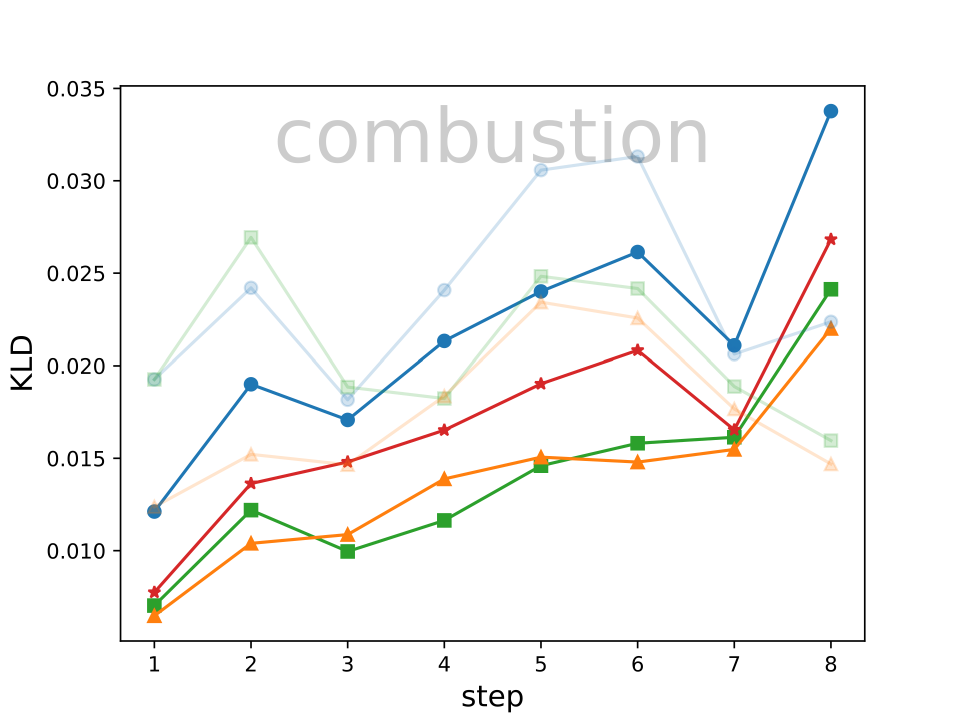}&
\includegraphics[width=0.24\linewidth]{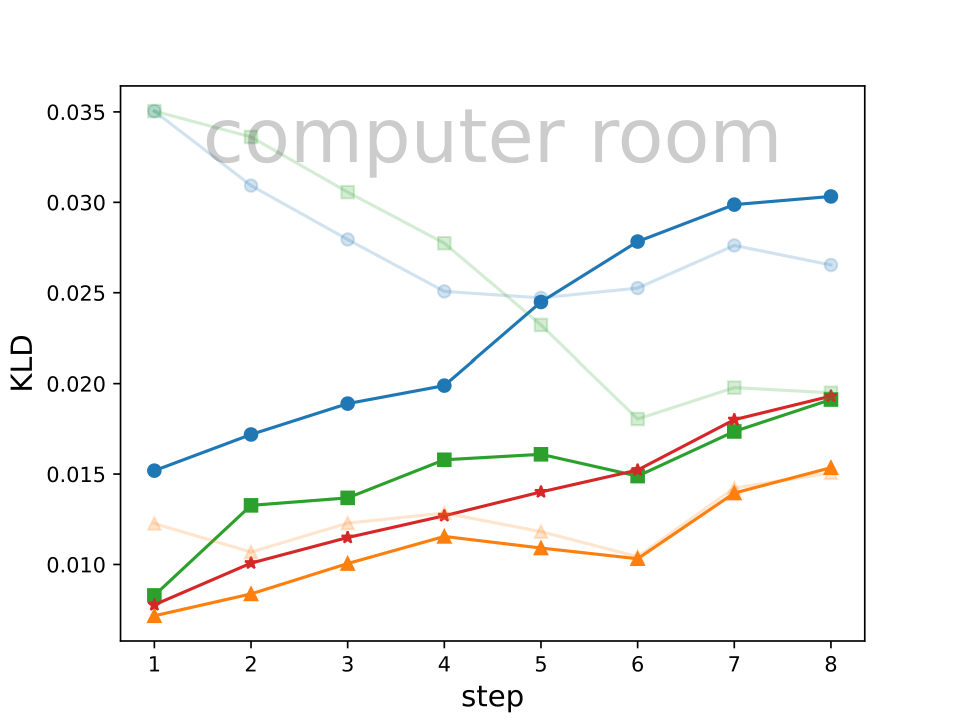}\\
\includegraphics[width=0.24\linewidth]{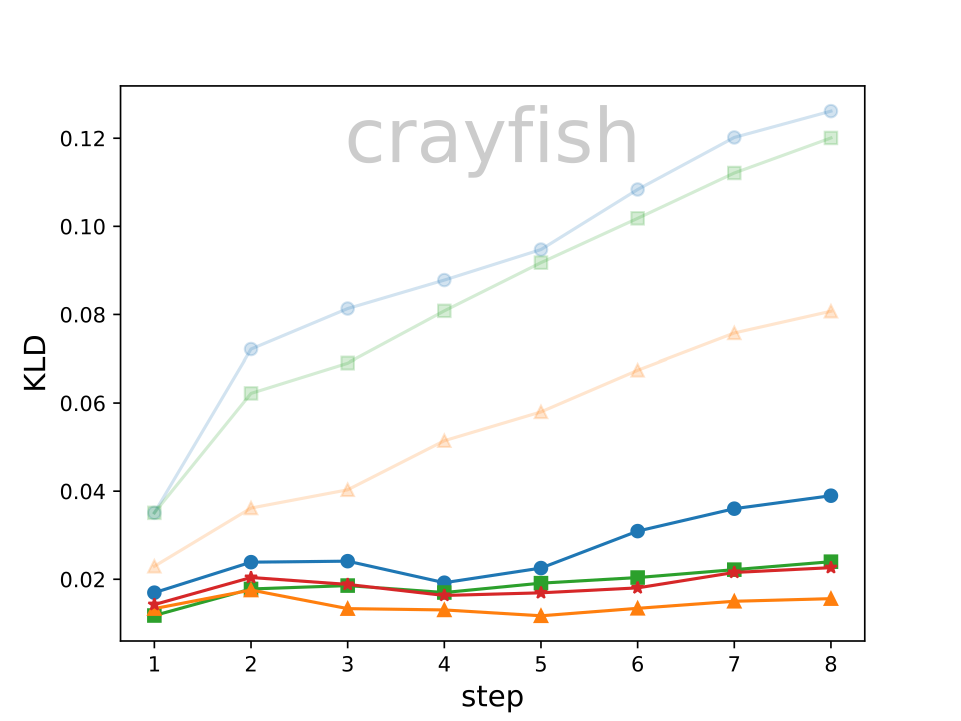}&
\includegraphics[width=0.24\linewidth]{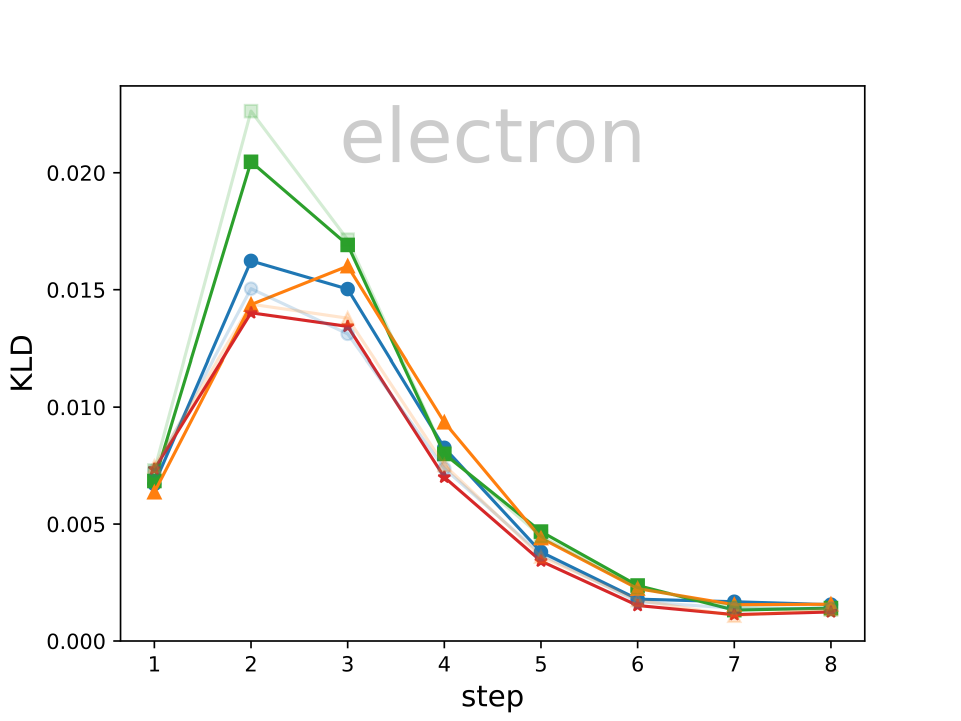}&
\includegraphics[width=0.24\linewidth]{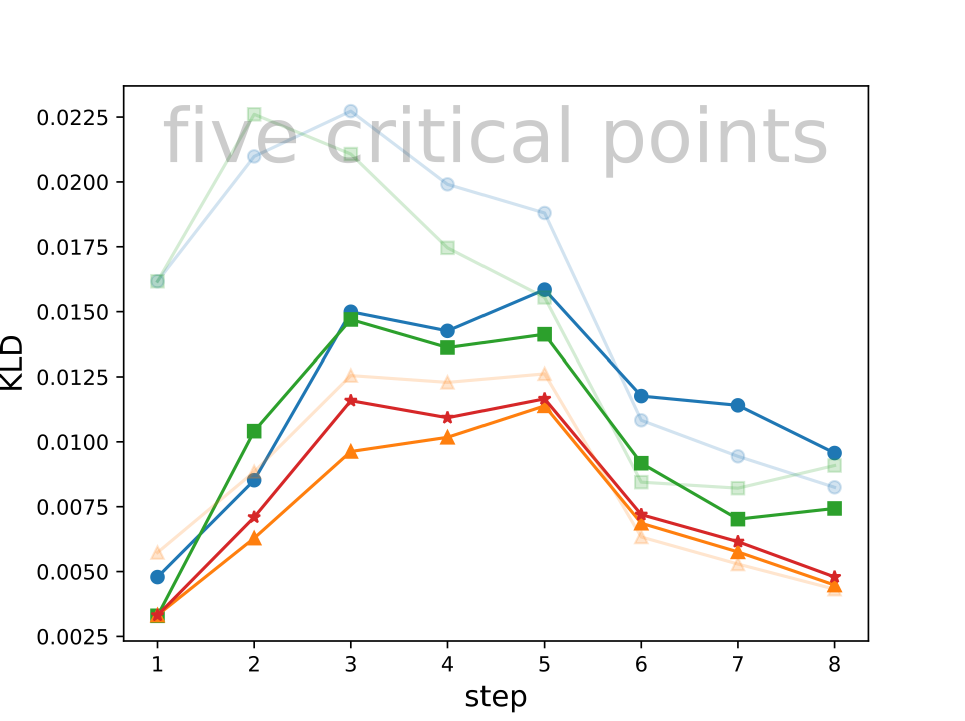}&
\includegraphics[width=0.24\linewidth]{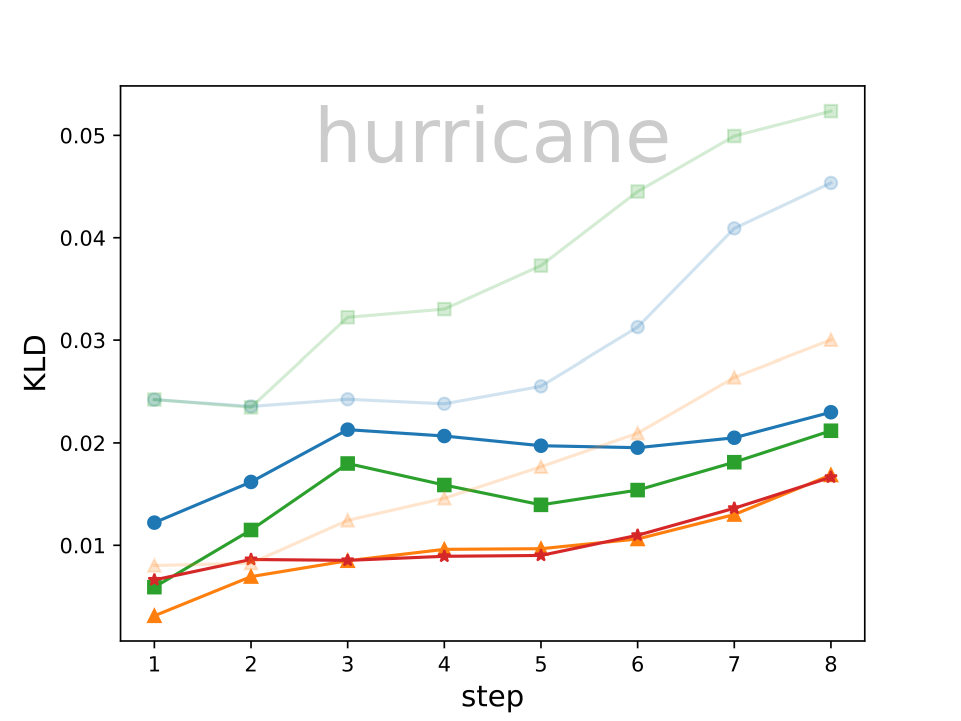}\\
\includegraphics[width=0.24\linewidth]{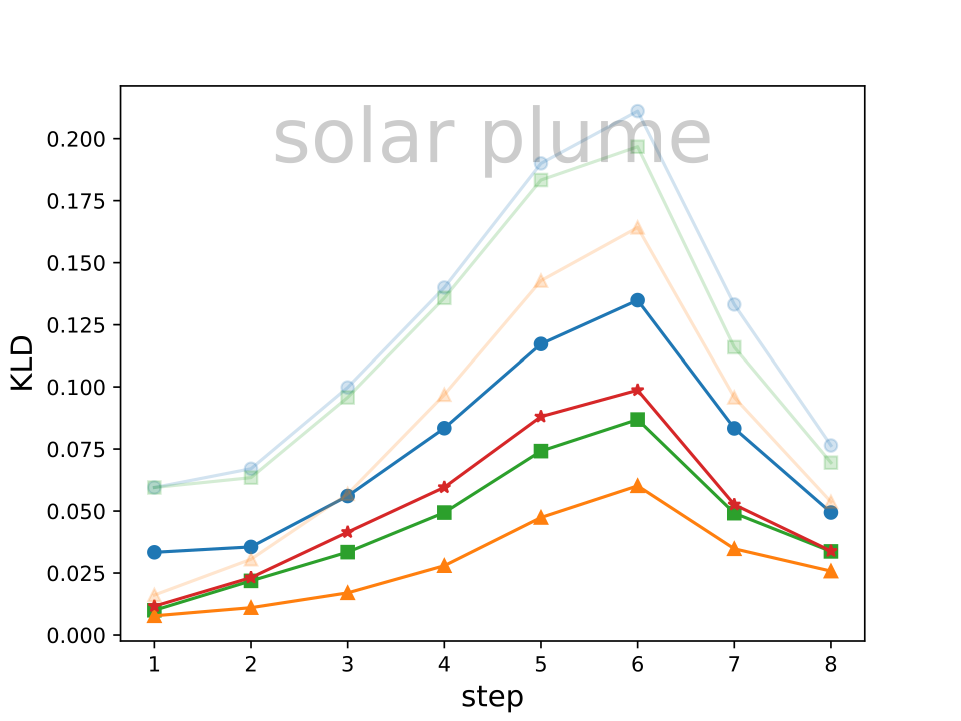}&
\includegraphics[width=0.24\linewidth]{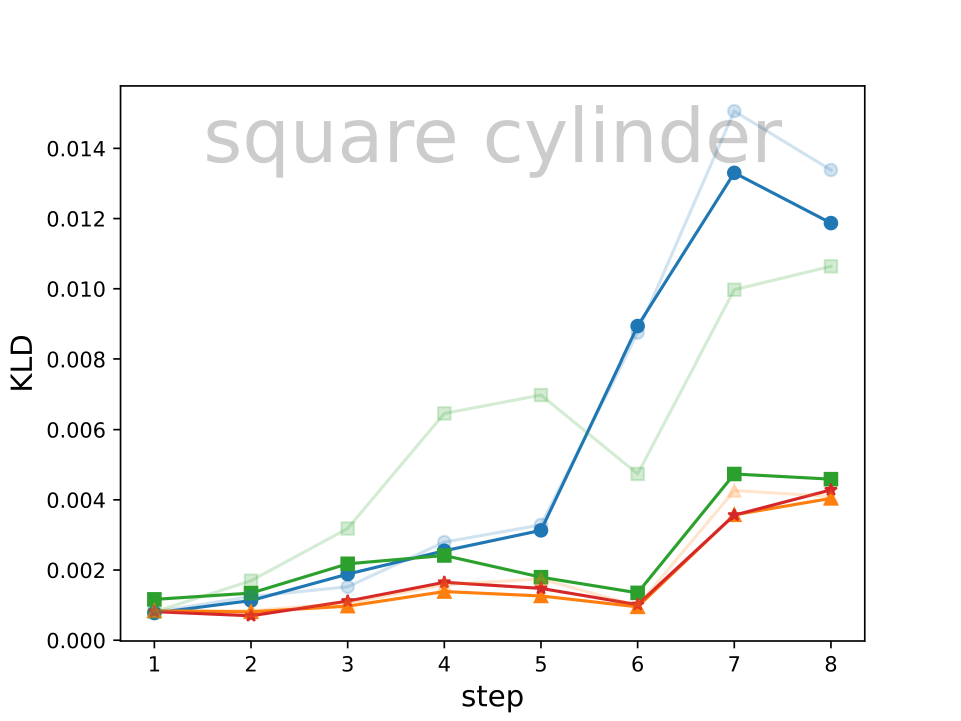}&
\includegraphics[width=0.24\linewidth]{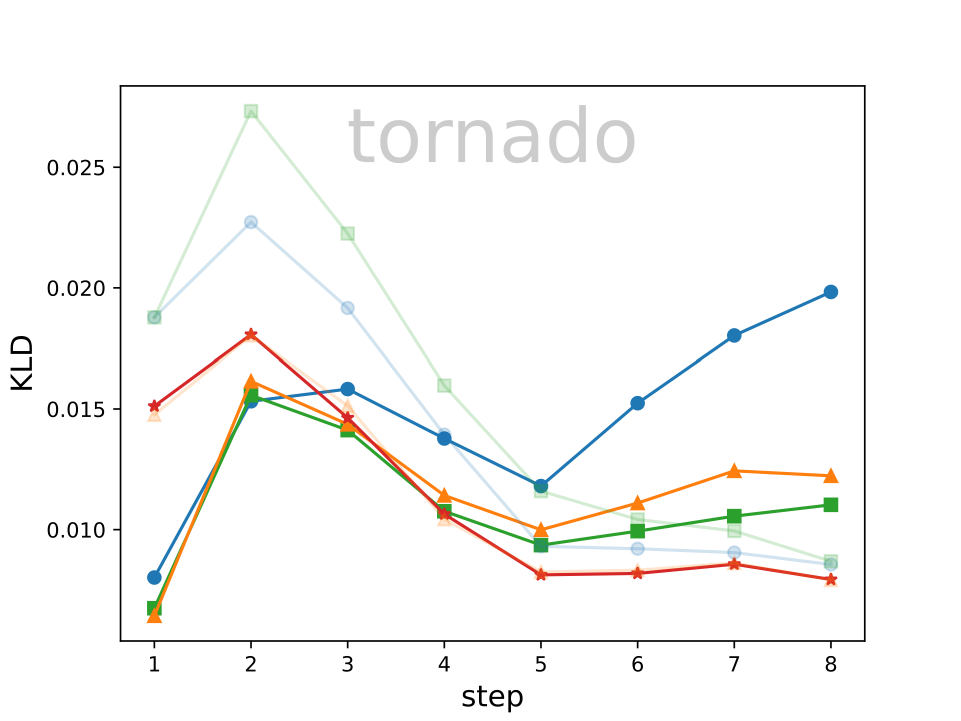}&
\includegraphics[width=0.24\linewidth]{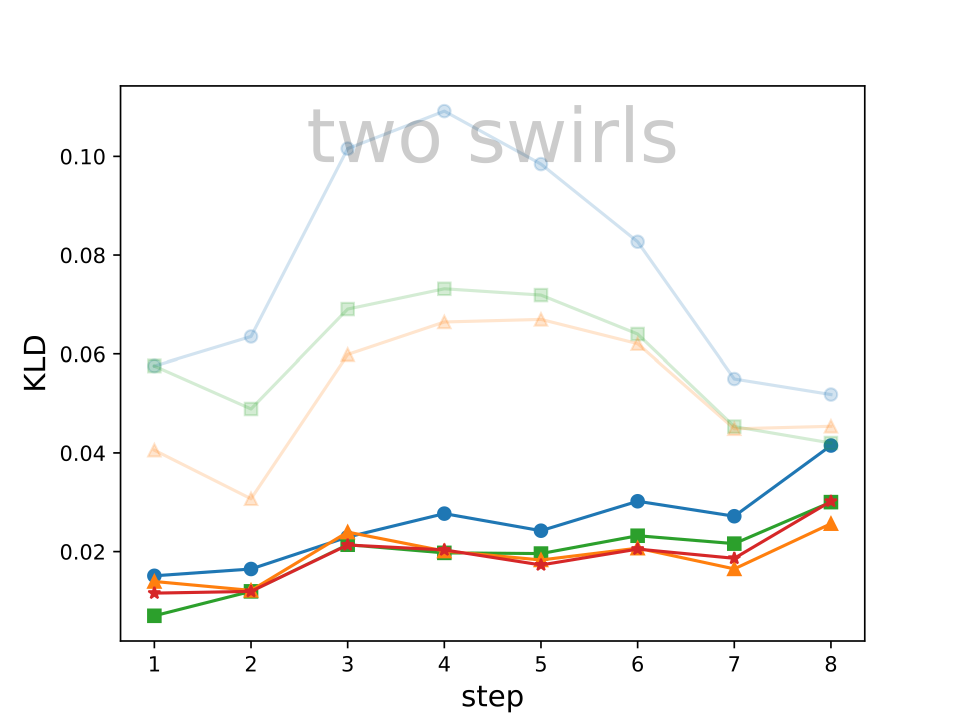}\\
\end{array}$
\end{center}
\caption{Estimation error of particle density (x-axis) over steps (y-axis) for each of the twelve data sets. The blue, green, orange, and red curves show the accuracy of {\fon}, {\semantic}, {\fixed}, and our {\ours}, respectively. Curves with lighter color show the estimation error of respective original networks without our edge optimization, and curves with darker color show the error of optimized networks.
 }
\label{fig:estimate}
\end{figure*}

\subsection{Particle Density Estimation}
\label{sec:density_estimate}
\hot{\bf Experiment setup.} For this task, we compute the three matrices $\matD$, $\matA$, and $\matT$ for each network and use Equation~\eqref{eq:estB} to estimate the numbers of particles over blocks at each step. We use Equation~\eqref{eq:loss} to accumulate the estimation error for the first eight steps. The reported error is divided by the number of particles, which converts the vector $\vecB$ back to a probability distribution and normalizes the loss. The initial positions of particles are evenly distributed in space for all the training, validation, and testing data. The ground-truth is obtained by tracing the particles with the fourth-order Runge-Kutta method and counting the number of particles in each data block. Table~\ref{table:estimate} shows the average estimation error and network size of each approach for each data set. 

\hot{\bf Impact of the order of network.} We first compare the networks of different orders using our {\ours} approach. In general, the estimation error decreases slightly and network size increases with the increase of order. The third-order network outperforms the second-order one for all data sets, and the fourth-order network outperforms the third-order one for most of the data sets, except the hurricane data set. But, in general, we find that the errors are similar across different orders. To balance between the error and the size of the network, we will use the third-order network for comparison among different approaches.

\hot{\bf Comparison with previous approaches.} Our {\ours} has smaller estimation errors than {\fon} and {\semantic} for all data sets. The average error of {\ours} (0.016) is more than $50\%$ smaller than the average of {\fon} (0.042) and {\semantic} (0.039). {\ours} also performs better that \fixed on most data sets (ten out of twelve), except the ABC and computer room. But the average improvement is smaller, with the average error of {\fixed} being 0.028. It should be noted that the {\fixed} achieves similar errors to {\ours} on many data sets, and the difference of average errors majorly come from a few data sets, such as the {\bernard}, crayfish, solar plume, and two swirls. For these data sets, {\fixed} has errors larger than 0.05. For the solar plume data set, {\ours} also has an error of 0.051, but for all the other data sets, the errors are smaller than 0.02. In terms of the size, {\fon} has the smallest number of nodes without considering HO-statuses. For the HON approaches, {\ours} is the smallest on average (454.3), which is $34.4\%$ smaller than {\semantic} (692.6) and $71.7\%$ smaller than {\fixed} (1606.6).

\hot{\bf Impact of our edge optimization.} All the existing approaches benefit from our optimization procedure. On average, the estimation errors reduce by 26.6\% for {\fon}, 45.9\% for {\semantic}, and 30.6\% for {\fixed}. The small size of {\fon} may limit the power to precisely describe the transition patterns, leading to the smallest improvement with our optimization. On the contrary, {\fixed} may already capture most transition patterns and has less room to improve, compared with {\semantic}. After the optimization, {\fon} still has the largest average error (0.023), and {\fixed} has the smallest error (0.013). {\ours} (0.016) and {\semantic} (0.016) have similar errors, which are close to that of {\fixed}. But {\ours} is 34.4\% smaller in size and achieves smaller errors on ten of the twelve data sets. However, in general, we find that the edge optimization is more important than the initialization of nodes, and the optimized performance is constrained by the size of the network. Figure~\ref{fig:estimate} shows the error estimation over tracing steps. We find that the estimation errors usually follow similar patterns for all approaches, except for those without edge optimization (lighter color curves). We can confirm that {\fixed} with our edge optimization usually deliver the best accuracy, but ours are often close. \hot{This implies that the existing graph-based approaches may be equipped with our network for better performance. For example, random walks may simulate the particle movement on our network to deliver better time performance than {\fixed} and more accurate estimation results than {\semantic} and {\fon}.}

\begin{figure*}[t]
\begin{center}
 $\begin{array}{c@{\hspace{0.02in}}c@{\hspace{0.02in}}c@{\hspace{0.02in}}c}
\includegraphics[width=0.24\linewidth]{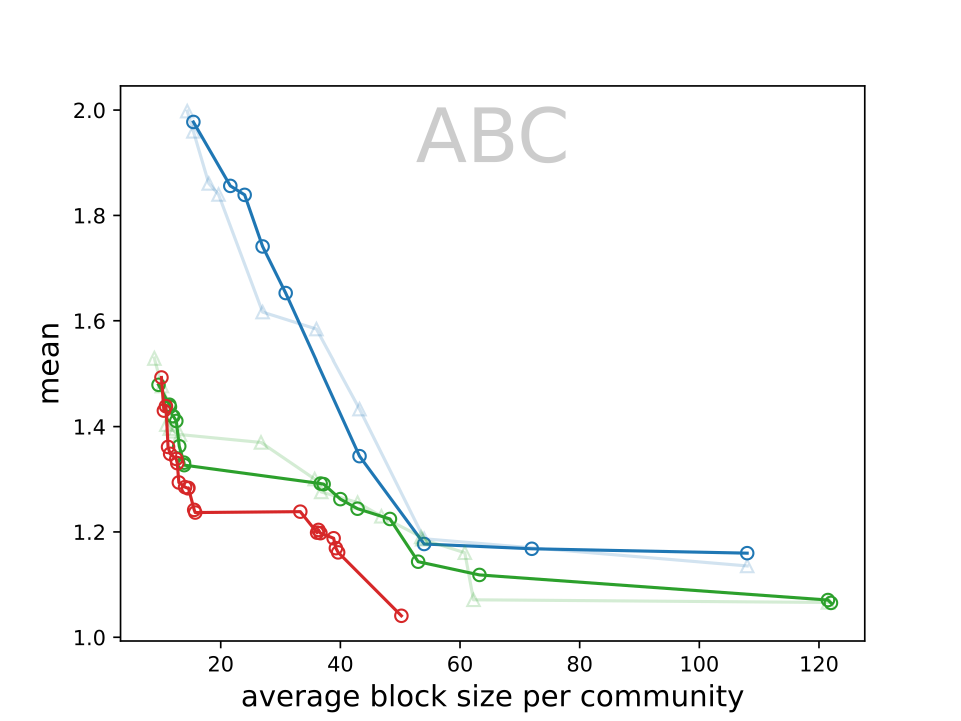}&
\includegraphics[width=0.24\linewidth]{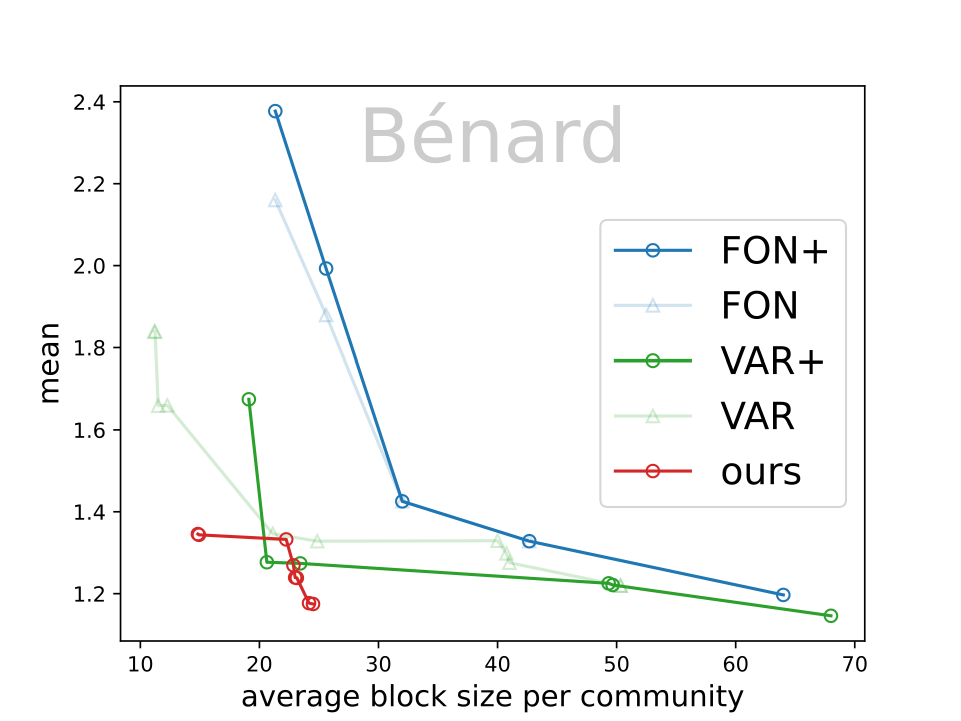}&
\includegraphics[width=0.24\linewidth]{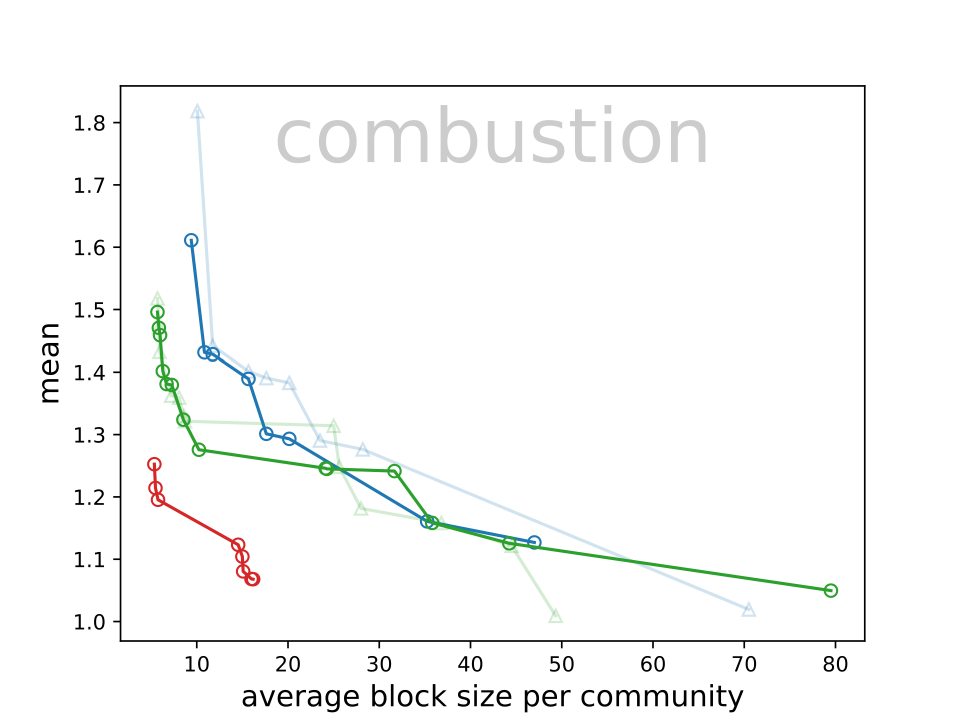}&
\includegraphics[width=0.24\linewidth]{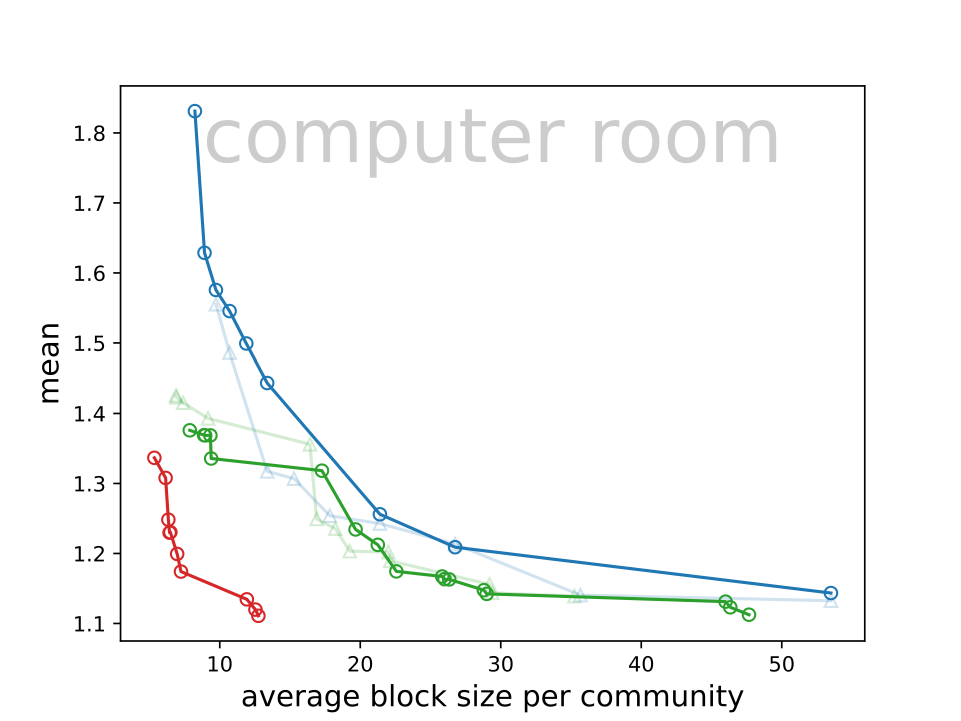}\\
\includegraphics[width=0.24\linewidth]{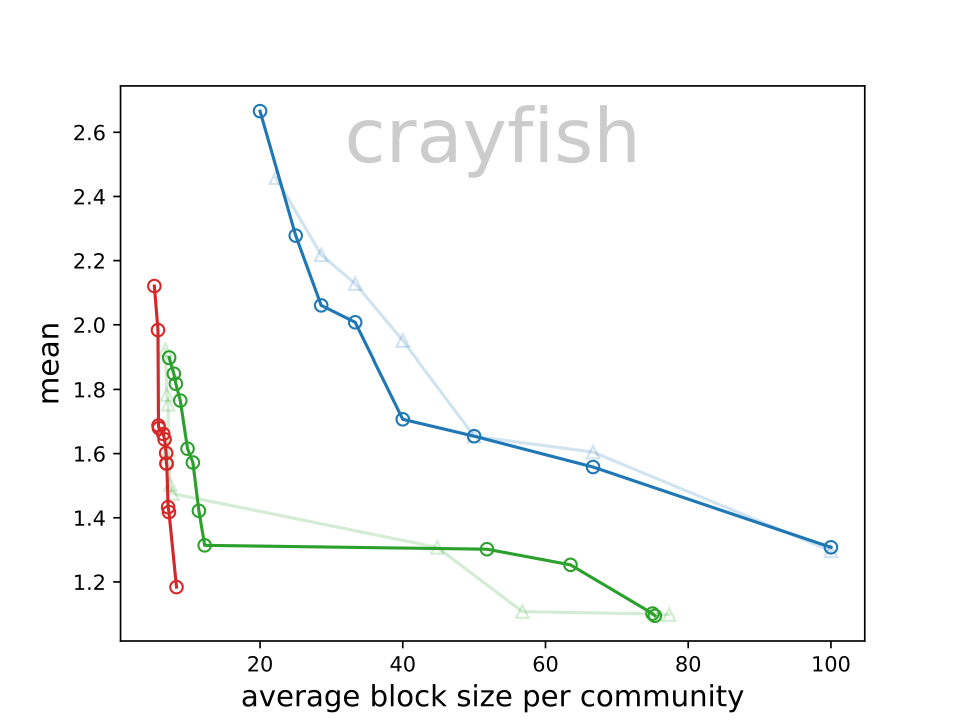}&
\includegraphics[width=0.24\linewidth]{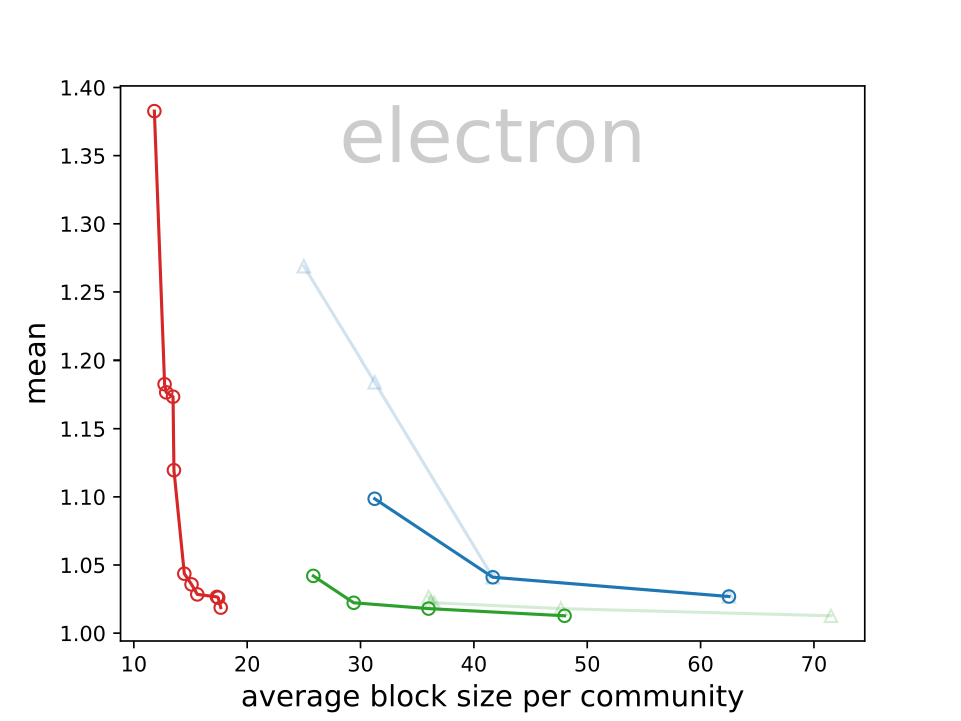}&
\includegraphics[width=0.24\linewidth]{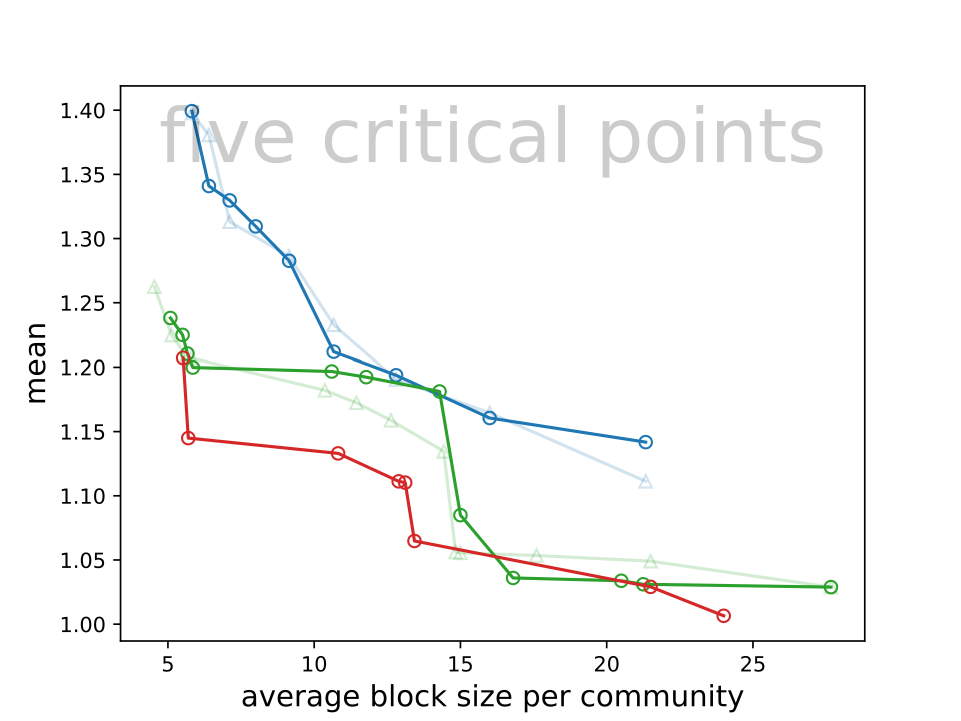}&
\includegraphics[width=0.24\linewidth]{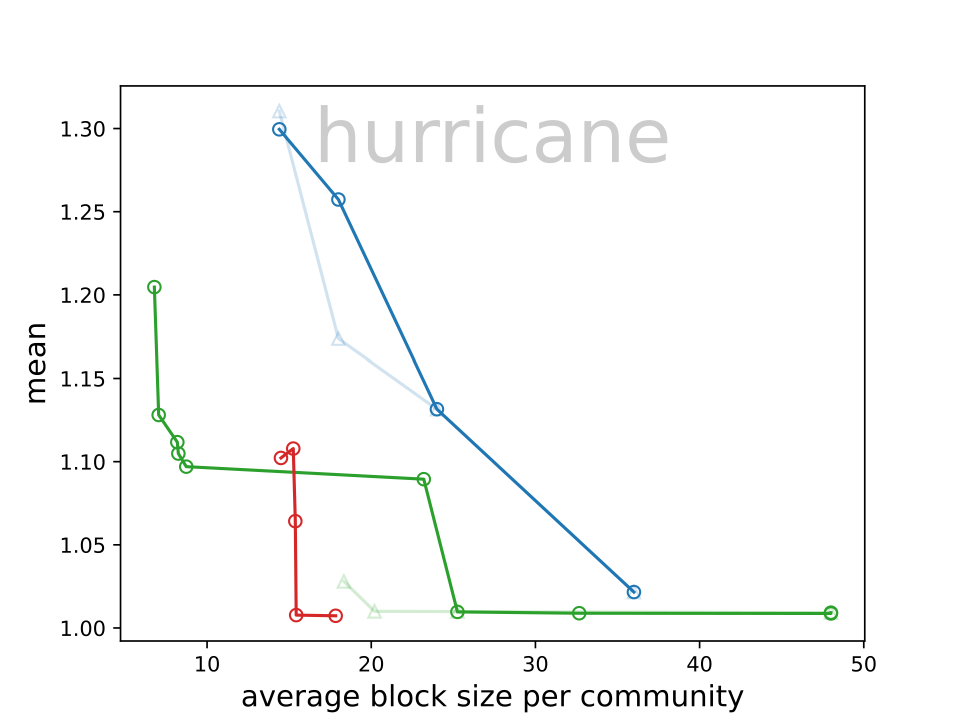}\\
\includegraphics[width=0.24\linewidth]{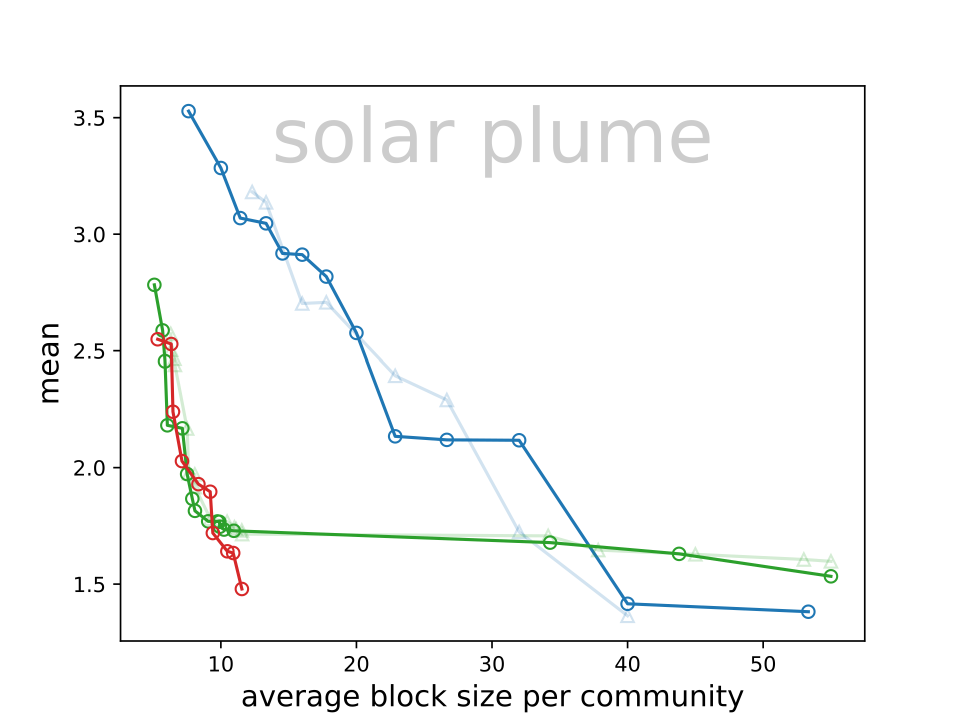}&
\includegraphics[width=0.24\linewidth]{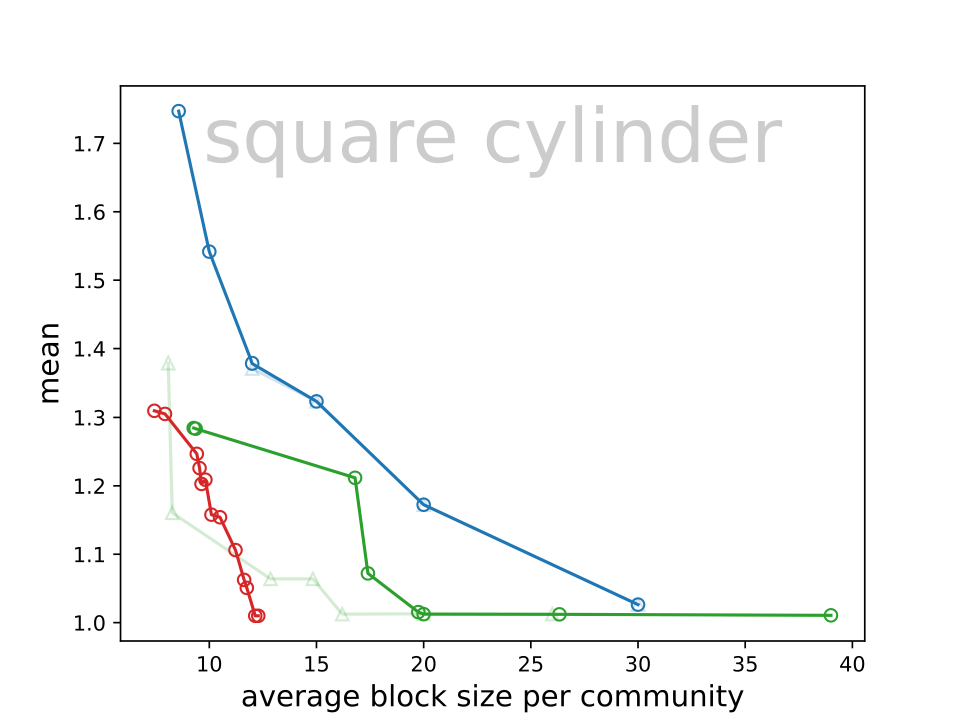}&
\includegraphics[width=0.24\linewidth]{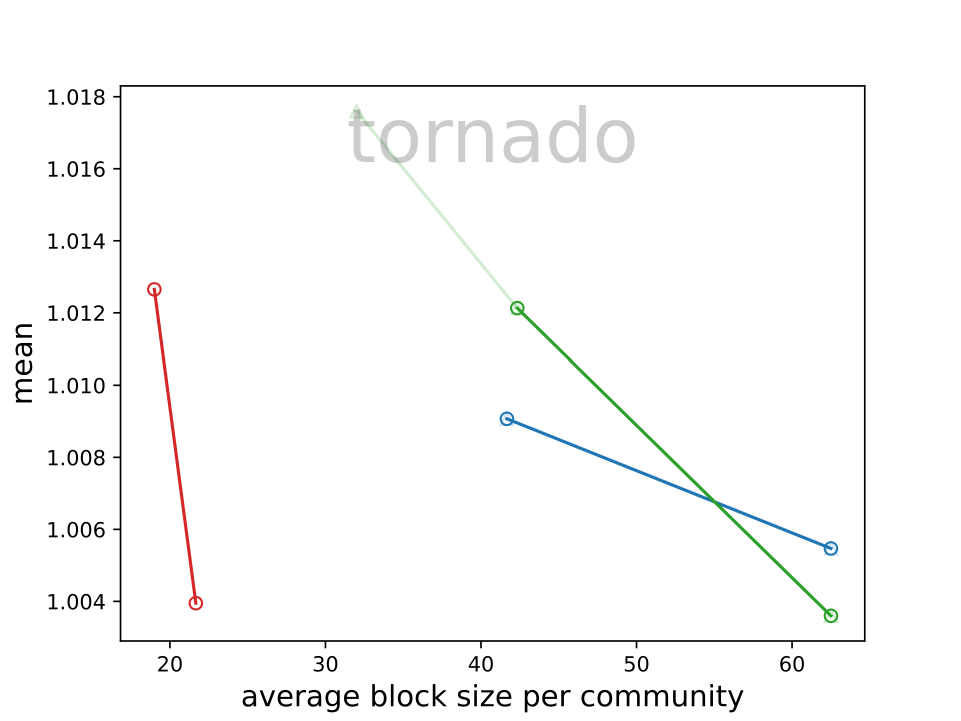}&
\includegraphics[width=0.24\linewidth]{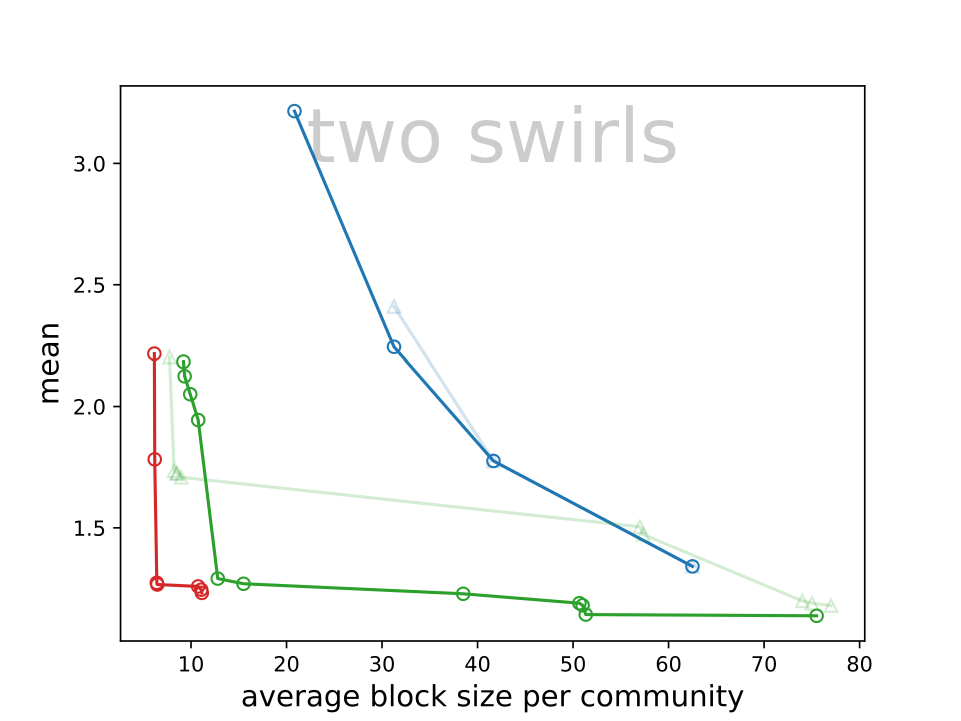}\\
\end{array}$
\end{center}
\caption{The average community size (x-axis) and the mean of the number of communities visited (y-axis). For the same community size, smaller means of communities visited is preferred. The blue curves denote the result of {\fon}, the green curves denote that of {\semantic}, and the red curves denote our {\ours}. Curves with lighter color show the result of respective original networks without our edge optimization, and curves with darker color show the result of optimized networks.}
\label{fig:fig_community_detection}
\end{figure*}

\subsection{Community detection}
\label{sec:community}
\hot{\bf Experiment setup.} In parallel particle tracing, data-parallelism and task-parallelism partition require effective flow field partition to reduce the amount of particle exchanging or data loading~\cite{zhang2018survey}.
In this task, we apply a community detection algorithm on {\ours} to examine its effectiveness to guide the flow field partition. We evaluate the partition results by tracing particles and counting the mean of the number of community visits. Specifically, we track the movement of particles on HO-nodes and increase the count when the community being visited changes. The count increases when a particle moves from a community to a different one (including a previously visited one). A count of one means that the particle always stays in a single community. In this way, the mean of community visits can be used to estimate the number of times that a particle will be exchanged during tracing when task-parallelism is used or the number of data loading needed when data-parallelism is employed.

We use InfoMap~\cite{rosvall2008maps}, a flow-based community detection method, to identify community modules in the network. This approach uses a parameter ``Markov-time" to control the resulting community resolution. Generally, higher ``Markov-time" results in a smaller number of communities with larger sizes. To avoid trial-and-error effort to select this parameter, we evenly sample the parameter from 0.5 to 3.5 with a step of 0.1, and record the resulting average community sizes and the average numbers of communities visited using a sample set of particles. The parameter values at the Pareto front are used in our testing stage with new sets of particles. For this task, we compare our approach with {\fonPlus} and {\semanticPlus}. We do not compare with {\fixedPlus} as that may produce many small communities; each contains a bundle of streamlines. 

\hot{\bf Community detection results.} Figure~\ref{fig:fig_community_detection} shows the average community sizes (in blocks) versus the mean of community visits for all data sets based on the testing set of particles. The average community size indicates the average amount of data to be loaded for each computation node. Given the resource limits, a preferred parameter value should lead to the smallest mean of community visits with an affordable community size. In Figure~\ref{fig:fig_community_detection}, we find that all curves decrease monotonically when the average community size increases. This indicates that the parameter values collected from the sampling particles can provide useful hints for the testing particles as well. Comparing with {\fonPlus} and {\semanticPlus}, our {\ours} produces smaller means of community visits in most cases. An exception is the solar plume data set, for which our {\ours} and {\semanticPlus} performs similarly for smaller community sizes. But our {\ours} rarely produces large communities, while {\semanticPlus} may produce much larger communities with a marginal decrease in the mean of community visits. Additionally, the HON approaches (i.e., {\ours} and {\semanticPlus}) outperforms {\fon} for almost all cases.

\begin{figure}[t]
\begin{center}
 $\begin{array}{c@{\hspace{0.02in}}c}
\includegraphics[width=0.50\linewidth]{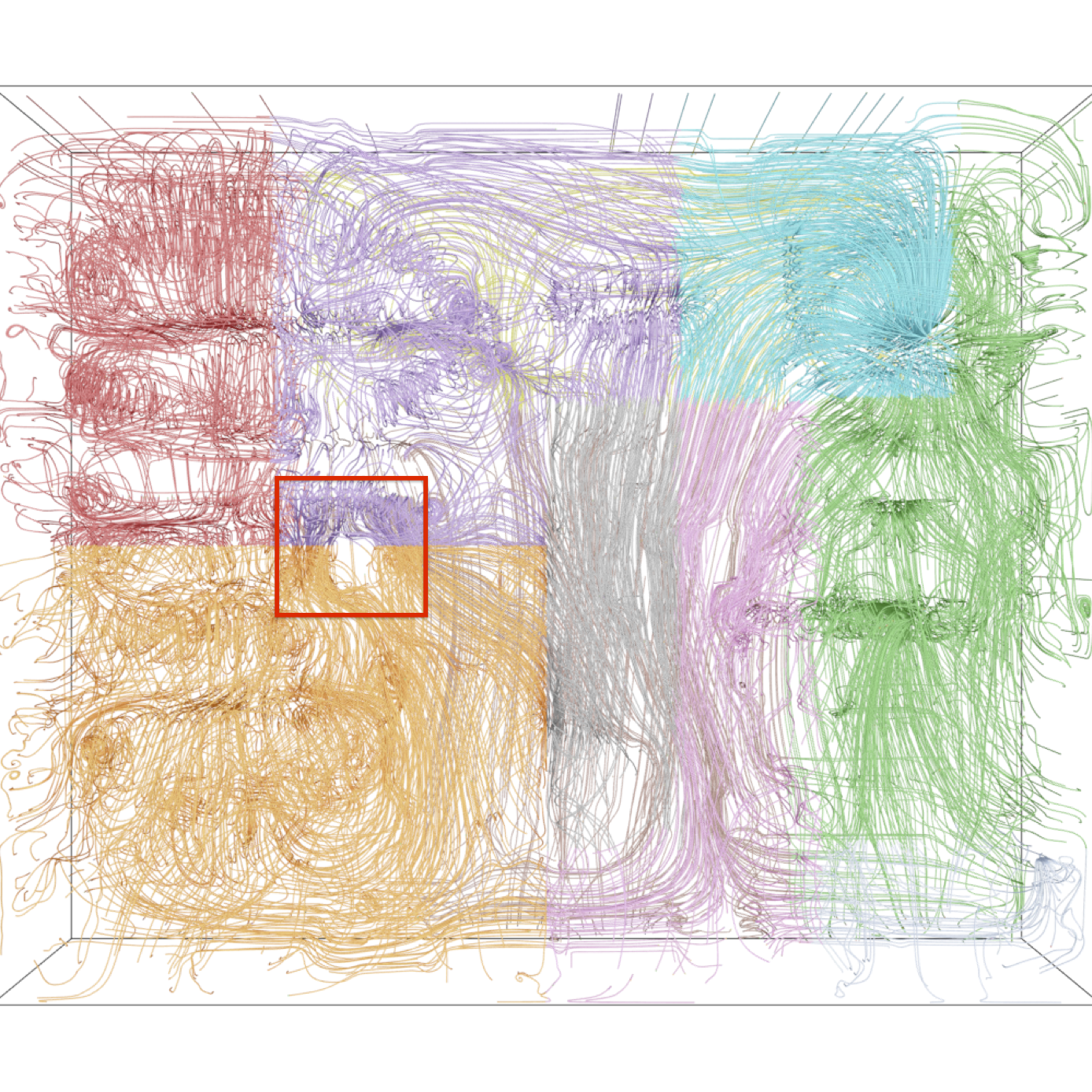}&
\includegraphics[width=0.30\linewidth]{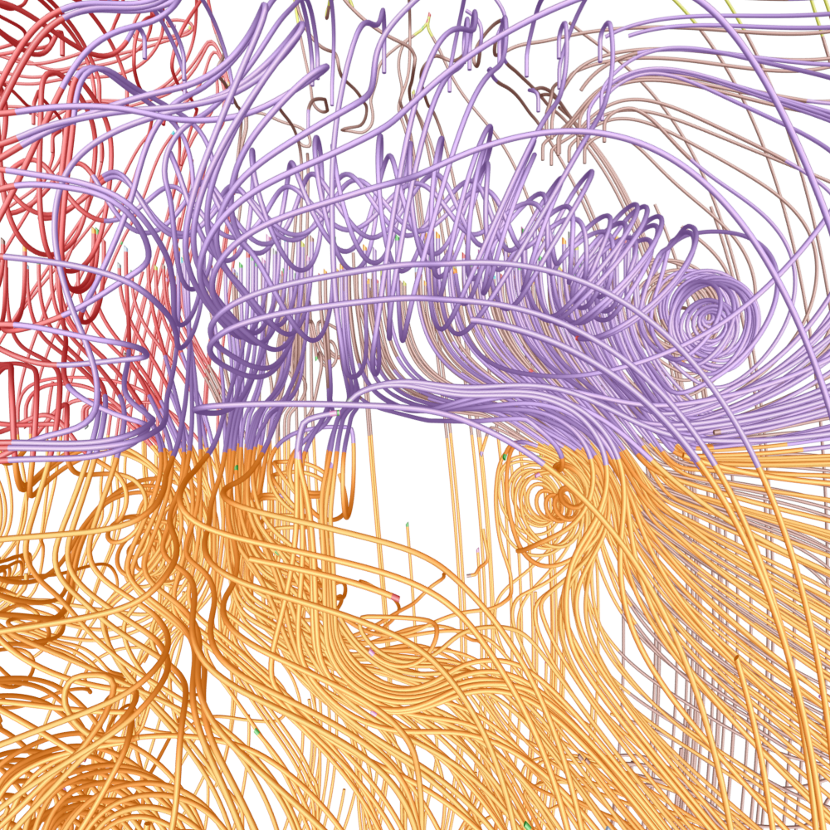}\\
\includegraphics[width=0.50\linewidth]{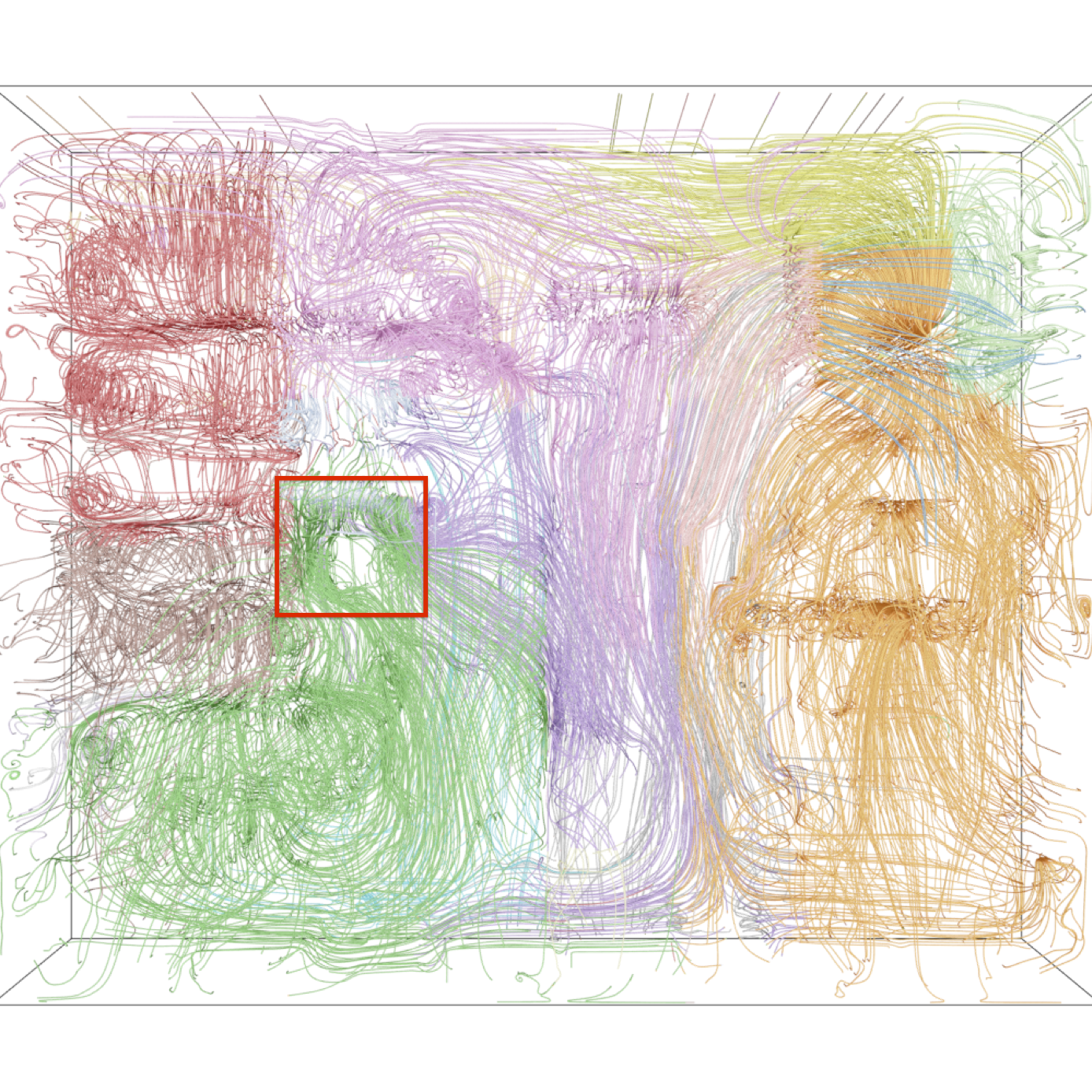}&
\includegraphics[width=0.30\linewidth]{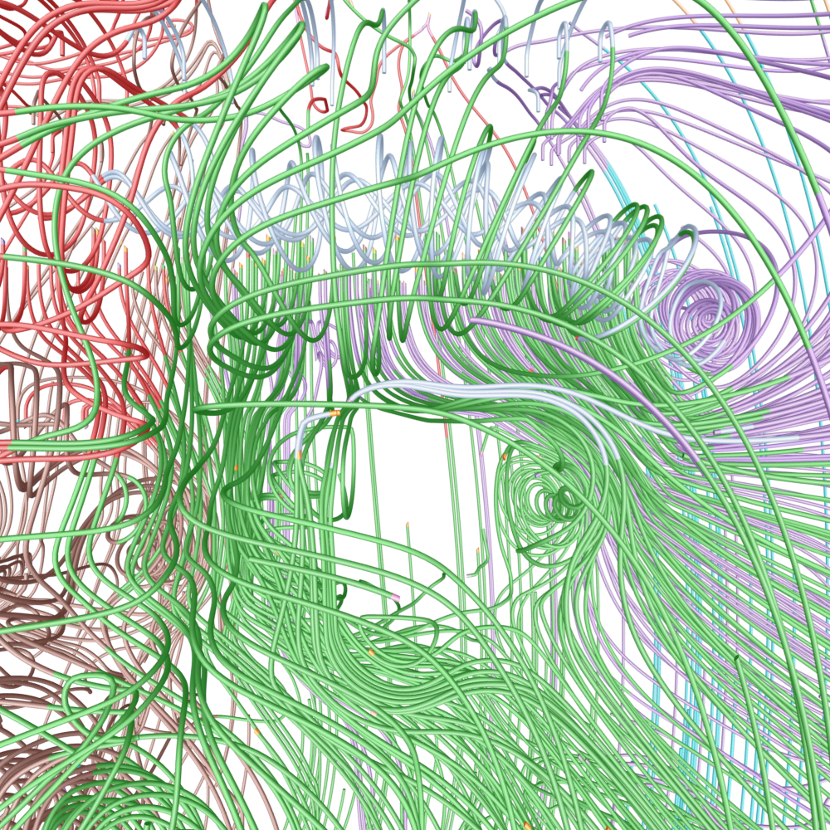}\\
\includegraphics[width=0.50\linewidth]{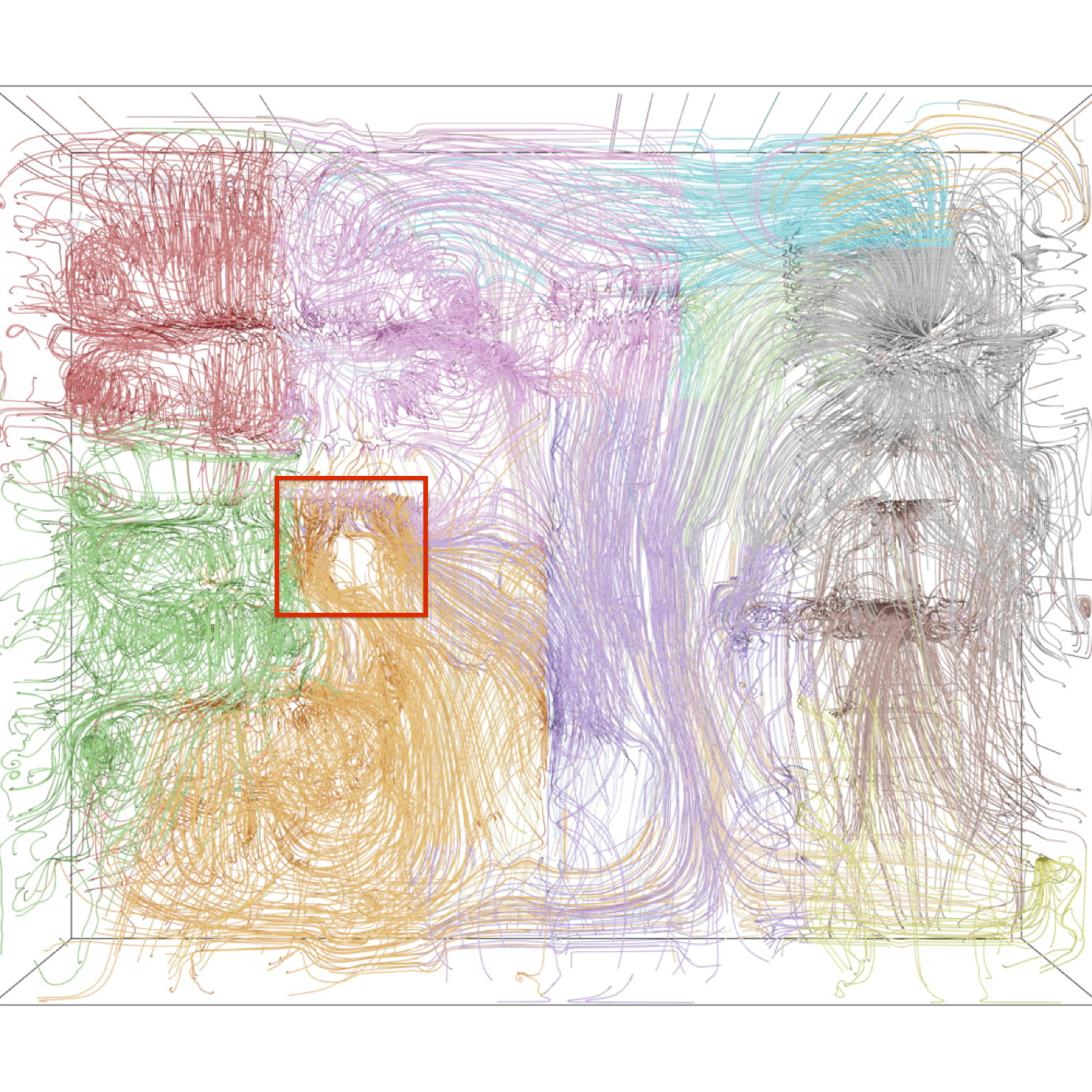}&
\includegraphics[width=0.30\linewidth]{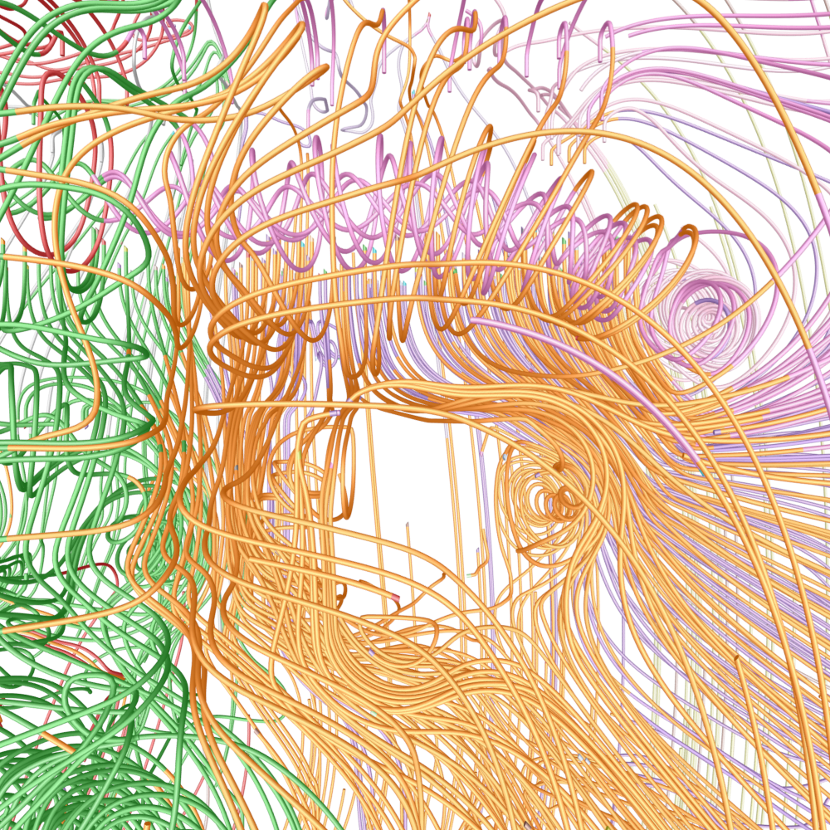}\\
\mbox{(a)} & \mbox{(b)}
\end{array}$
\end{center}
\caption{\hot{The data partition based on the community detection using the computer room data set. The streamline segments are colored by the communities. Note that a streamline may exhibit multiple colors when it passes multiple communities. (a) shows the partition results of the entire field, and (b) magnifies the regions in the respective red rectangles. The first, second, and third rows show the results of {\fonPlus}, {\semanticPlus}, and our {\ours}, respectively.}}
\label{fig:fig_community_3D_size}
\end{figure}

Figure~\ref{fig:fig_community_3D_size} visualizes an example of the data partition results with similar performance (the mean of community visits) using different approaches. Note that the communities are detected on the HO-nodes level. Therefore, a data block may belong to multiple communities for the HON approaches. Overall, we find that the HON approaches tend to produce smaller and overlapping communities for similar performance. This can be confirmed by the quantitative results: {\ours} produces an average community size of 5.37 with a mean of community visits of 1.37; {\semanticPlus} produces an average size of 6.89 with a mean of visits of 1.43; and, {\fon} produces an average size of 9.72 and a mean of visits of 1.55. We should note that the small and overlapping communities require more computation nodes but fewer resources on each node, which is more suitable for today's parallel computation hardware. Furthermore, Figure~\ref{fig:fig_community_3D_size} (b) confirms that the power of HONs resides in its ability to subdivide blocks. The HON approaches produce ``soft" boundaries of communities by assigning the particles of different behaviors in the boundary blocks into different communities. In contrast, FON only partitions the data over blocks, and, therefore, it creates ``hard" boundaries that may separate a flow feature in an unnatural way.

\begin{figure}[t]
\begin{center}
\includegraphics[width=0.98\linewidth]{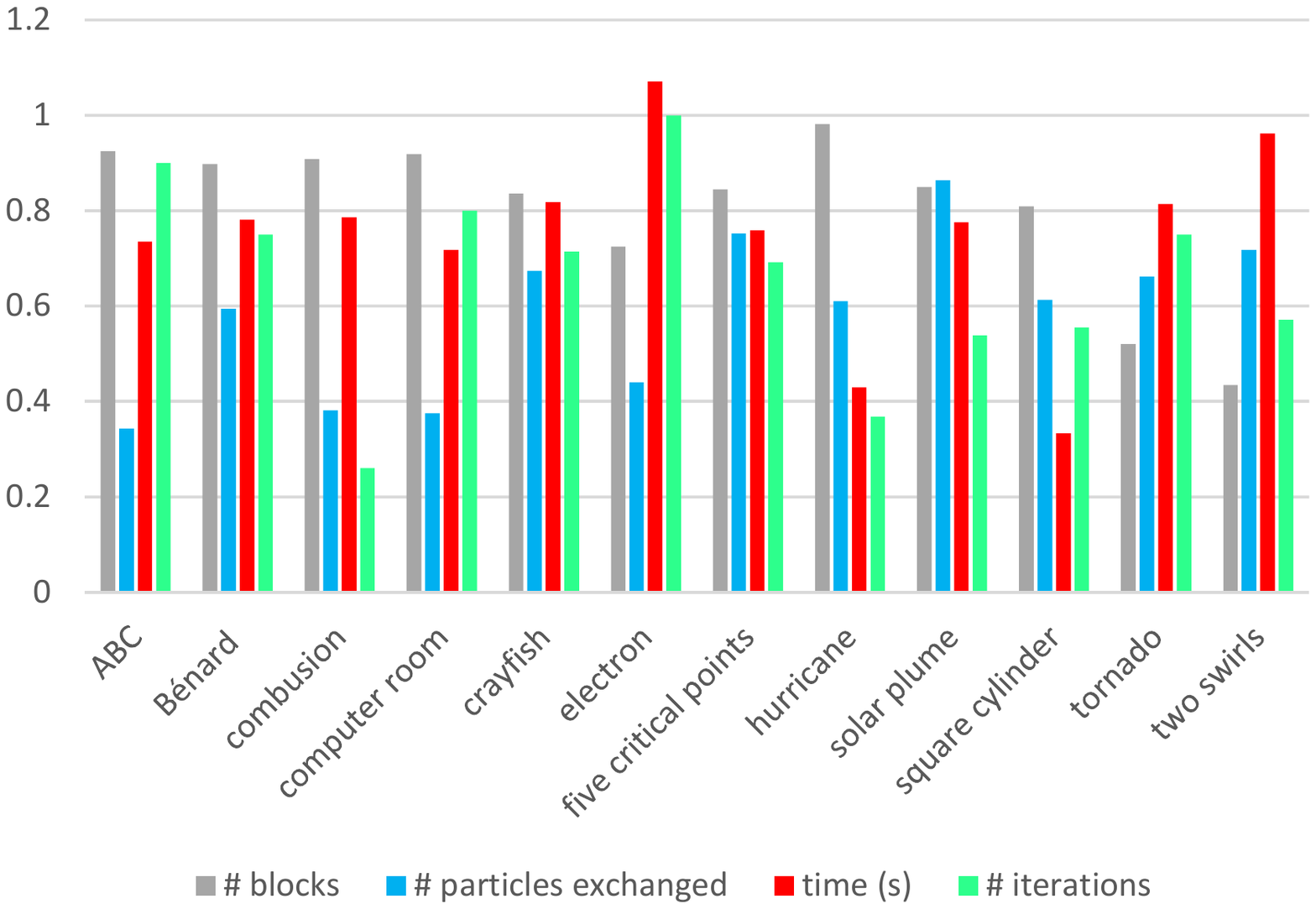}
\end{center}
\caption{\hot{Comparison of the parallel particle tracing performance delivered by the {\ours} communities and {\fon} communities. The gray bars are the ratios of the average sizes of {\ours} communities (in numbers of blocks) over that of {\fon} communities. The blue bars are the ratios of the numbers of particle exchanged when using {\ours} over that using {\fon}. The red bars are the ratios of timing in seconds using {\ours} over that using {\fon}. The green bars are the ratios of the numbers of iterations needed using {\ours} over that using {\fon}.}}
\label{fig:timing}
\end{figure}

\begin{figure}[t]
\begin{center}
$\begin{array}{c@{\hspace{0.02in}}c@{\hspace{0.02in}}c}
\includegraphics[width=0.45\linewidth]{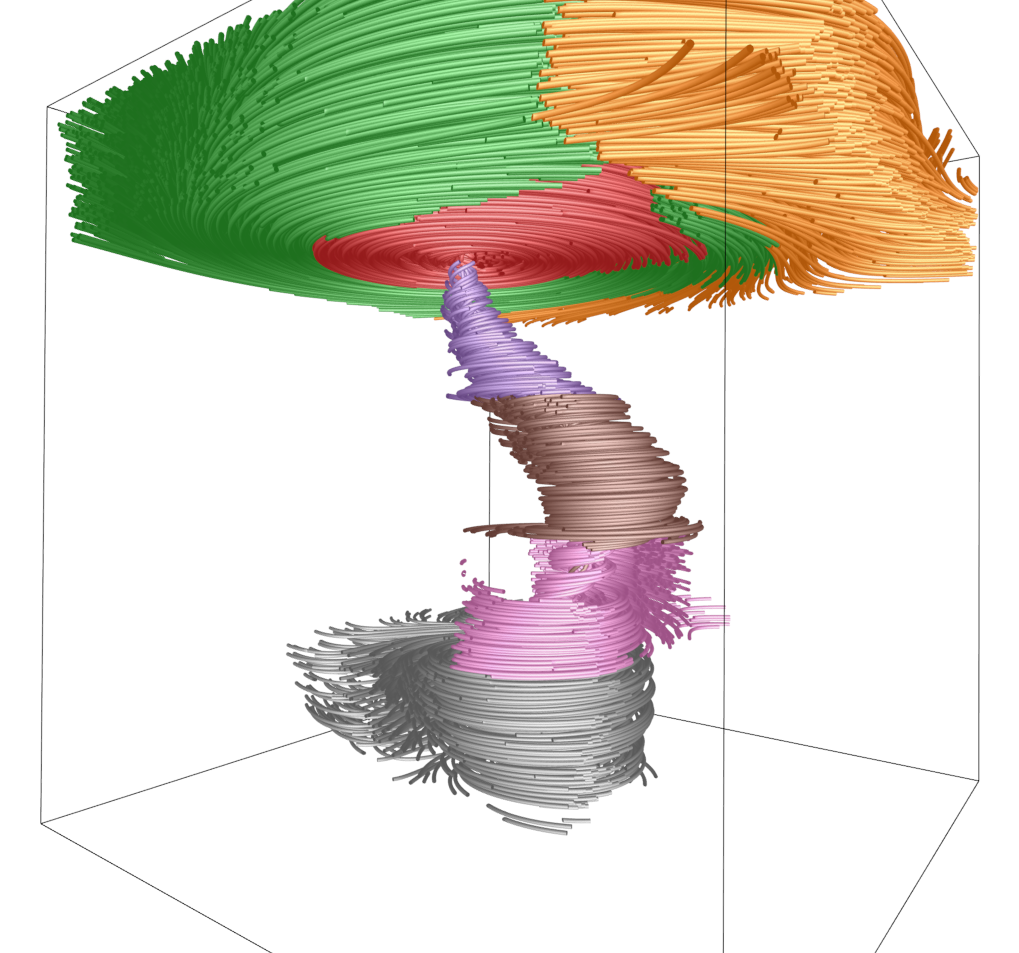}&
\includegraphics[width=0.45\linewidth]{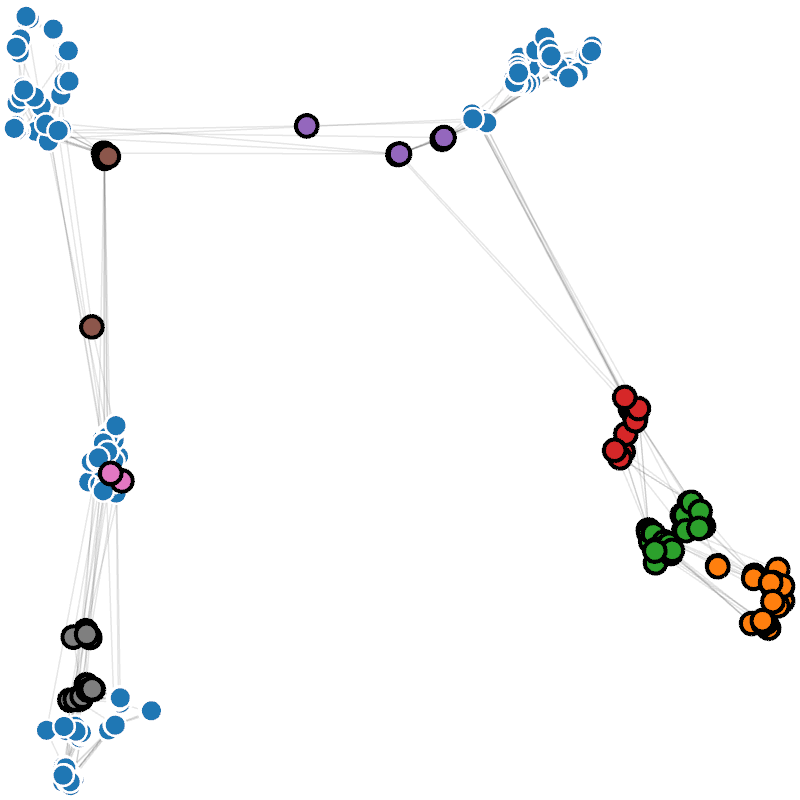}\\
\mbox{(a)} & \mbox{(b)}\\
\includegraphics[width=0.45\linewidth]{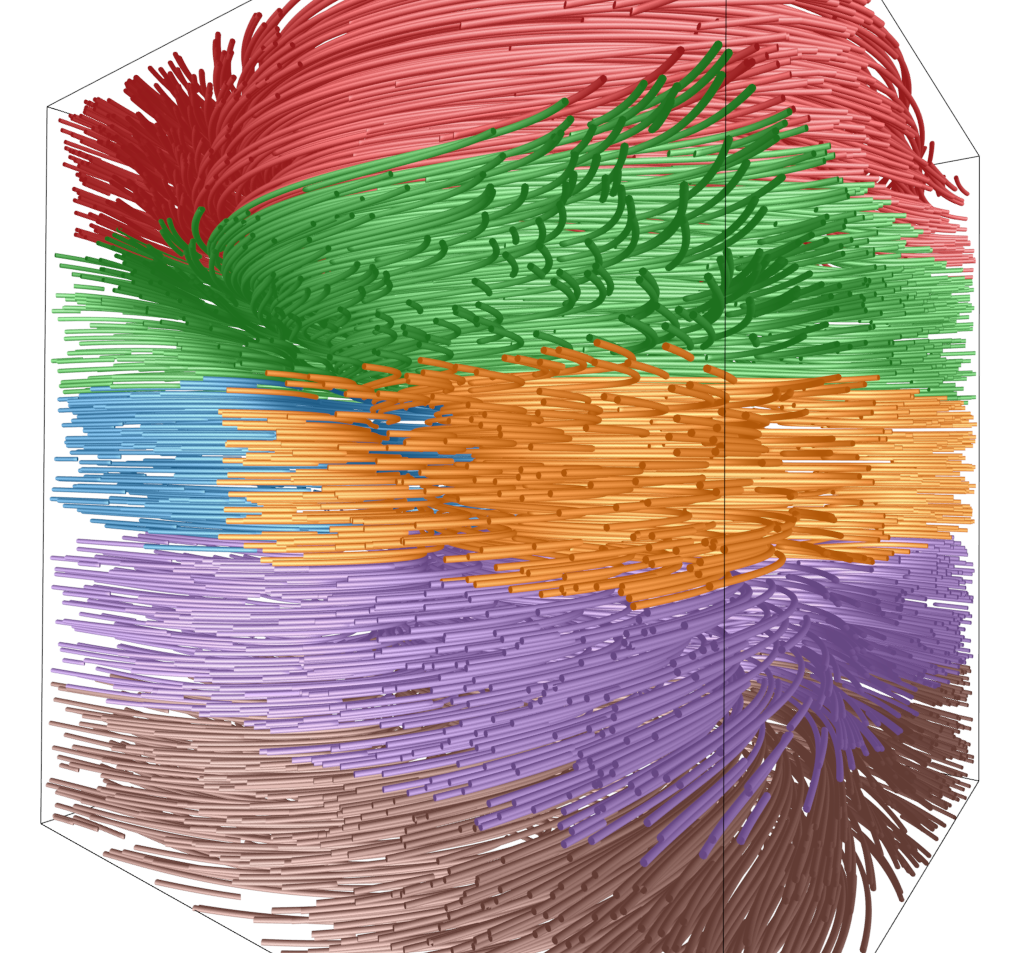}&
\includegraphics[width=0.45\linewidth]{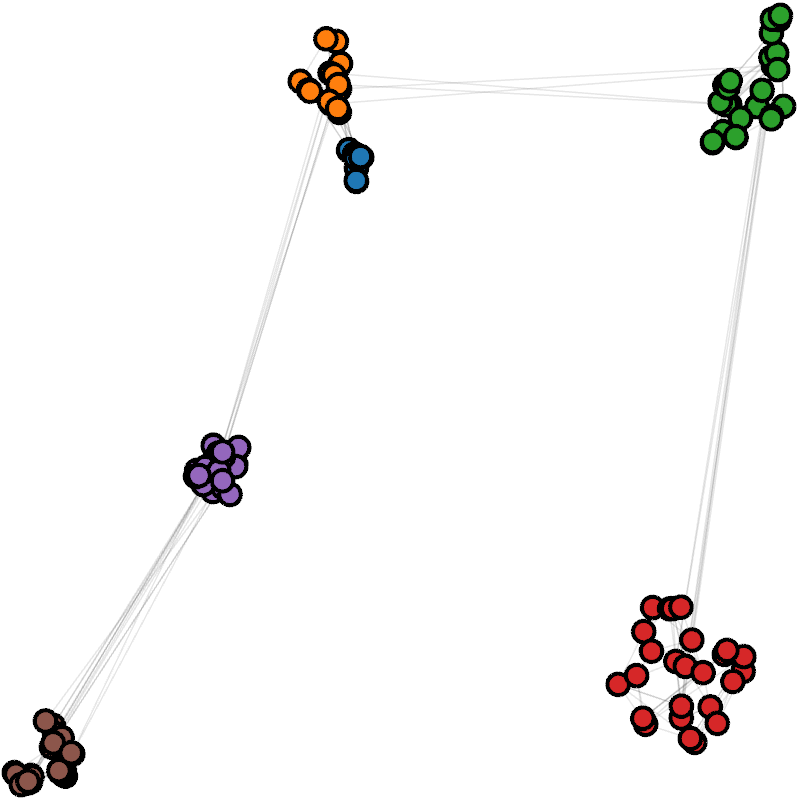}\\
\mbox{(c)} & \mbox{(d)}
\end{array}$
\end{center}

\caption{Visualization of networks and streamline segments corresponding to the selected nodes using the tornado data set. The first and the second row show the networks and streamline segments of our {\ours} and {\fon}, respectively.}
\label{fig:fig_case_1}
\end{figure}

\hot{
{\bf Preliminary timing results.} We implement a simple parallel particle tracing strategy using MPI to examine the performance delivered by the traditional {\fon} communities and the {\ours} communities. With this strategy, each computation node loads the data blocks corresponding to one community and processes the particles in that community. The computation is performed in multiple iterations. In each iteration, a computation node traces every particle assigned to the node until the tracing is finished or the particle goes out of the community. After each iteration, all computation nodes exchange the particles which are not completely traced. The particles are sent to a destination node based on its HO-states. The iteration repeats until all particles are fully traced. The experiment is conducted on a CPU cluster with sixteen computation nodes, where each node has an Intel Xeon E5-2692 CPU running at 2.2GHz with 64GB memory.

For fair comparison, we use configurations that produces similar average block sizes for {\fon} and {\ours}, guided by Figure~\ref{fig:fig_community_detection}. This ensure that both {\fon} and {\ours} consumes similar amount of resources. Figure~\ref{fig:timing} shows the comparison in four aspects: the average number of blocks, the number of particles exchanged during tracing, the time to trace all particles, and the number of iterations to perform all tracing steps. Note that, as the data sets vary significantly in these four numbers, we use the ratios between the numbers of {\ours} and those of {\fon} for easier comparison across data sets.

In Figure~\ref{fig:timing}, we can see that we allow {\fon} to load more blocks on average for all data sets, because the community detection algorithm cannot produce communities with exactly the same number of blocks. For most data sets, the numbers of blocks for {\fon} and {\ours} are similar with a ratio close to one. Two exceptions are the tornado and two swirls, for which even the smallest communities produced by {\fon} are still much larger than {\ours}. With $80.4\%$ of resources consumed per computation node, {\ours} still outperforms {\fon} in all aspects. Compared with {\fon}, {\ours} requires $58.6\%$ of the numbers of particles exchanged during tracing and $65.8\%$ of the numbers of iterations required, leading to a $74.9\%$ of running time on average. 

However, we should note that the naive parallel particle tracing does not fully leverage the power of {\ours}. On average, the reduction in run time ($25.1\%$) is not as significant as the reduction in the number of particles exchanged ($41.4\%$) and the number of iterations ($34.2\%$). For the electron data set, this is even more obvious. With the same number of iterations and only $44.0\%$ of particles exchanged, {\ours} even requires slightly more time ($7.1\%$) to trace the particles than {\fon}. This is due to the unbalanced workload across computation nodes and iterations. In the later iterations, several particles traveling between a few computation nodes may delay the entire computation. As the total amount of computation is constant, the inefficient execution in these iterations reduces the overall utilization of computation power, which leads to a longer execution time.

A more sophisticated parallel particle tracing algorithm may avoid this by balancing the workload, and {\ours} can also be beneficial in this aspect for two reasons. First, {\ours} provides a more accurate estimation of the particle densities, as shown in Table~\ref{table:estimate}. This can provide a more accurate estimation of the number of particles processed by each computation node. Second, {\ours} distinguishes different patterns inside a block. As the trajectories of particles may vary significantly in length, {\ours} may facilitate a more accurate estimation of the amount of computation required by different kinds of particles. The improvement in these two aspects may boost the performance of workload estimation, and, therefore, enhance the workload distribution.
}

\hot{
\subsection{Visual Exploration}
\label{sec:visualization}
We further examine the constructed network using the graph layout with the LinLog energy model~\cite{JGAA-154}. We implement a simple exploration system that supports brushing and linking between the graph visualization and the streamline visualization. Once a node is selected in the graph visualization, the corresponding streamline segments related to that node are visualized in the same color.
}

\begin{figure}[t]
\begin{center}
$\begin{array}{c@{\hspace{0.02in}}c@{\hspace{0.02in}}c}
\includegraphics[width=0.45\linewidth]{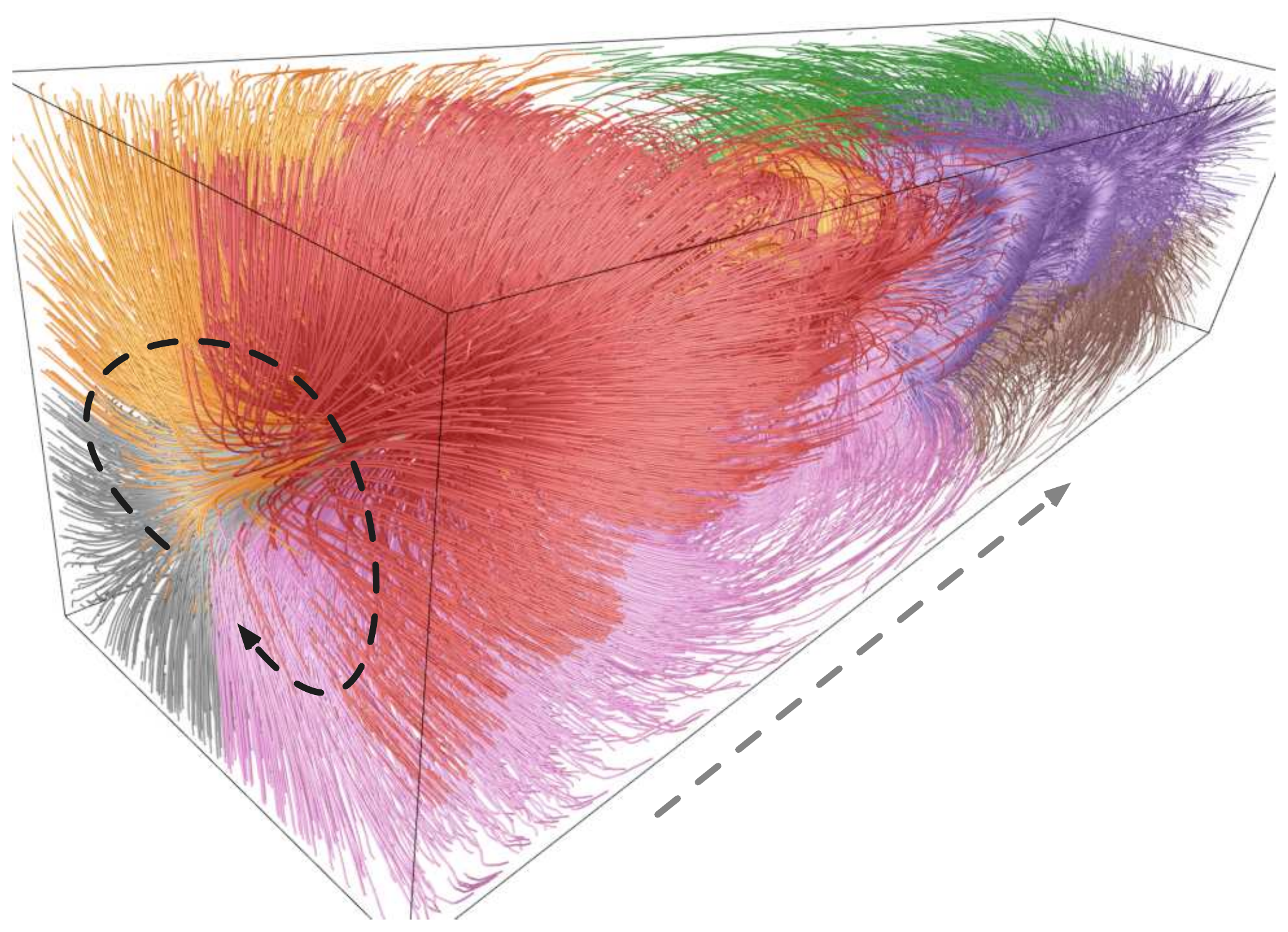}&
\includegraphics[width=0.45\linewidth]{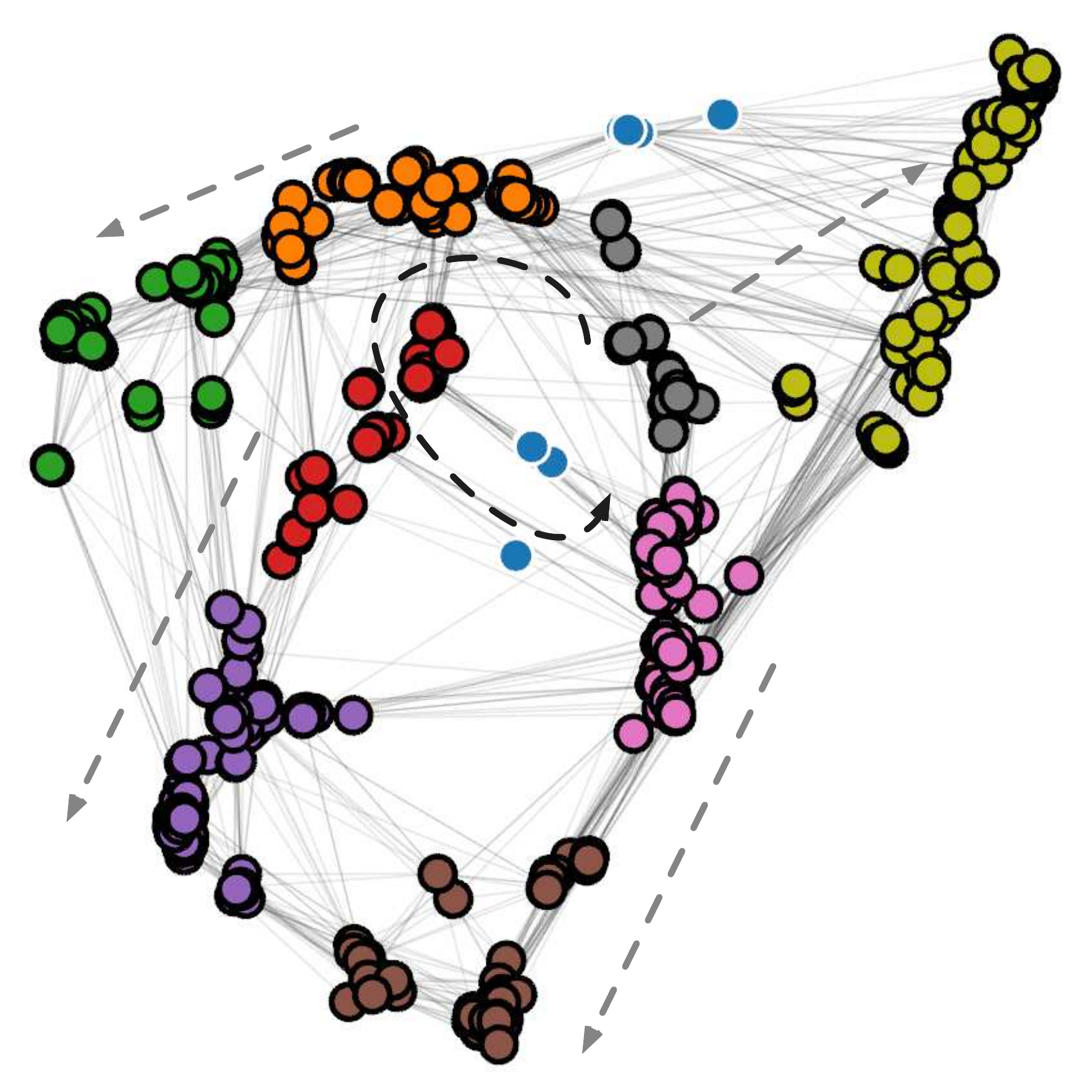}\\
\mbox{(a)} & \mbox{(b)}\\
\includegraphics[width=0.45\linewidth]{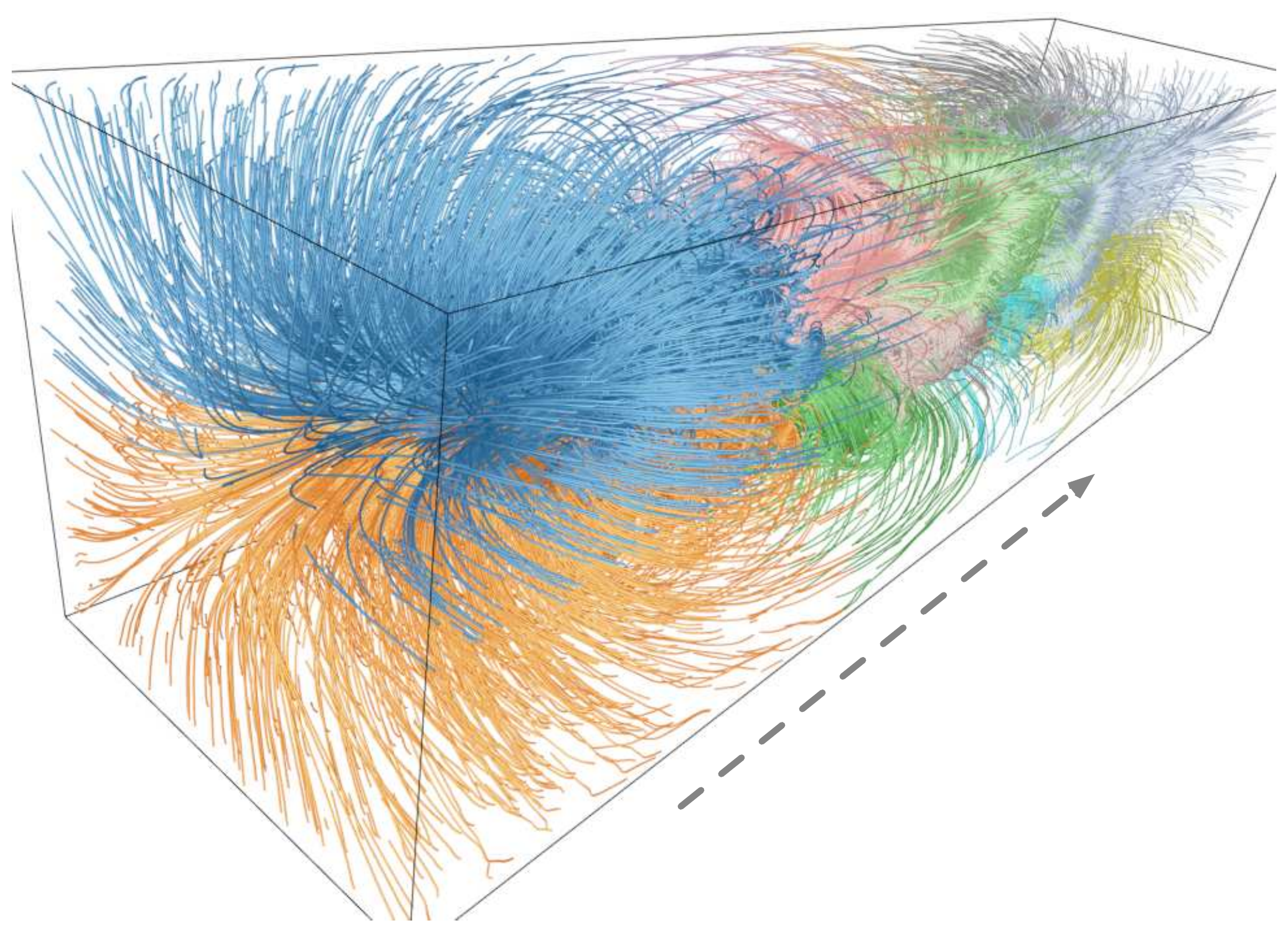}&
\includegraphics[width=0.45\linewidth]{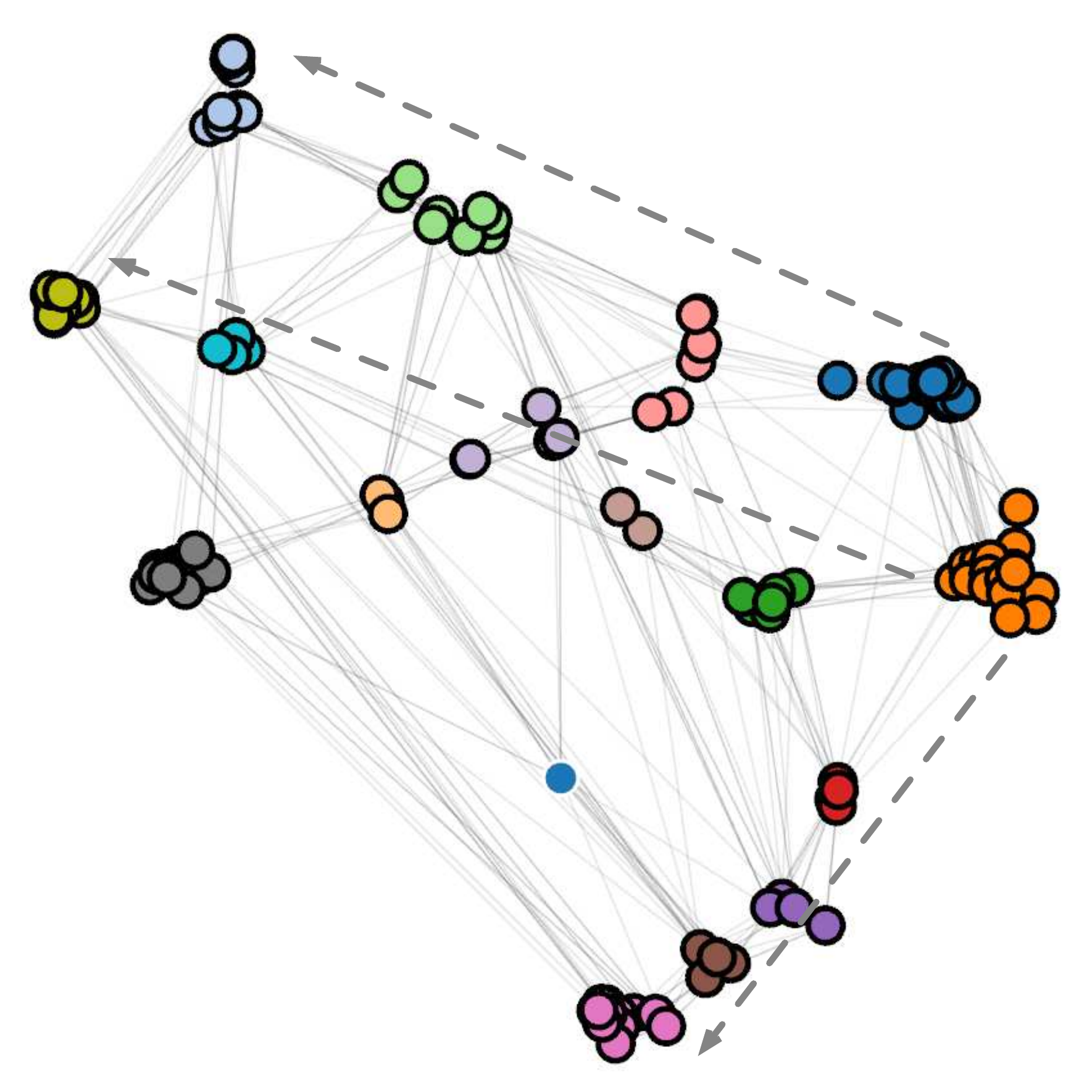}\\
\mbox{(c)} & \mbox{(d)}
\end{array}$
\end{center}
\caption{Visualization of networks and streamline segments corresponding to the selected nodes using the plume data set. The first and the second row show the networks and streamline segments of our {\ours} and {\fon}, respectively.}
\label{fig:fig_case_2}
\end{figure}

{\bf Tornado}. The tornado data set is divided into $5 \times 5 \times 5$ blocks. In Figure~\ref{fig:fig_case_1} (b) and (d), we find that both {\ours} and {\fon} exhibit five groups of nodes, corresponding to the five vertical layers of the data set. Additionally, for {\ours}, most of the five groups of nodes demonstrate a finer level of structures. For example, the group of nodes at the bottom-right corner can be easily divided into three smaller groups, as shown in orange, green, and red in Figure~\ref{fig:fig_case_1} (b). Their corresponding streamline segments locate at the top layer of the data set, where the segments related to the red nodes occupy the central region, as shown in Figure~\ref{fig:fig_case_1} (a). For other groups, we can also observe some nodes that are visually separable from other nodes, as colored in purple, brown, pink, and gray. These nodes correspond to streamlines at the core of the tornado as well. This shows that {\ours} captures the core of the tornado. The network produced by {\ours} is consistent with our understanding of the data set: only the particles around the core of the tornado will move across vertical block layers, while the particles at the outer layer of the tornado will mostly move horizontally. The particle transitions along the vertical core enhance the connections among HO-nodes corresponding to the core of the tornado and drag them away from other nodes at the same vertical layer. However, {\fon} fails to separate the tornado core from other streamline segments, as the tornado core does not align with the boundaries of blocks. 


{\bf Solar Plume}. The solar plume data set is partitioned into $4 \times 4 \times 10$ data blocks. In Figure~\ref{fig:fig_case_2} (b), we find that the nodes in {\ours} form eight distinct groups in the graph layout. The four groups at the center (red, orange, gray, and light purple) correspond to the four sectors of the crown of the solar plume. The node groups and the sectors follow the same order (as indicated by the black arrow), preserving the neighboring relations between the groups. The node groups at the outer ring (green, purple, brown, and yellow) correspond to the tail of the solar plume. They also follow the same order, and each group stays close with the corresponding group in the inner layer. In Figure~\ref{fig:fig_case_2} (c), {\fon} does not distinguish the four sectors at the crown clearly. Instead, the upper and lower blocks form two groups of nodes (blue and orange) in the graph. We can see two clear paths in the upper region of the graph, corresponding to the blocks in the front. We can also see a path in the bottom-left region, but the fourth path is not clearly visible.

\begin{figure}[t]
\begin{center}
$\begin{array}{c@{\hspace{0.02in}}c@{\hspace{0.02in}}c@{\hspace{0.02in}}c}
\includegraphics[width=0.32\linewidth]{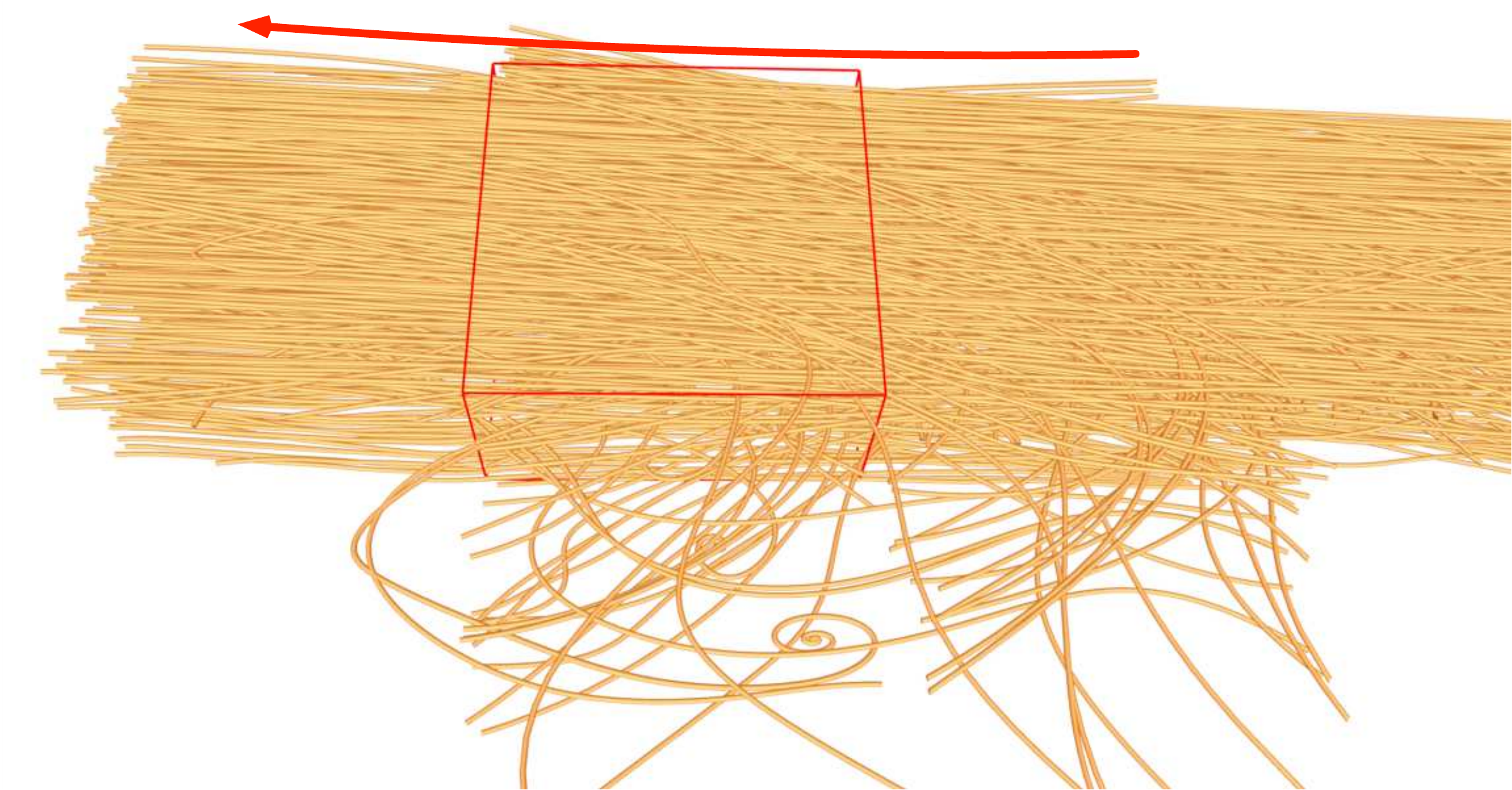}&
\includegraphics[width=0.32\linewidth]{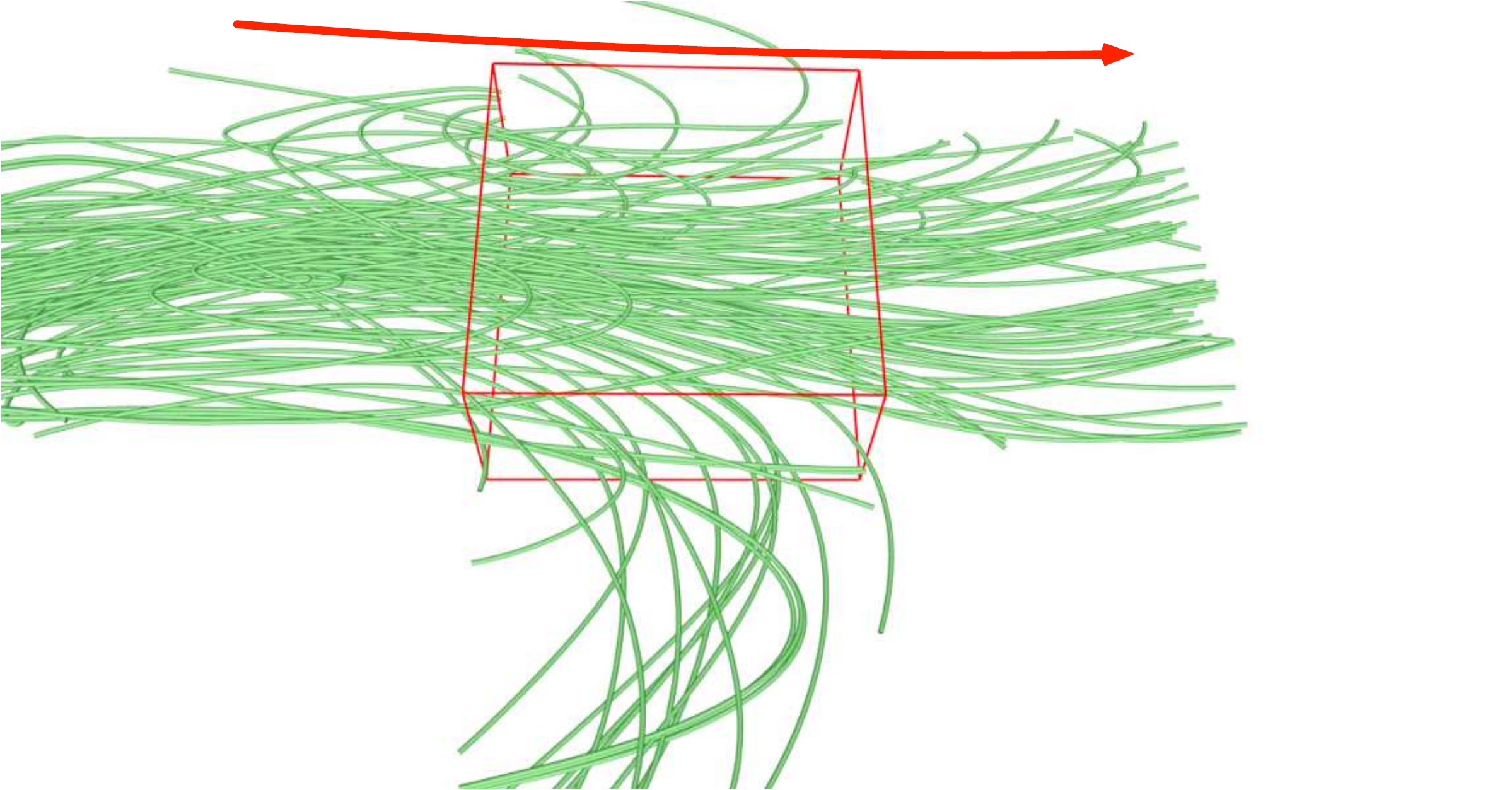}&
\includegraphics[width=0.32\linewidth]{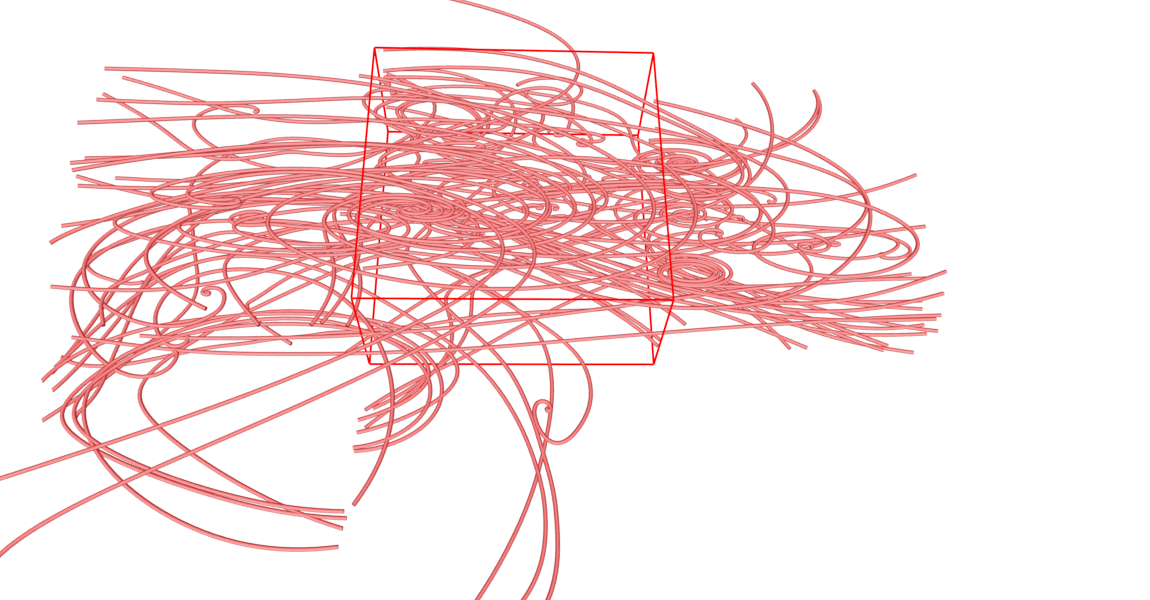}\\
\mbox{(a)} & \mbox{(b)} & \mbox{(c)}\\
\includegraphics[width=0.32\linewidth]{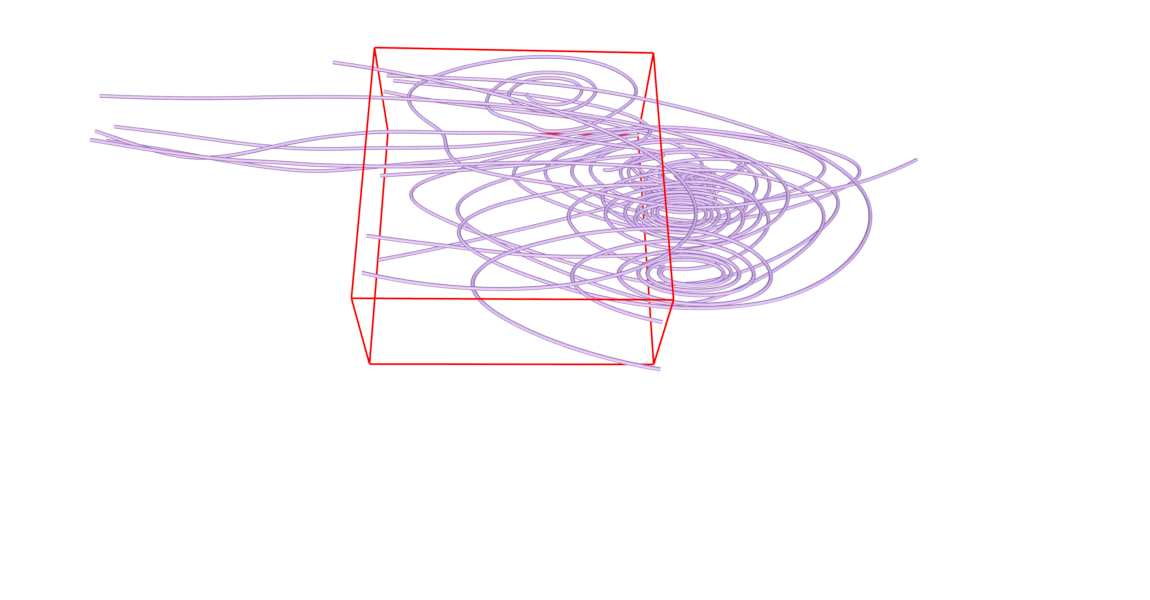}&
\includegraphics[width=0.32\linewidth]{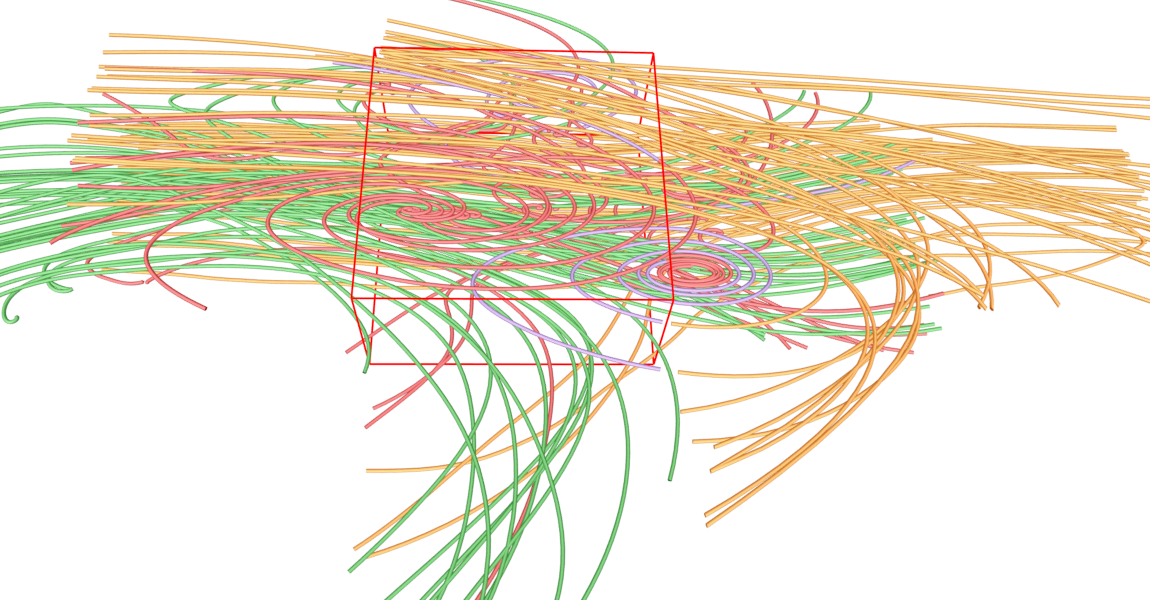}&
\includegraphics[width=0.32\linewidth]{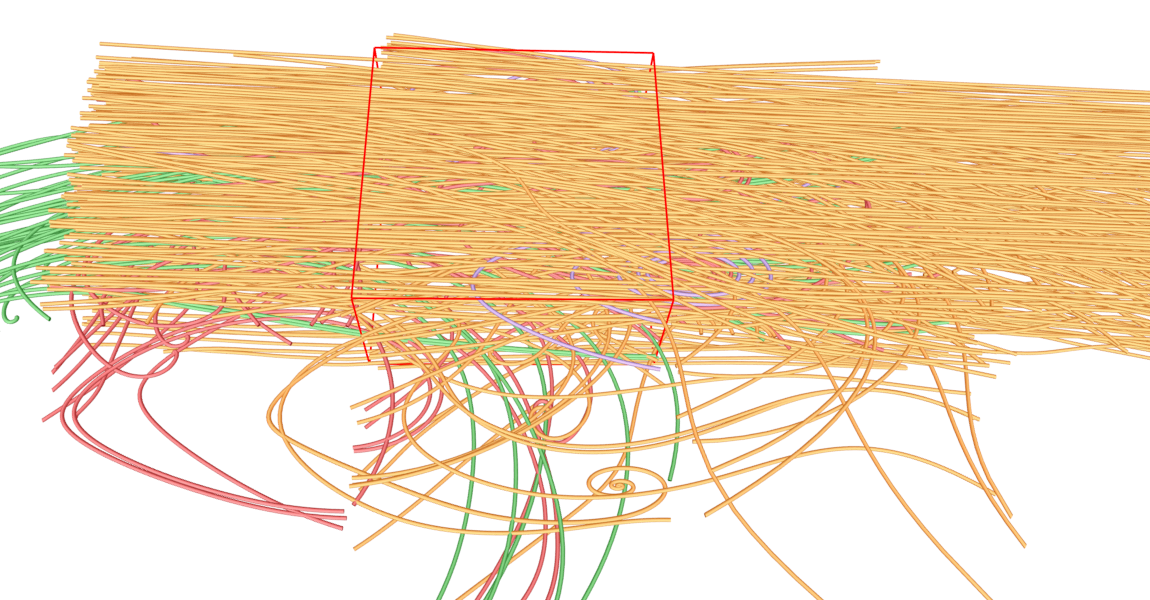}\\
\mbox{(d)} & \mbox{(e)} & \mbox{(f)}
\end{array}$
\end{center}
\caption{\hot{Flow patterns related to a single block in the unsteady ECMWF data set. (a) to (d) show the pathlines related to the four major HO-nodes in the block highlighted in red box, respectively. (e) and (f) shows the pathlines of different HO-nodes at the time step 2 and 8, respectively. The HO-nodes are produced by our {\ours} approach.}}
\label{fig:unsteady-vis}
\end{figure}

\begin{figure}[t]
\begin{center}
$\begin{array}{c@{\hspace{0.02in}}c@{\hspace{0.02in}}c}
\includegraphics[width=0.45\linewidth]{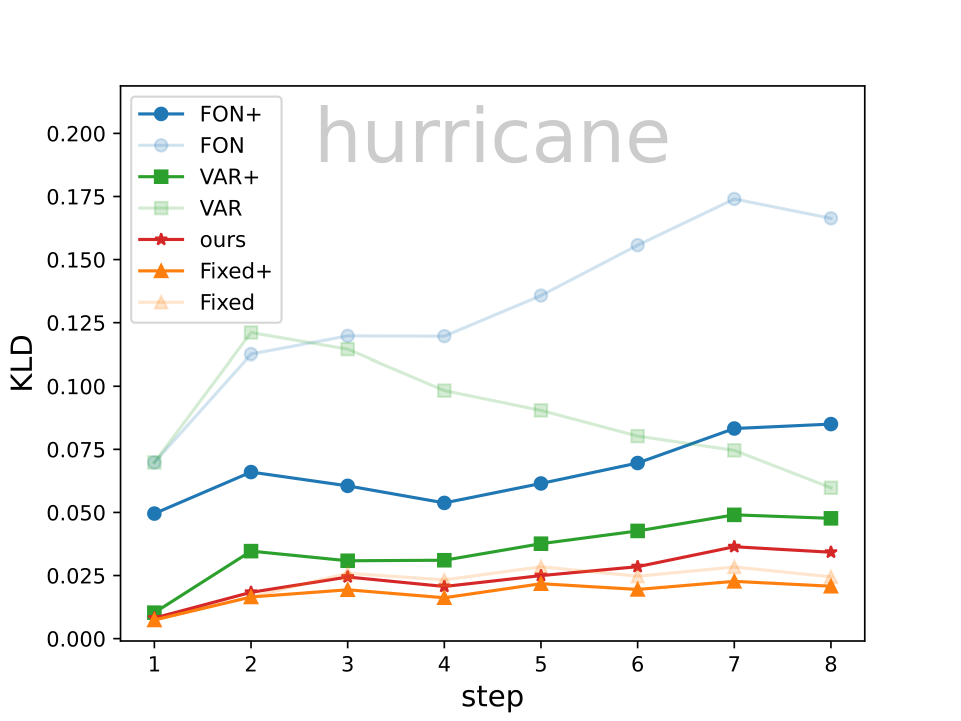}&
\includegraphics[width=0.45\linewidth]{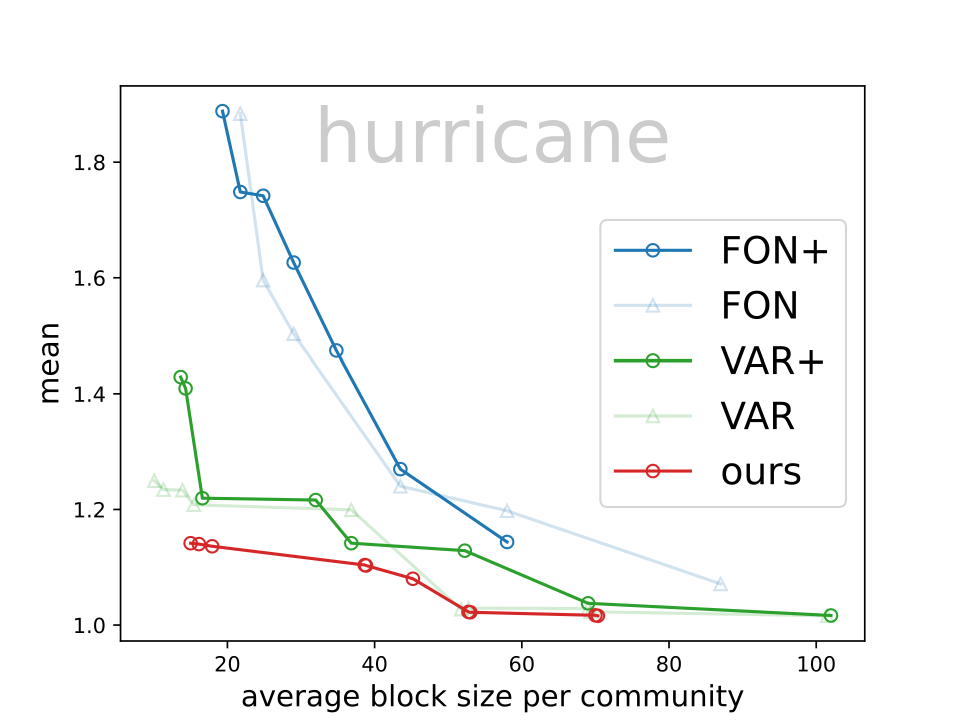}\\
\includegraphics[width=0.45\linewidth]{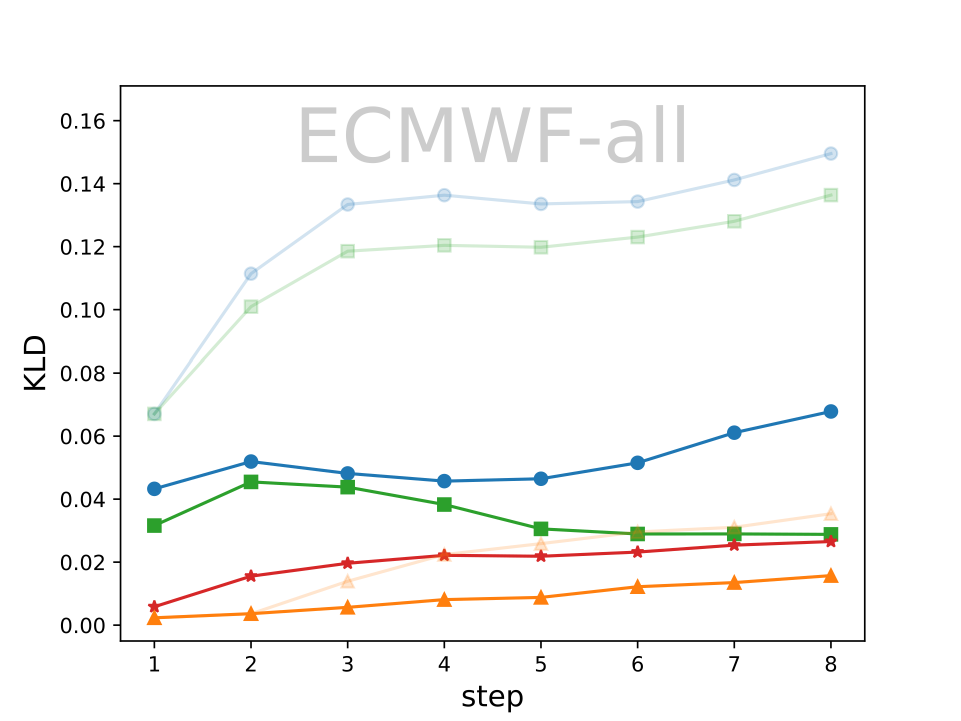}&
\includegraphics[width=0.45\linewidth]{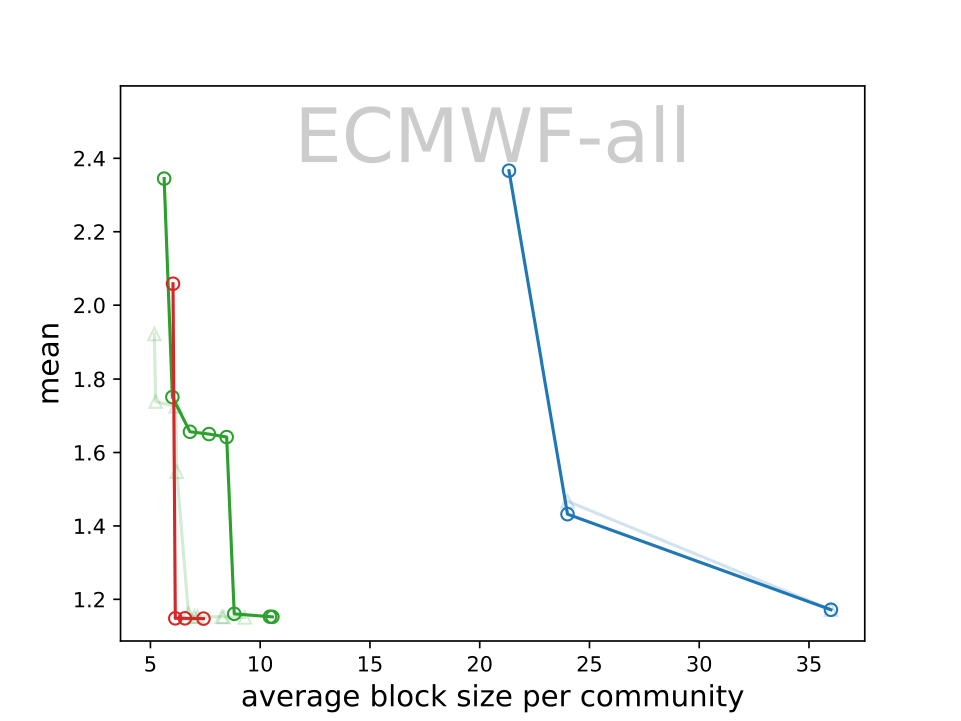}\\
\includegraphics[width=0.45\linewidth]{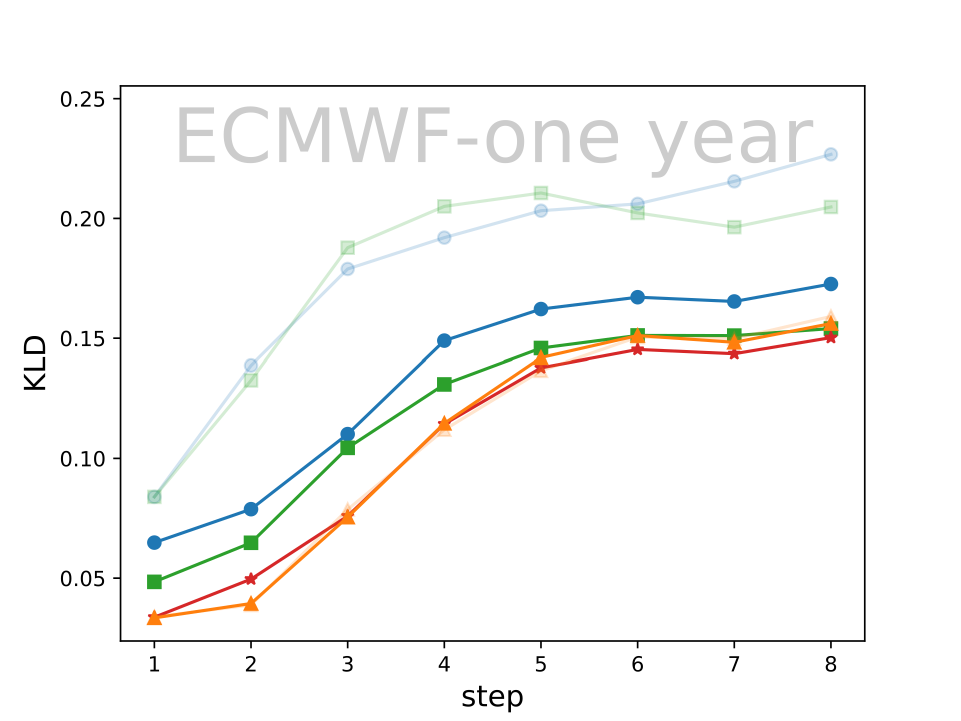}&
\includegraphics[width=0.45\linewidth]{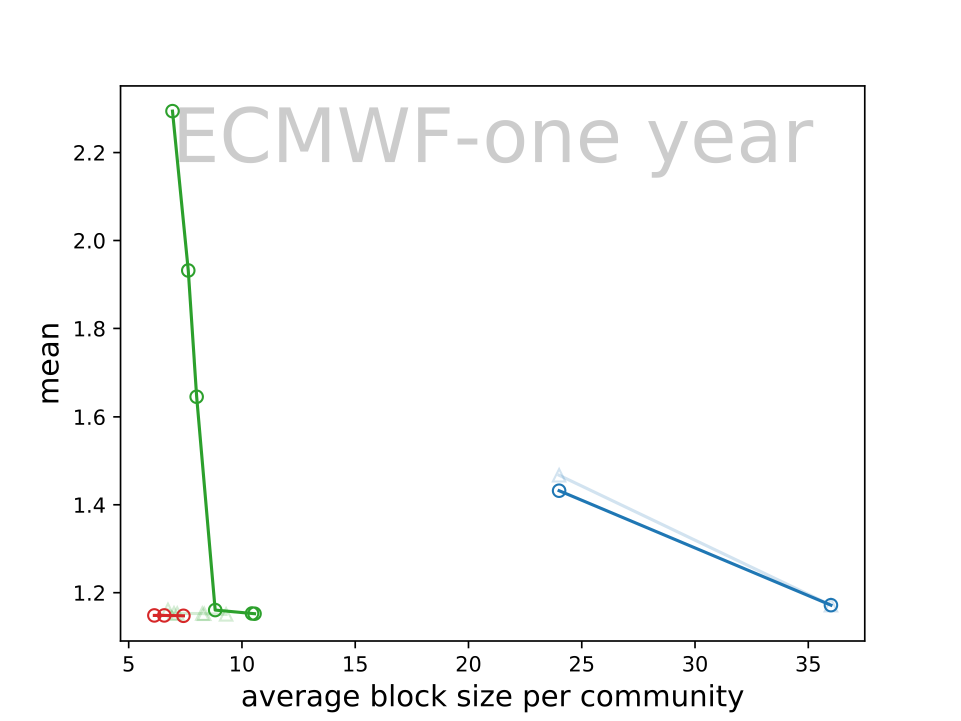}\\
\mbox{(a)} & \mbox{(b)}
\end{array}$
\end{center}
\caption{\hot{The particle density estimation (a) and the community detection (b) results for the unsteady flow fields. The first row shows the results of the hurricane data set using all time steps. The second row shows the results of the ECMWF data set using all time steps for training and testing. The third row shows the results of the ECMWF data set using one year (12 time steps) for training and two years (24 time steps) for testing.}}
\label{fig:unsteady}
\end{figure}

\hot{
\subsection{Preliminary Experiment with Unsteady Flows}
\label{sec:unsteady}

{\bf Experiment setup.} We experiment with our FlowHON on two unsteady flow data sets: the hurricane data set, and the European Center for Medium Range Weather Forecasts (ECMWF) data set. The hurricane data set contains 48 time steps (one per hour) over the Mexico Gulf, while the ECMWF data set contains 44 time steps (one per month) over the earth. Therefore, the hurricane data set depicts a relatively short-term atmospherical flow in a local region, and the ECMWF data set describes a long-term global unsteady flow. For each time step, we generate one thousand pathlines starting from that time step. For the ECMWF data set, we use a very small time interval between tracing steps as the interval between the time steps is relatively large. Therefore, the pathline may be similar to streamlines traced at individual time steps for this data set. The pathlines are first converted into sequences of blocks. Then, the networks are constructed from the block sequences using exactly the same scheme as the construction for steady flows. 

{\bf Exploration results.} For unsteady flow fields, the trajectories of particles inside each data block may be even more diversified. Figure~\ref{fig:unsteady-vis} (a) to (d) shows the pathlines corresponding to four major HO-nodes in the selected block in red. The HO-nodes are listed in the order of the number of pathlines related to them. The orange and the green pathlines represent the two most prominent patterns, which seem similar in shape. But further investigation shows that the two groups of pathlines move in the opposite direction: the orange ones move westward, and the green ones move eastward, as indicated by the red arrows. The other two groups of pathlines both reveal spiral patterns. The red group corresponds to spirals going outward from the selected block, while the purple group corresponds to spirals moving between the selected block and the neighboring block on the right. Furthermore, these patterns may change over time. For example, Figure~\ref{fig:unsteady-vis} (e) and (f) show that the westward wind (orange) is much stronger at the time step 8 than the time step 2. This demonstrates the power of {\ours} to distinguish different flow patterns inside a single block without consulting additional time step information. In contrast, {\fon} assumes all particles in a block follow the same transition distribution. Therefore, {\fon} can only guess that a particle has similar chances to move eastward and westward at any time step, which is inaccurate.

{\bf Quantitative results.} Figure~\ref{fig:unsteady} compares the performance of different networks on the particle density estimation and the community detection. The first two rows show the results using all time steps for training and testing. In terms of the density estimation, we find that the networks with our edge optimization clearly outperform the original approaches, especially for {\fon} and {\semantic}. For {\fixed}, the edge optimization does not lead to an obvious improvement. This reveals that the finest-level HO-states play a major role in the estimation, while the optimization only recovers the information on a simplified network with a reduced number of nodes. For the unsteady flows, we find that our {\ours} outperforms {\semanticPlus} and is very close to {\fixedPlus}. This demonstrates the effectiveness of our node initialization and iterative update scheme. For the ECMWF data set, as it contains many periodic flow patterns over years, we further examine whether the data in previous years can be used to predict the transitions in later years. In the third row of Figure~\ref{fig:unsteady}, we find that the edge optimization still provides a performance gain, but the estimation error clearly increases after a few time steps for all approaches. This may indicate that the difference between transition patterns of different years may require a more sophisticated optimization technique. In terms of the community detection result, we still find that {\ours} outperforms {\fonPlus} and {\semanticPlus} as the particles visit fewer communities on average with smaller community sizes.  
}

\hot{
\subsection{Domain Expert Feedback}
We invite Dr. Jingkun Chen, an expert in fluid dynamics, to evaluate the effectiveness of {\ours}. Dr. Chen has more than ten years of experience in this field, and his recent work focuses on developing scalable computational tools for various scientific domains, including atmospheric science. The evaluation was performed in two stages. In the first stage, Dr. Chen was introduced to the basic concepts of {\ours} and the exploration interface using the tornado data set. In the second stage, Dr. Chen used the interface to explore the ECMWF data set and compared the {\fon} and {\ours} on it, as this data set is closely related to his application of interest.

Dr. Chen stated that ``{\ours} is useful in discovering features as the complex flow regions often seem to be denser than other regions." This may be related to the fact that {\ours} will produce more HO-nodes in the blocks where particles exhibit more diverse transition patterns. He further commented that ``The higher-order nodes separate the feature patterns from the other flow lines, which is useful to study the specific physical phenomenon. For example, in the atmospherical data sets, the spiraling pattern, such as typhoons, is often studied. {\ours} enables the selection of this kind of features, but in {\fon}, this pattern is hidden in the laminar flows and cannot be selected." This is indeed consistent with the findings in our exploration. However, in terms of the overall structure of the graphs, Dr. Chen stated that ``The graph visualization of the two networks ({\fon} and {\ours}) reveal similar structures."  

Dr. Chen also mentioned some desired features that may be supported by {\ours} in the future. He stated that ``When an HO-node is selected, it will be more convenient if some functions are provided to explore its neighbors in the graph. For the atmospherical science, this may be helpful to study the water vapor transmission, the pollution diffusion, and the energy circulation." As {\ours} tends to provide a more deterministic transition behavior among nodes, we feel that this is a promising direction to explore. For example, starting from one node, {\ours} can better estimate the regions that will be influenced by the selected node. He also mentioned that ``{\ours} is an interesting idea to handle the dependencies in computation. It will be interesting to see whether {\ours} can work with the computation tools we are developing."
}
\section{Conclusions and future work}

We propose {\ours} that describes particle transitions at block level using higher-order networks. We formulate the higher-order network construction as an optimization problem of three linear layers, corresponding to the distribution from blocks to higher-order states, the aggregation from higher-order states to higher-order nodes, and the transitions among higher-order nodes. Higher-order nodes subdivide the flow behaviors in individual blocks, leading to better accuracy in describing particle transitions among blocks. The higher-order network connects the higher-order nodes and describes the transition patterns at a larger scale. 
\hot{We evaluate the effectiveness of our approach through comparison with existing graph-based approaches from different aspects. We experiment on both steady and unsteady flow fields using three downstream tasks. Our results show that {\ours} outperforms existing approaches in most cases, and our optimization approach could help existing approaches boost their performance as well. In addition, we conduct an empirical evaluation with a domain expert to examine the effectiveness of FlowHON in exploring flow fields and discuss the potential applications.}

In the future, we would like to explore the following directions.
\hot{First, we would like to develop scalable parallel particle tracing techniques on top of the {\ours}. Our simple parallel particle tracing platform with the {\ours} has demonstrated the potential of {\ours} in reducing particle exchanges and tracing iterations. But workload balance in the platform remains an open question. We would like to further explore whether the higher-order nodes will provide more accurate information to estimate the workload in each community as well. Based on the accurate transition pattern and workload estimation, we may be able to develop a more sophisticated parallel tracing platform for large-scale, efficient particle tracing.}
Second, we would like to investigate the on-the-fly construction of the {\ours}. In this way, the network is dynamically updated during the particle tracing, which could improve the performance of later tracing steps. This may be useful for unsteady flow fields with the assumption that the transition patterns do not change abruptly.
Third, we would like to seek a better balance between the explainability of our linear transformation-based approach and the power of deep learning approaches. Deep neural networks may be used to identify meaningful higher-order states with even longer dependencies. 


\vspace{-0.1in}
\bibliographystyle{abbrv}
\bibliography{reference}

\begin{IEEEbiography}[{\includegraphics[width=1in,height=1.25in,clip,keepaspectratio]{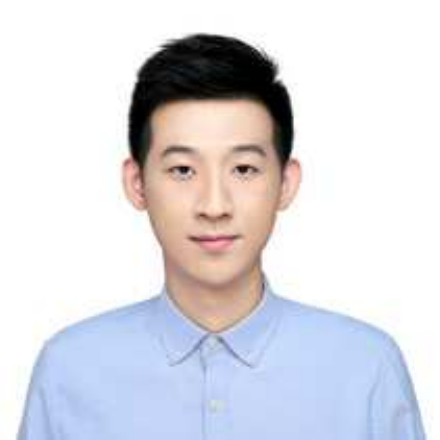}}]{Nan Chen} 
is currently a research assistant at National University of Singapore. He received a B.E.\ degree in computer engineering from Sun Yat-sen University in 2019, and a Master of Computing degree from National University of Singapore in 2022. His research interests are scientific visualization and multivariate data exploration. 
\end{IEEEbiography}

\begin{IEEEbiography}[{\includegraphics[width=1in,height=1.25in,clip,keepaspectratio]{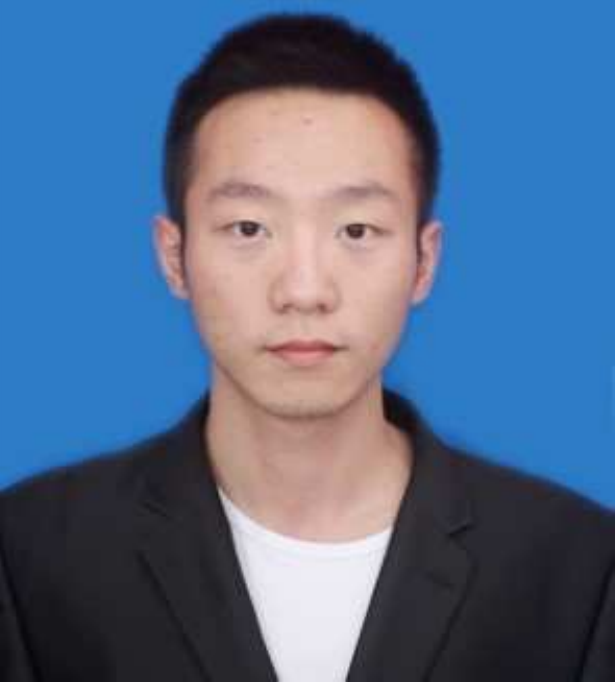}}]{Zhihong Li} 
is a master student at Sun Yat-sen University. He received a B.E.\ degree in computer science from Liaoning University in 2020. His research interests are flow visualization and deep learning for scientific visualization.
\end{IEEEbiography}

\begin{IEEEbiography}[{\includegraphics[width=1in,height=1.25in,clip,keepaspectratio]{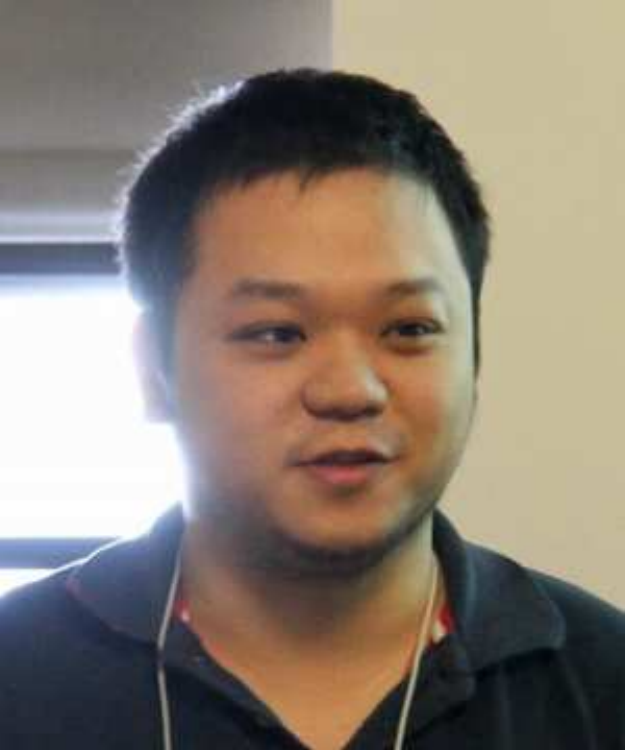}}]{Jun Tao} 
is an associate professor of computer science at Sun Yat-sen University and National Supercomputer Center in Guangzhou. He received a Ph.D.\ degree in computer science from Michigan Technological University in 2015. Dr.\ Tao's major research interest is scientific visualization, especially on applying information theory, optimization techniques, and deep learning to flow visualization and multivariate data exploration.
\end{IEEEbiography}

\end{document}